\documentclass[a4paper,12pt,headsepline,twoside]{scrartcl}
\newcommand{\titleDocument}{Bachelor Thesis}

\usepackage{graphicx}

\usepackage{amsmath}
\newcommand\norm[1]{\left\lVert#1\right\rVert}
\newcommand\abs[1]{\left\vert#1\right\vert}
\newcommand\normalized[1]{\left\langle#1\right\rangle}
\DeclareMathOperator{\fft}{FFT}
\DeclareMathOperator{\accsum}{accsum}
\DeclareMathOperator{\rank}{rank}

\DeclareMathOperator*{\argmin}{arg\,min}
\DeclareMathOperator*{\argmax}{arg\,max}

\usepackage{xfrac}

\usepackage{tikz}
\usetikzlibrary{fit,positioning}

\usepackage{pgf-pie}
\newif\ifpienumberinlegend
\pgfkeys{/number in legend/.code=
    \expandafter\let\expandafter\ifpienumberinlegend
    \csname if#1\endcsname
    \ifpienumberinlegend

    \def\beforenumber##1\afternumber{}%
    \fi,
    /number in legend/.default=true
}

\usepackage{pgfplots}
\usepackage{pgfplotstable}

\usepackage{tabu}
\usepackage{xhfill}
\usepackage{multirow}
\newcommand{\ditto}{---\textquotedbl---}
\newcommand{\specialcell}[2][c]{\begin{tabular}[#1]{@{}c@{}}#2\end{tabular}}

\usepackage{float}

\usepackage{xargs}

\usepackage{hhline}

\usepackage[hypcap=true]{caption}

\makeatletter

\usepackage[german,english]{babel}

\usepackage[right]{eurosym}

\usepackage[utf8]{inputenc}

\usepackage[T1]{fontenc}

\usepackage{lmodern}
\usepackage{fix-cm}

\usepackage{longtable}

\usepackage{geometry}

\usepackage{fancybox}

\usepackage{diagbox}

\usepackage[hyphens,obeyspaces,spaces]{url}

\usepackage[hang,marginal,stable]{footmisc}

\usepackage{subcaption}

\usepackage{amssymb}

\usepackage{xcolor}% http://ctan.org/pkg/xcolor
\definecolor{tu-blue}{cmyk}{1,0.7,0,0}
\definecolor{tu-red}{cmyk}{0,1,1,0}
\definecolor{emerald}{RGB}{46,204,113}

\usepackage[bookmarksnumbered,pdftitle={\titleDocument},hyperfootnotes=false]{hyperref} 
\hypersetup{colorlinks, citecolor=emerald, linkcolor=tu-blue, urlcolor=tu-blue}

\graphicspath{{abb/}}

\usepackage{fancyhdr} %Paket laden
\usepackage{array}

\usepackage{setspace}

\usepackage{capt-of}

\usepackage{makeidx}

\usepackage{listings}
\makeindex

\usepackage{nomencl}

\makenomenclature
\begin{document}

\newcommand\citeneeded{\textcolor{tu-red}{ \textbf{??CITE??} }}
\newcommand\todo[1]{
  \par
  \textcolor{tu-red}{\textbf{TODO:}} #1
  \par
}

% hier werden die Trennvorschläge inkludiert
%hier müssen alle Wörter rein, welche Latex von sich auch nicht korrekt trennt bzw. bei denen man die genaue Trennung vorgeben möchte
\hyphenation{
Film-pro-du-zen-ten
Lux-em-burg
Soft-ware-bau-steins
zeit-in-ten-siv
Kai-sers-lau-tern
}

%Schriftart Helvetica
%\changefont{phv}{m}{n}

% Leere Seite am Anfang
%\newpage
%\thispagestyle{empty} % erzeugt Seite ohne Kopf- / Fusszeile
%\section*{ }

% Titelseite %
%------------------ !!! RE-INCLUDE !!!------------------------
% das Papierformat zuerst
%\documentclass[a4paper, 11pt]{article}

% deutsche Silbentrennung
%\usepackage[ngerman]{babel}

% wegen deutschen Umlauten
%\usepackage[ansinew]{inputenc}

% hier beginnt das Dokument
%\begin{document}

\thispagestyle{empty}

%\begin{figure}[t]
% \includegraphics[width=0.6\textwidth]{abb/fh_koeln_logo}
%\end{figure}

\begin{figure}[t]
 \centering
 \includegraphics[width=0.6\textwidth]{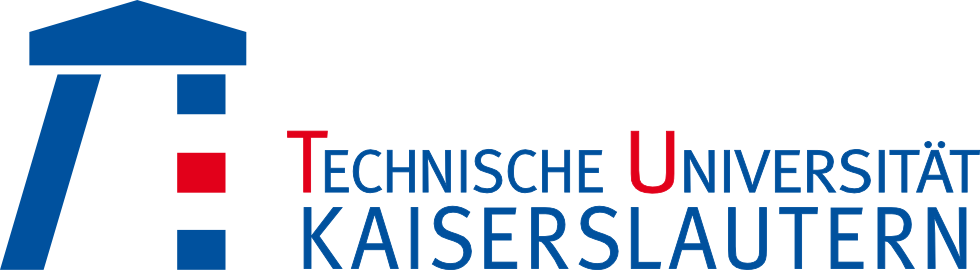}
~~~~~~~~~~
 \includegraphics[width=0.3\textwidth]{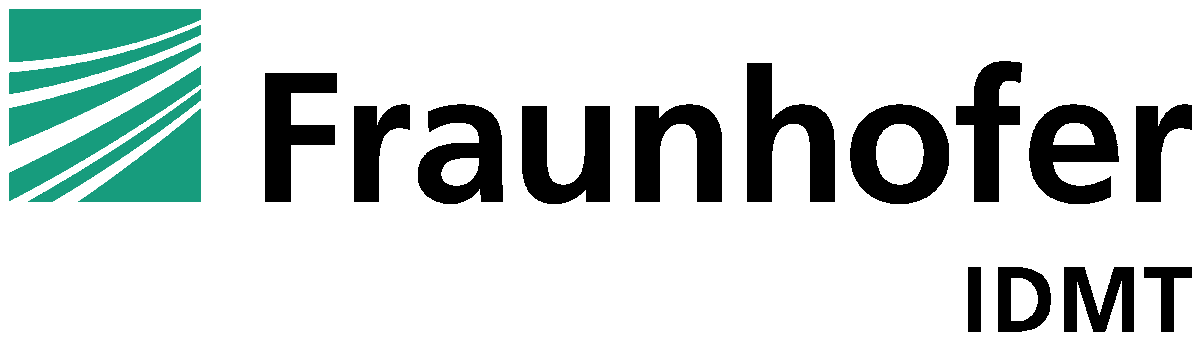}
\end{figure}

\begin{verbatim}


\end{verbatim}

\begin{center}
\Large{Technische Universität}\\
\Large{Kaiserlautern}\\
\end{center}

\begin{center}
\Large{Fachbereich Informatik}
\end{center}
\begin{verbatim}


\end{verbatim}

\begin{center}
\doublespacing
\textbf{\LARGE{\titleDocument}}\\
\singlespacing
\end{center}
\begin{verbatim}











\end{verbatim}
\begin{center}
\textbf{zur Erlangung des akademischen Grades \\ Bachelor of Science}
\end{center}
\begin{verbatim}





\end{verbatim}
\begin{flushleft}
\begin{tabular}{llll}
\textbf{Thema:} & & A hybrid approach to supervised machine learning\\
& & for algorithmic melody composition & \\
& & \\
\textbf{Autor:} & & Rouven Bauer <r\_bauer11@cs.uni-kl.de>& \\
\\
& & \\
\textbf{Version vom:} & & 23. Oktober 2016 &\\
& & \\
\textbf{1. Betreuer:} & & Prof. Dr. Prof. h.c. Andreas Dengel &\\
\textbf{2. Betreuer:} & & Dr. rer. nat. Stephan Baumann  &\\
\textbf{externe Betreuung:} & & Dr.-Ing. Estefanía Cano Cerón  &\\
\end{tabular}
\end{flushleft}
%------------------ !!! RE-INCLUDE !!!------------------------

\newpage
\thispagestyle{empty} % erzeugt Seite ohne Kopf- / Fusszeile
\section*{ }
\cleardoublepage
\setcounter{page}{1}

% römische Numerierung
%\pagenumbering{arabic}

% 1.5 facher Zeilenabstand
\onehalfspacing

% Sperrvermerk
%\input{sperrvermerk}

% Danksagung
\section*{Acknowledgements}

This work would not have been possible without the great help and supervision of Dr.-Ing.~Estefan\'ia~Cano~Cer\'on at Fraunhofer IDMT in Ilemnau. Thanks for all the time, inspiring discussions and ideas. Thanks also to the supervisors at DFKI Kaiserslautern Dr. Stephan Baumann and Prof. Anreas Dengel as well as all other people giving their support in manners like administrative support, guidance, proof-reading and motivation.

\newpage
\thispagestyle{empty} % erzeugt Seite ohne Kopf- / Fusszeile
\section*{ }

% Einleitung / Abstract
\section*{Abstract}

\begin{abstract}
\noindent\textbf{English:}\\
In this work we present an algorithm for composing monophonic melodies similar in style to those of a given, phrase annotated, sample of melodies. For implementation, a hybrid approach incorporating parametric Markov models of higher order and a contour concept of phrases is used. This work is based on the master thesis of Thayabaran Kathiresan (2015). An online listening test conducted shows that enhancing a pure Markov model with musically relevant context, like count and planed melody contour, improves the result significantly.
\end{abstract}

\begin{abstract}
\noindent\textbf{German:}\\
\begin{otherlanguage}{german}
In dieser Arbeit wird ein Algorithmus präsentiert, der einstimmige Melodien komponiert. Diese sind stilistisch ähnlich zu Beispielmelodien, deren Phrasengrenzen zuvor annotiert wurden. Dazu wird ein Hybridansatz verfolgt, der parametrische Markow-Modelle höherer Ordnung, sowie ein Konturenkonzept implementiert. Diese Arbeit basiert auf der Masterarbeit von Thayabaran Kathiresan (2015). Ein Onlineumfrage hat ergeben, dass das erweitern eines reinen Markow-Modells mit musikalisch relevantem Kontext, wie Zählzeit und geplanter Melodiekontur, die Ergebnisse signifikant verbessert.
\end{otherlanguage}
\end{abstract}

\newpage
\thispagestyle{empty} % erzeugt Seite ohne Kopf- / Fusszeile
\section*{ }

% einfacher Zeilenabstand
\singlespacing

% Inhaltsverzeichnis anzeigen
\newpage
\tableofcontents

\newpage

% das Abbildungsverzeichnis
%\newpage
% Abbildungsverzeichnis soll im Inhaltsverzeichnis auftauchen
% Abbildungsverzeichnis endgueltig anzeigen
%\addcontentsline{toc}{section}{List of Figures}
\listoffigures

% das Tabellenverzeichnis
%\newpage
% Tabellenverzeichnis soll im Inhaltsverzeichnis auftauchen
%\addcontentsline{toc}{section}{List of Tables}
% \fancyhead[L]{Abbildungsverzeichnis / Abkürzungsverzeichnis} %Kopfzeile links
% Tabellenverzeichnis endgueltig anzeigen
\listoftables

%% WORKAROUND für Listings
%\makeatletter% --> De-TeX-FAQ
%\renewcommand*{\lstlistoflistings}{%
%  \begingroup
%    \if@twocolumn
%      \@restonecoltrue\onecolumn
%    \else
%      \@restonecolfalse
%    \fi
%    \lol@heading
%    \setlength{\parskip}{\z@}%
%    \setlength{\parindent}{\z@}%
%    \setlength{\parfillskip}{\z@ \@plus 1fil}%
%    \@starttoc{lol}%
%    \if@restonecol\twocolumn\fi
%  \endgroup
%}
%\makeatother% --> \makeatletter
% das Listingverzeichnis
%\newpage
% Listingverzeichnis soll im Inhaltsverzeichnis auftauchen
%\addcontentsline{toc}{section}{Listingverzeichnis}
%\fancyhead[L]{Abbildungs- / Tabellen- / Listingverzeichnis} %Kopfzeile links
%\renewcommand{\lstlistlistingname}{Listingverzeichnis}
%\lstlistoflistings
%%%%

%% das Abkürzungsverzeichnis
%\newpage
%% Abkürzungsverzeichnis soll im Inhaltsverzeichnis auftauchen
%\addcontentsline{toc}{section}{Index of Abbreviations}
%% das Abkürzungsverzeichnis entgültige Ausgeben
%\fancyhead[L]{Index of Abbreviations} %Kopfzeile links
%\input{latex_einstellungen/abkuezungen/abkuerzungen}
%\printnomenclature

% das Abkürzungsverzeichnis
\newpage
% Abkürzungsverzeichnis soll im Inhaltsverzeichnis auftauchen
\addcontentsline{toc}{section}{Definitions}
% das Abkürzungsverzeichnis entgültige Ausgeben
\fancyhead[L]{Definitions} %Kopfzeile links
\section*{Definitions}
\label{sec:definitions}
In this thesis we use the following mathematical notation:
\begin{itemize}
\item Sets
\begin{itemize}
\item $ \abs{X} $ for a set $ X $ is it's cardinality (i.e. the number of elements contained).
\item $ [a,\,b] \subset \mathbb{R},\ a,b \in \mathbb{R} $ is an interval of the real numbers. It contains all elements $ x \in \mathbb{R}\colon a \leq x \leq b $.
\item $ (a,\,b) \subset \mathbb{R},\ a,b \in \mathbb{R} $ is an interval of the real numbers. It contains all elements $ x \in \mathbb{R}\colon a < x < b $.
\item Combinations like $ [a,\,b) $ and $ (a,\,b]$ are possible as well.
\item $ \{a,\,\ldots,\,b\} \subset \mathbb{Z},\ a,b \in \mathbb{Z} $ is an interval of integers. It contains all elements $ x \in \mathbb{Z}\colon a \leq x \leq b $.
\end{itemize}
\item Functions
\begin{itemize}
\item $ f\colon X \mapsto Y $ is a function (or mapping), called $ f $, from the set $ X $ to the set $ Y $.
\item $ \varphi_{\mu, \sigma^2}(x) = \frac{1}{\sigma\sqrt{2\pi}}e^{-\frac{(x-\mu)^2}{2\sigma^2}} $ is the Gaussian distribution with $ \mu $ as mean and $ \sigma^2 $ as standard deviation.
\item $ \fft (f) $ is the fast Fourier transform of the function $ f $.
\item $ \fft^{-1}(f) $ is the inverse fast Fourier transform of the function $ f $.
\end{itemize}
\item Vectors
\begin{itemize}
\item $ \vec{x} $ is a vector.
\item For an n-dimensional vector $ \vec{x} \in \mathbb{R}^n $ we define $ \dim(\vec{x}) = n $ as the dimension of the vector.
\item $ \norm{\vec{x}} = \sqrt{\sum\limits_i \vec{x}_i ^{\,2}} $ is the Euclidean length of a vector $ \vec{x} $.
\item $
\normalized{\vec{x}} = 
\begin{cases}
  \hfil\frac{\vec{x}}{\norm{\vec{x}}} & \norm{\vec{x}} \neq 0 \\
  \hfil\vec{0}                        & \text{else}
\end{cases}
$ is a vector normalized by it's Euclidean length.
\item $ \vec{a}\bigodot\vec{b} = \vec{c} $ with $ \vec{c}_i = \vec{a}_i\cdot\vec{b}_i $ is the element wise multiplication of two vectors.
\item $ \vec{b} = \accsum(\vec{a}) $, the accumulative sum defined element-wise as 
$\vec{b_i} = \sum\limits_{j \leq i} \vec{a_j} $.
\end{itemize}
\end{itemize}

\newpage
\thispagestyle{empty} % erzeugt Seite ohne Kopf- / Fusszeile
\section*{ }

% Definiert Stegbreite bei zweispaltigem Layout
\setlength{\columnsep}{25pt}

%%%%%%% EINLEITUNG %%%%%%%%%%%%
%\twocolumn
\newpage
\fancyhead[L]{\nouppercase{\leftmark}} %Kopfzeile links

% 1,5 facher Zeilenabstand
\onehalfspacing

\section[Introduction]{Introduction}
\label{sec:introduction}
Artificial intelligence has made huge progress over the last decades. One topic still fascinating and demanding for further research is \emph{computational creativity} (sometimes also referred to as \emph{artificial creativity}). Either attempting to understand what human creativity is and how it works by developing computational models for it or driven by the attempt to outperform humans at one of the tasks they are probably still better in than computers. Composing music is one of those creative tasks. Music is a concept known to all cultural groups around the world. It plays an important role in many social contexts like religion and plain entertainment.

The idea of automatic composition is an old one. W. A. Mozart proposed such a method in 1798 \cite{Mozart1798}. Since there were no computers at his time he utilized a pair of dice to randomly recombine a set a bars to compose \emph{Countrydances}. There were as many as $ 1.85 \cdot 10^{17} $ different pieces that could be composed with his method. It is an example of automatically composing a melody with harmonization arranged for piano all at once. It is possible though, to subdivide the composing process into several task. Examples are \emph{rhythm generation}, \emph{harmonization} and \emph{melody composition}. This work focuses on the latter one.

\subsection{Motivation and Goal}
\label{sec:motivation}
One definition of music is ``organized sound''. This definition is a very broad one. We want to focus on western style music, more exactly melodies, that can be written down in a traditional staff. It's worth noting that written music still leaves a lot of interpretive freedom, like slight variations in pitch or time as well as agogics, for each musician that may play it. This music interpretation process is a different topic that goes beyond the scope of our work. An example of such work was proposed by Flossmann et al. \cite{Flossmann2013}. For a broad overview of the topic the interested reader is referred to the survey by Kirke and Miranda \cite{Kirke2009}.

What our system tries to achieve is composing monophonic melodies similar to those of a given training set of melodies. It utilizes higher order Markov models and other statistical models to do so. We decided to follow this approach as this work builds upon the master thesis of Kathiresan \cite{Kathiresan2015} which incorporates Markov models as well. The source code can be found on GitHub (\url{https://github.com/roba91/melody-composer}).

A system like this could find applications in generating melodies for practicing an instrument or music theoretical tasks like harmonization with as many melodies as a student pleases. Since it's more fun and therefore more motivating to play or analyze melodies that are pleasant and don't feel unnatural or artificial, we tried to improve the original system by Kathiresan (see Section \ref{subsubsec:master-thesis}) in a way to compose more human-like melodies than it was able to create before.

\section{Previous Work}
\label{sec:previous_work}
\subsection{State of the Art}
\label{subsec:state_of_the_art}
There are several ways to structure the previous work on the topic of algorithmic composition. Here, we will stick to the taxonomy as proposed by Fernández and Vico \cite{Fernandez2013}. Their survey paper provides a great overview of the state of the art and is recommended to the reader for a more in depth insight.

% taxonomy
% \TC@ntainer[styles]{width}[name]{content}
\newcommandx\TC@ntainer[4][1={}, 3={}]{%
  \tikz\node[inner sep=0pt,outer sep=0pt,fill=white,fill opacity=0,text width=#2,align=center,text opacity=1,text height={},#1](#3) {#4};%
}
% \TContainer[styles][name]{content}
\newcommandx\TContainer[3][1={}, 2={}]{%
  \TC@ntainer[#1]{\textwidth}[#2]{#3}%
}
% \TBox[styles]{width}[name]{content}
\newcommandx\TBox[4][1={}, 3={}]{%
  \tikz\node[inner sep=3pt,outer sep=0pt,draw,fill=black,fill opacity=0.1,thick,text width=#2-7pt,align=center,rounded corners=0.2cm,minimum height=1cm,minimum width=0,text height={},text opacity=1,#1](#3) {#4};%
}
\newbox\@LBox
\newbox\@RBox
\newlength\@LHeight
\newlength\@RHeight
\newlength\@Diff
% \LRXSameHeight{\TBox or \TC@ntainer} [styles1]{width1}{content1} [styles2]{width2}{content2}
\newcommandx\LRXSameHeight[7][2={}, 5={}]{%
  \setbox\@LBox\hbox{#1[#2]{#3}{#4}}%
  \setbox\@RBox\hbox{#1[#5]{#6}{#7}}%
  \@LHeight=\ht\@LBox
  \@RHeight=\ht\@RBox

  % new line fixes things?!?
  \ifnum\@LHeight>\@RHeight
  	\settoheight\@RHeight{#1[#5,minimum height=\@LHeight]{#6}{\vbox to \@LHeight{#7}}}%
  	\setlength\@Diff{\@LHeight}%
  	\addtolength\@Diff{-\@RHeight}%
  	\addtolength\@LHeight{\@Diff}%
    #1[#2]{#3}{#4} % space for more nice!
    #1[#5,minimum height=\@LHeight]{#6}{\vbox to \@LHeight{#7}}
  \else
  	\settoheight\@LHeight{#1[#2,minimum height=\@RHeight]{#3}{\vbox to \@RHeight{#4}}}%
  	\setlength\@Diff{\@RHeight}%
  	\addtolength\@Diff{-\@LHeight}%
  	\addtolength\@RHeight{\@Diff}
    #1[#2,minimum height=\@RHeight]{#3}{\vbox to \@RHeight{#4}} % space for more nice!
    #1[#5]{#6}{#7}%
  \fi  
}
\newcommand\LRContainerSameHeight{%
  \LRXSameHeight{\TC@ntainer}%
}
\newlength\@boxIboxheight
\newlength\@boxIIboxheight
% assuming border widths of box1 and box2 are equal
\newcommandx\LRBoxSameHeight{%
  \LRXSameHeight{\TBox}%
}
\newcommandx\TestBox[4][1={}, 3={}]{
	1: #1, 2: #2, 3: #3, 4: #4
}

%\tracingall
\begin{figure}[htb]
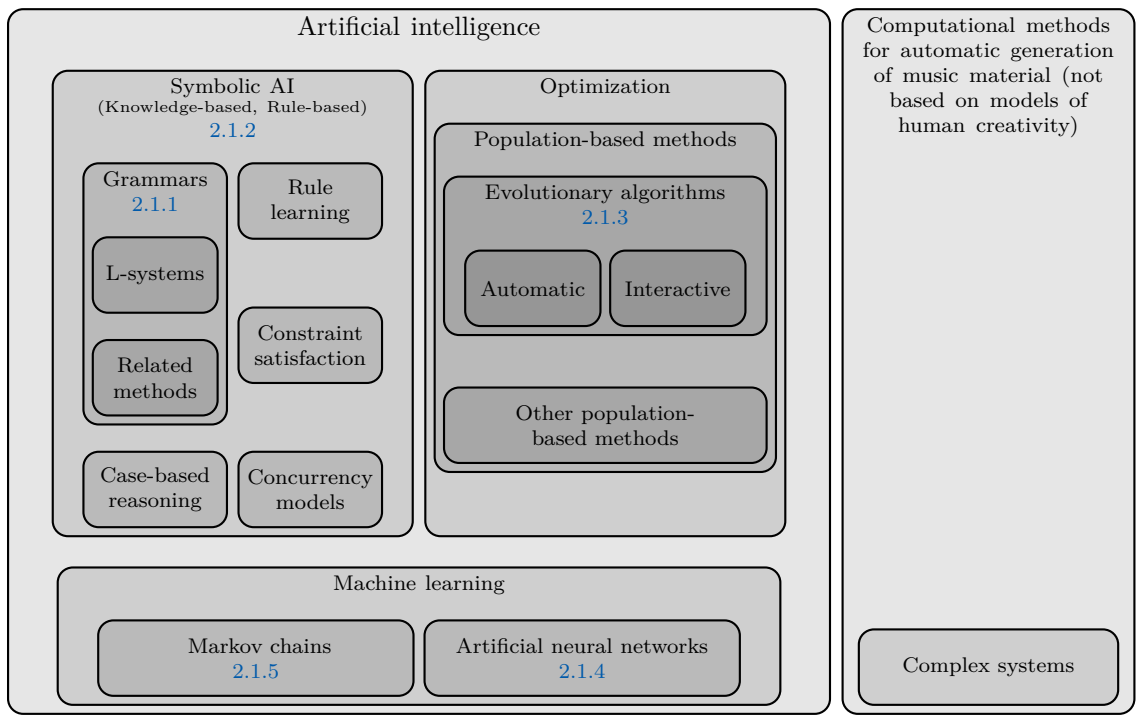

\LRBoxSameHeight{0.70\textwidth}{
    \footnotesize Artificial intelligence\\
    \vspace{1em}
    \LRBoxSameHeight{0.45\textwidth}{
      \scriptsize Symbolic AI\\
      \tiny (Knowledge-based, Rule-based)\\
      \scriptsize\ref{subsubsec:gmSymbolicAI}\\
      \vspace{1em}
      \LRContainerSameHeight{0.425\textwidth}{
        \TBox{\textwidth}{
          \scriptsize{Grammars\\
          \ref{subsubsec:gmGrammars}}\\
          \vspace{1em}
          \TBox{\textwidth}{
            \scriptsize L-systems
          }\\
          \vspace{1em}
          \TBox{\textwidth}{
            \scriptsize Related methods
          }%
        }\\
        \vspace{1em}
        \TBox{\textwidth}{
          \scriptsize Case-based reasoning
        }%
      }{0.425\textwidth}{
        \TBox{\textwidth}{
		  \scriptsize Rule learning
		}%
        \vfill
        \TBox{\textwidth}{
		  \scriptsize Constraint satisfaction
		}%
        \vfill
        \TBox{\textwidth}{
		  \scriptsize{Concurrency models}
		}%
	  }	
    }{0.45\textwidth}{
      \scriptsize Optimization\\
      \vspace{1em}
      \TBox{\textwidth}{
        \scriptsize Population-based methods\\
        \vspace{1em}
        \TBox{\textwidth}{
          \scriptsize{Evolutionary algorithms\\
          \ref{subsubsec:gmEvolutionaryAlgorithms}}\\
          \vspace{1em}
          \TBox{0.45\textwidth}{
            \scriptsize Automatic
          } % space for nice!
          \TBox{0.45\textwidth}{
            \scriptsize Interactive
          }%
        }
        \vspace{1em}\\
        \TBox{\textwidth}{
          \scriptsize Other population-based methods
        }%
      }%
    }\\
    \vspace{1em}
    \TBox{0.9\textwidth}{
      \scriptsize Machine learning\\
      \vspace{1em}
      \TBox{0.45\textwidth}{
        \scriptsize{
          Markov chains\\
          \ref{subsubsec:gmMarkovChains}\\
        }
      } % space for nice!
      \TBox{0.45\textwidth}{
        \scriptsize{Artificial neural networks\\
        \ref{subsubsec:gmNeuralNetworks}}
      }%
    }%
  }%
  {0.25\textwidth}{
    \scriptsize Computational methods for automatic generation of music material
          (not based on models of human creativity)\\
    \vfill
    \TBox{0.95\textwidth}{
      \scriptsize Complex systems
    }%
}

 \caption[Taxonomy of methods of previous work]{Taxonomy of approaches in previous work of computer composed music as proposed by Fernández and Vico \cite{Fernandez2013}. Not all fields are covered in our work because we want to focus on the major approaches and those relevant to ours.}
\end{figure}

It is worth mentioning that it is very difficult in general to clearly categorize all previous works by their approach. This is because many of them are hybrid approaches (like the one proposed by us), utilizing more the one \emph{artificial intelligence} (AI) approach. Nevertheless, we categorized them by their most dominating approach.
  
\subsubsection{Generating Melodies with Grammars}
\label{subsubsec:gmGrammars}
\emph{Lindenmayer Systems} (L-systems) and \emph{Chomsky grammars} have in common that they are an intuitive choice for algorithmic composition. Music is very often repetitive and self-similar. This self-similarity can be found on multiple scales in music. In groups of just a few notes up to bars, phrases, whole compositions (as in movements) or sometimes even across pieces. This self-similarity corresponds to the hierarchical structure of grammars.

Grammars are defined by a set of \emph{production rules} that are applied recursively in so called \emph{derivations} to substitute parts of a symbol chain with other symbols. L-systems work similarly, but in every derivation step, all possible rules are applied simultaneously. Such a generated symbol chain must then be mapped to a musical object. Often, this is a simple translation from the grammar alphabet to aspects of a composition, such as pitches, durations, or chords. Moorer \cite{Moorer1972} and Langston \cite{Langston1989} proposed systems that apply these methods.

One important step in grammar driven composition is to generate the rules that define the grammar. There are two basic strategies for this: 1) One can either derive the rules manually from music theory as Rader \cite{Rader1974} did, or 2) the recently more often chosen strategy, is to abstract from the musical structure of a training dataset to obtain those rules as Gillick et al. \cite{Gillick2009} did.

\subsubsection{Generating Melodies with other Methods of symbolic AI}
\label{subsubsec:gmSymbolicAI}
\emph{Rule-based systems} are another intuitive choice for algorithmic composition. Music theory offers a lot of rules on how to compose music, arrange it and other tasks. Some examples of such rules that are well-founded and well-defined in music theory are counter point \cite{Morales1995} and basso continuo \cite{Wead2007}.

The rule set used for composing is commonly a static one during execution; however, there are some proposals that use a combination of static and dynamically learned rules \cite{Schwanauer1993}. The Literature also provides systems that derive their rules completely from training data \cite{Spangler1999}, or combine rule templates with training data \cite{Morales1995}.

Another approach from the field of symbolic AI is the formulation of a set of constraints. The musical task is then accomplished by solving the resulting \emph{constraint satisfaction problem} (CSP). A lot of work using CSPs was done on solving classical problems like harmonization or counterpoint composition \cite{Ramirez1998}. But other aspects of musical composition have been addressed as well. Two consecutive CSPs can be used to compose music following a storyboard modeling the mood of the composition as a function of time as shown by Zimmermann \cite{Zimmermann2001}. The first CSP generates a harmony progression from the storyboard while the second one then composes a four-part harmonization from the first's output. Another application of CSPs was proposed by Laruson and Kuuskankare \cite{Laurson2000}. They used constraints to ensure human playability of a melody in respect to the fingering for example.

\emph{Case-based reasoning} (CBR) is a further approach to implement rules of composition tasks \cite{Pereira1997}. CBR is an approach where the system starts off with a set of cases for which the solution to the problem is known. If a solution for a new (unknown) case is needed, the system tries to find a similar case and to modify the solution for it in order to solve the new case. This case-solution pair can then be added to the set of known cases.

\subsubsection{Generating Melodies with Evolutionary Algorithms}
\label{subsubsec:gmEvolutionaryAlgorithms}
Evolutionary algorithms in general start with an initial set (called \emph{population}) of candidate solutions (called \emph{individuals}, \emph{creatures} or \emph{phenotypes}) to a problem. In iterative steps, new populations (called \emph{generations}) are generated so that they are better candidate solutions. Therefore, a \emph{fitness function} must be defined that describes the quality of the candidates. The fitter individuals of the latest generation are selected and modified by so called \emph{cross-overs} and \emph{mutations} which are combinations and random changes of the individuals to form the next generation.

This approach has been used in two major forms: Espí et al. \cite{Espi2007} and Jensen \cite{Jensen2011} are recent examples of works using a well-defined fitness function. Nevertheless, it is very hard, if not impossible, to find a good mathematical model of aesthetic perception. Thus, approaches with human interaction replacing or complementing a defined fitness function by manually rating individuals were proposed. MacCallum \cite{MacCallum2012} published one of many examples.

\subsubsection{Generating Melodies with Neural Networks}
\label{subsubsec:gmNeuralNetworks}
\emph{Artificial neural networks}, called \emph{neural networks} for simplicity in this text, are a machine learning approach inspired by nature. These networks are composed of \emph{layers} built out of so called \emph{neurons}. In feed-forward networks, neurons take some or all (depending on the type of layer) neuron outputs of the previous layer as inputs. The inputs are then multiplied by \emph{weights} (usually different for all inputs), summed up and mapped with an \emph{activation function} (most simple activation function is the signum function). The weights are what is altered when a neural network is trained \cite{Kriesel2007, Duda2012}. \emph{Recurrent neural networks} like \emph{LSTMs}\footnote{long short-term memories} have more complicated link structures between \emph{neurons} that allow the network for having an internal state similar to a memory \cite{Hochreiter1997}. This comes in handy when one does not want to compose a melody as a whole but sequentially.

Neural networks as a tool in AI have experienced an enormous push forward in the last years as computing power grew and the training of \emph{deep neural networks} became a task with manageable time consumption \cite{Schmidhuber2015}. Hence, it's not surprising that people have proposed neural network driven approaches to computer composed music. The main advantage of those systems is their ability to capture repetitions of patterns over an arbitrary long period of time like Colombo et al. showed \cite{Colombo2016}. On the other hand, the main disadvantage of deep neural networks is their need for a high number of examples (training data). Their results suggest that their model reproduces rhythmical structures well but is not able to capture high level melodic structure like phrase repetitions in rhythm or melody. Another example of a melody composing algorithm utilizing LSTMs among other approaches has been proposed by Coca~et~al.~\cite{Coca2013}.

\subsubsection{Generating Melodies with Markov Chains}
\label{subsubsec:gmMarkovChains}
\emph{Markov Chains} are a statistical model introduced by the Russian mathematician A. A. Markov in the early 20\textsuperscript{th} century \cite{Hayes2013}. They can be thought of as directed graphs where the vertices are events/states, edges are transitions and edge weights are the transition probabilities between events/states. In practical implementations Markov models are mostly represented with probability matrices. A main problem of Markov models used for algorithmic composition tasks is that the next state only depends on the current one. For melodies, this results in randomly wandering around pitch sequences. The results can be improved by using Markov models of higher order ($m$-th order). Instead of only depending on the current state, the current and the last $m-1$ states are taken into account for generating the next one.

There are two main ways to obtain the transition probabilities: either derive them by hand i.e. construct and tweak them until the results are satisfying, or learn them from a training set of melodies. The former approach was often taken by composers aiming for supporting their composition process while the latter one was focused on in more scientific contexts \cite{Fernandez2013}.

The training process for Markov models is easy. For all states, all transitions are counted and then divided by the total number or transitions. Here is a small example: our training data contains four different symbols ``A'', ``B'', ``C'', and ``D''. The training data consists of three documents that look like this: \{``ABBA'', ``ACDC'', ``ACAB''\}.
For a second order Markov model, two blank symbols are prepended to all documents: \{``\textvisiblespace\textvisiblespace{}ABBA'', ``\textvisiblespace\textvisiblespace{}ACDC'', ``\textvisiblespace\textvisiblespace{}ACAB''\}. Now all following symbols for each bi-gram (that is not the last one of a document) are counted (see Table \ref{tab:bigram-count}). Finally all rows are normalized to sum up to one. The result is called a transition matrix. It contains the probability of next states for a series of previous states.

\begin{table}[htb]
  \begin{center}
    \begin{tabular}{c|c|*{4}{r|}}
      \multicolumn{2}{c}{}  & \multicolumn{4}{c}{next symbol}\\
      \hhline{~~----}
      \multicolumn{2}{c|}{}           &   A &   B &   C &   D \\
      \hhline{~-----}
      \multirow{7}{*}{bi-gram} &  \textvisiblespace\textvisiblespace &   3 &   0 &   0 &   0  \\
      \hhline{~-----}
                               &  \textvisiblespace{}A &   0 &   1 &   2 &   0  \\
      \hhline{~-----}
                               &  AB &   0 &   1 &   0 &   0  \\
      \hhline{~-----}
                               &  BB &   1 &   0 &   0 &   0  \\
      \hhline{~-----}
                               &  AC &   1 &   0 &   0 &   1  \\
      \hhline{~-----}
                               &  CD &   0 &   0 &   1 &   0  \\
      \hhline{~-----}
                               &  CA &   0 &   1 &   0 &   0  \\
      \hhline{~-----}
    \end{tabular}
    \caption[Transition matrix example]{An example of counting transitions to train a Markov model of second order. This transition matrix is calculated by normalizing each row to sum up to one.}
    \label{tab:bigram-count}
  \end{center}    
\end{table}

It was already discovered in early years of the computer age \cite{Moorer1972} that for lower order Markov models the resulting melodies wander aimlessly around in an unmusical way as mentioned above. For higher orders, one needs a big amount of training data or the results will be plain copies or at most recombinations of large musical fragments of the input. For intermediate orders the results are reasonable. One disadvantage, however, cannot be overcome by choosing a higher order Markov models: the composition process is only aware of a very local excerpt of it's own composition. In this thesis we try to overcome this issue by modifying the melody composition process of the Markov model with other statistical concepts (see Section~\ref{sec:proposed_method}).

\subsection{Related Work}
\subsubsection{``Automatic Melody Generation'' by Thayabaran Kathiresan}
\label{subsubsec:master-thesis}
In 2015, Kathiresan proposed a work on composing melodies with first order Markov models trained with a single Melody \cite{Kathiresan2015}. The Markov model was post processed with a CSP approach to include none or one of the four constraints: ``Include a note '', ``Include several notes '', ``Use only a set of notes '' and ``Include a given pattern of notes ''. Another approach using a hidden Markov model with categorical distributions was described as well. During evaluation, the first approach turned out to work better though. For this reason, we will focus on the formerly described approach throughout this work and call it KATH for shortness.

\begin{figure}[ht]
  \center
  \includegraphics[scale=0.5]{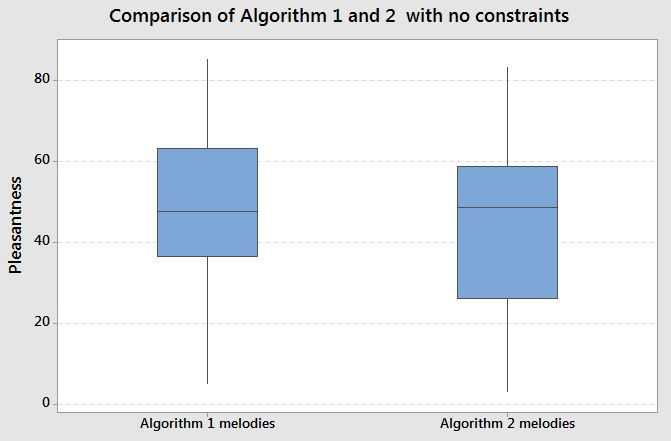}
  \caption[Pleasantness of Kathiresan's results]{How pleasant participants of Kathiresan's listening test found the presented melodies. \emph{Algorithm 1} is the one we used to compare our work to as it outperforms \emph{Algorithm 2}.\\
      \begin{footnotesize}
        Image extracted from Kathiresan's work \cite{Kathiresan2015}.
      \end{footnotesize}}
  \label{fig:kath-pleasantness}
\end{figure}

Kathiresan evaluated the work by conducting a listening test. The participants were asked to classify the results as ``human composed '', ``algorithm composed '' or ``not sure ''. They were also asked to rate the pleasantness of the melodies on a scale from ``excellent '' over ``good '', ``fair '' and ``poor '' to ``bad ''. His results are visualized in Table \ref{tab:kath-confusion} and Figure \ref{fig:kath-pleasantness}.

\begin{table}[ht]
  \begin{center}
    \begin{tabu}{ l | l | c | c | c |}
       \multicolumn{2}{l}{} & \multicolumn{3}{c}{Predicted}\\
       \hhline{~~---}
       \multicolumn{2}{l|}{} & \specialcell{Human\\Composed} & \specialcell{Algorithm\\Composed} & \specialcell{Not sure} \\
       \hhline{~----}
       \multirow{2}{*}{Original} & Human Composed & 73\,\% &  19\,\% &  8\,\% \\
       \hhline{~----}
                             & Algorithm Composed & 22.5\,\% &  69\,\% &  8.5\,\%\\
       \hhline{~----}                
    \end{tabu}
    \caption[Confusion matrix of Kathiresan's results]{Confusion matrix of the results of Kathiresan's listening test. The rows represent the true class of the melodies while the columns represent the class the participants assigned the the melodies presented.\\
      \begin{footnotesize}
        Table extracted from Kathiresan's work \cite{Kathiresan2015}.
      \end{footnotesize}}
    \label{tab:kath-confusion}
  \end{center}
\end{table}

This thesis bases on the code and the idea but takes it further to overcome some of its main issues:
\begin{itemize}
\item Our main focus lies on compensating the lack of global composition awareness of Markov models as described in Section \ref{subsubsec:gmMarkovChains}.
\item We also try to restructure the composition process so that the length parameter to it can be given in bars instead of number of notes which makes less sense from a musical point of view.
\item Another major point of improvement is that our algorithm is capable of learning from an arbitrary number of input melodies while Kathiresan's was restricted to one.
\item We decide to deprecate the constraint feature in our code as it was implemented in a rather trivial manner with a lot of randomness and less systematic or musical considerations involved, leaving an improved implementation open as future work.
\end{itemize}

\section{Musical Terms}
\label{sec:musical-terms}
\subsection{Key and Mode}
\label{subsec:key-mode}
\begin{figure}[htbp]
  \center
  \begin{subfigure}{.5\textwidth}
  \center
  \resizebox{.85\textwidth}{!}{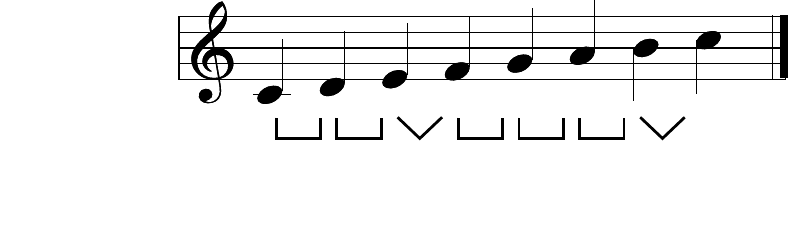}
  \caption{\emph{c major} scale}
  \end{subfigure}%
  \begin{subfigure}{.5\textwidth}
  \center
  \resizebox{.85\textwidth}{!}{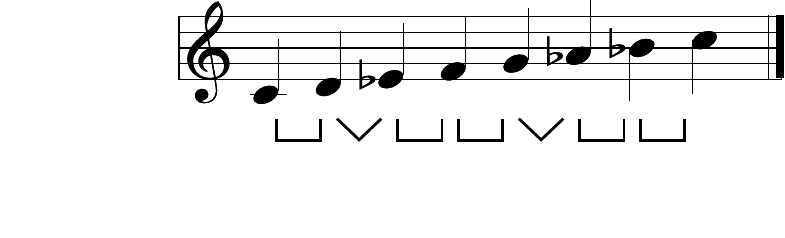}
  \caption{\emph{c minor} scale}
  \end{subfigure}\\%
  \caption[Intervals of major and minor scale]{Scales of \emph{c major} and \emph{c minor} with tonal intervals.}
  \label{fig:scales}
\end{figure}

The \emph{mode} of a musical piece defines a series of \emph{intervals} (i.e. pitch distances) between the notes of a scale (i.e. consecutive pitches in a key---see below). Some examples of modes are \emph{major} and \emph{minor}. The corresponding scales can be found in Figure \ref{fig:scales}.

The key defines a the \emph{root note} (i.e. where the scale starts) and the mode (see above) of a piece.

\subsection{Beats, Bars, and Time Signature}
\label{subsec:beats-bars-time-signature}
\begin{figure}[htbp]
  \center
  \hfil
  \begin{subfigure}{.47\textwidth}
  \center
  \resizebox{.9\textwidth}{!}{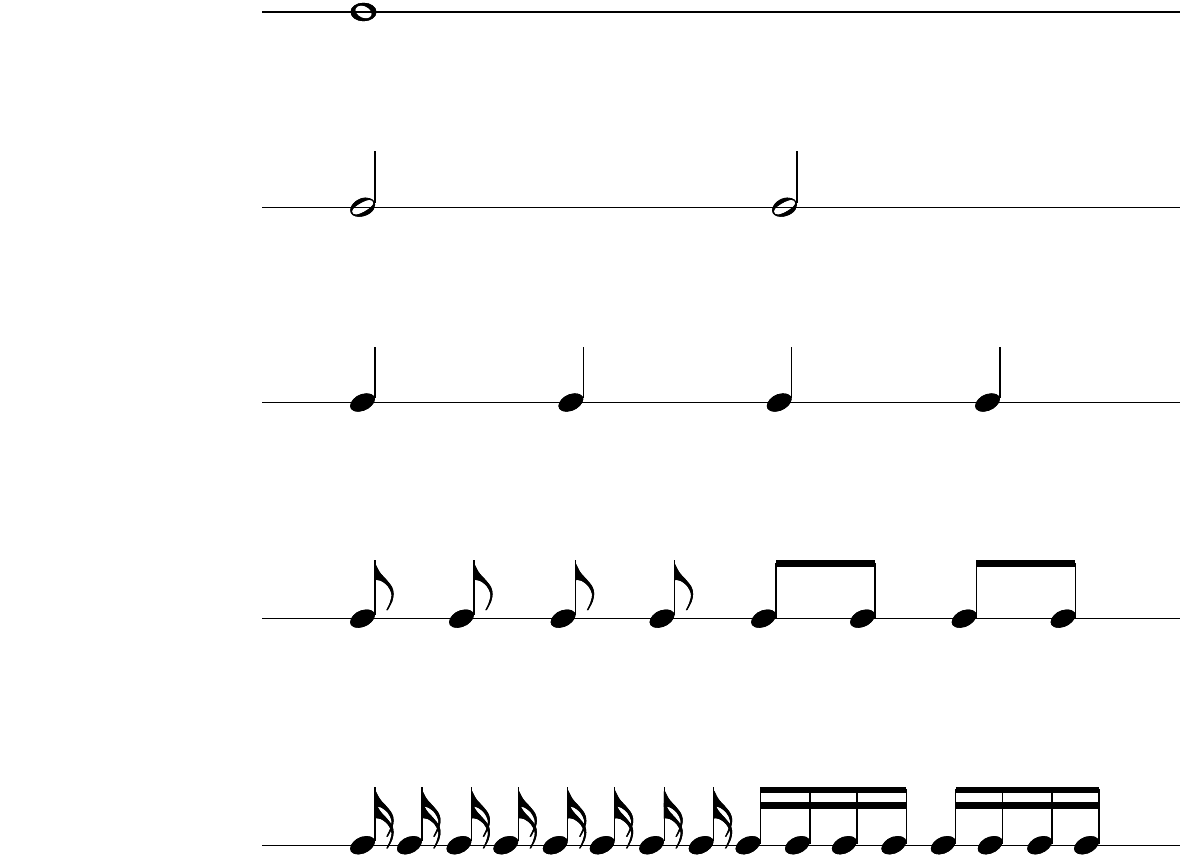}
  \caption{Binary division of a \emph{whole note}. Both representations of 8\textsuperscript{th} and 16\textsuperscript{th} notes are equivalent.}
  \end{subfigure}
  \hfil
  \begin{subfigure}{.47\textwidth}
  \center
  \resizebox{.9\textwidth}{!}{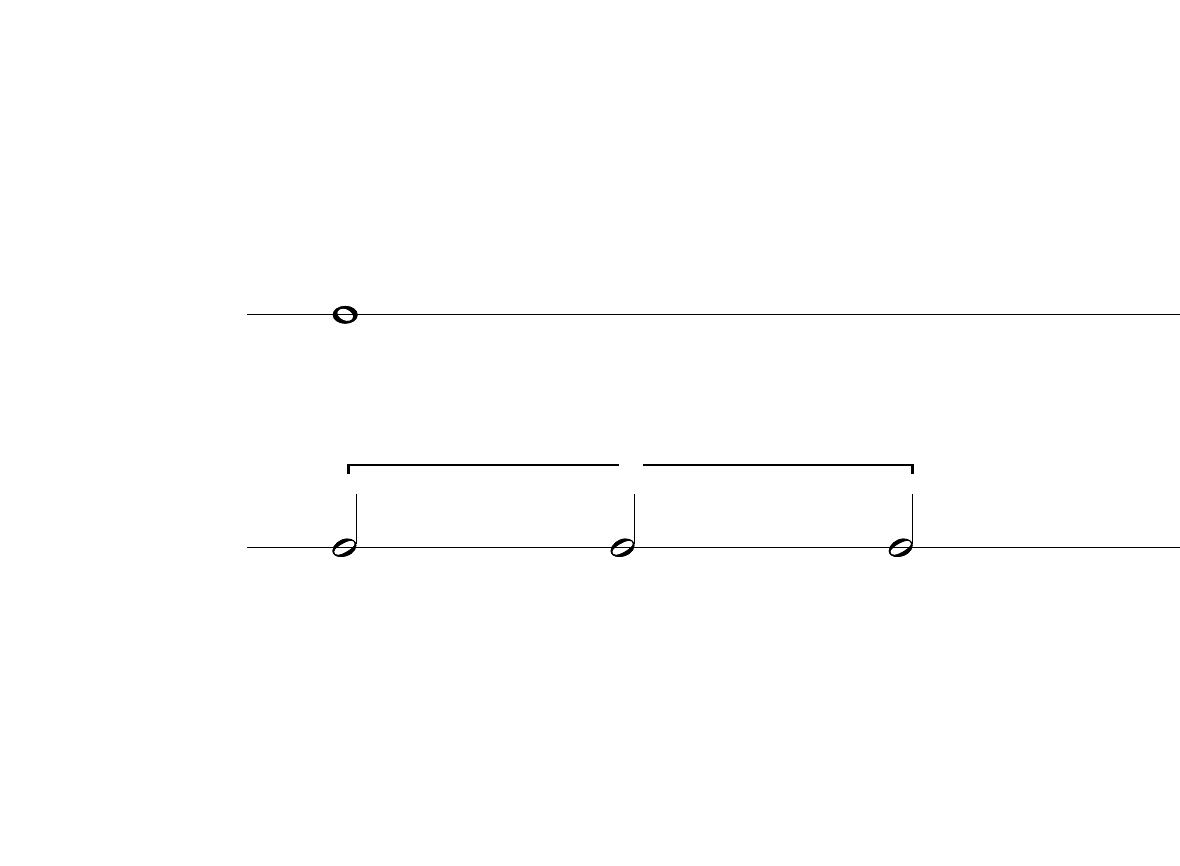}
  \caption{A \emph{whole note} can be divided into three \emph{half triplets}. This applies to other note durations as well.}
  \end{subfigure}%
  \hfil
  \caption[Note durations]{Note durations and their relations.}
  \label{fig:note-durations}
\end{figure}

First, let's have a look at \emph{note durations} (usually referred to as note values). There are \emph{whole notes}\footnote{Longer notes exist, but do not matter in this work.} that can either be divided binary or ternary (using triplets) as shown in Figure \ref{fig:note-durations}. A time signature
(e.g.\,\raisebox{-.3\height}{%% Creator: Inkscape inkscape 0.48.4, www.inkscape.org
%% PDF/EPS/PS + LaTeX output extension by Johan Engelen, 2010
%% Accompanies image file 'music_time_signature.pdf' (pdf, eps, ps)
%%
%% To include the image in your LaTeX document, write
%%   \input{<filename>.pdf_tex}
%%  instead of
%%   \includegraphics{<filename>.pdf}
%% To scale the image, write
%%   \def\svgwidth{<desired width>}
%%   \input{<filename>.pdf_tex}
%%  instead of
%%   \includegraphics[width=<desired width>]{<filename>.pdf}
%%
%% Images with a different path to the parent latex file can
%% be accessed with the `import' package (which may need to be
%% installed) using
%%   \usepackage{import}
%% in the preamble, and then including the image with
%%   \import{<path to file>}{<filename>.pdf_tex}
%% Alternatively, one can specify
%%   \graphicspath{{<path to file>/}}
%% 
%% For more information, please see info/svg-inkscape on CTAN:
%%   http://tug.ctan.org/tex-archive/info/svg-inkscape
%%
\begingroup%
  \makeatletter%
  \providecommand\color[2][]{%
    \errmessage{(Inkscape) Color is used for the text in Inkscape, but the package 'color.sty' is not loaded}%
    \renewcommand\color[2][]{}%
  }%
  \providecommand\transparent[1]{%
    \errmessage{(Inkscape) Transparency is used (non-zero) for the text in Inkscape, but the package 'transparent.sty' is not loaded}%
    \renewcommand\transparent[1]{}%
  }%
  \providecommand\rotatebox[2]{#2}%
  \ifx\svgwidth\undefined%
    \setlength{\unitlength}{15.95276318bp}%
    \ifx\svgscale\undefined%
      \relax%
    \else%
      \setlength{\unitlength}{\unitlength * \real{\svgscale}}%
    \fi%
  \else%
    \setlength{\unitlength}{\svgwidth}%
  \fi%
  \global\let\svgwidth\undefined%
  \global\let\svgscale\undefined%
  \makeatother%
  \begin{picture}(1,1.21084039)%
    \put(0,0){\includegraphics[width=\unitlength]{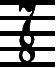}}%
  \end{picture}%
\endgroup%
}\,) is defined by two integers. The lower one (always a power of 2) determines what note duration is considered as a \emph{beat}, while the upper one determines how many beats are in a \emph{bar}. The bar boundaries are indicated by horizontal lines in the staff, which is the system of line(s) the notes are written on.

\subsection{Count and Off-beat}
\label{subsec:count-off-beat}
\begin{figure}[htb]
  \center
  \begin{subfigure}{.9\textwidth}
  \center
  \resizebox{.9\textwidth}{!}{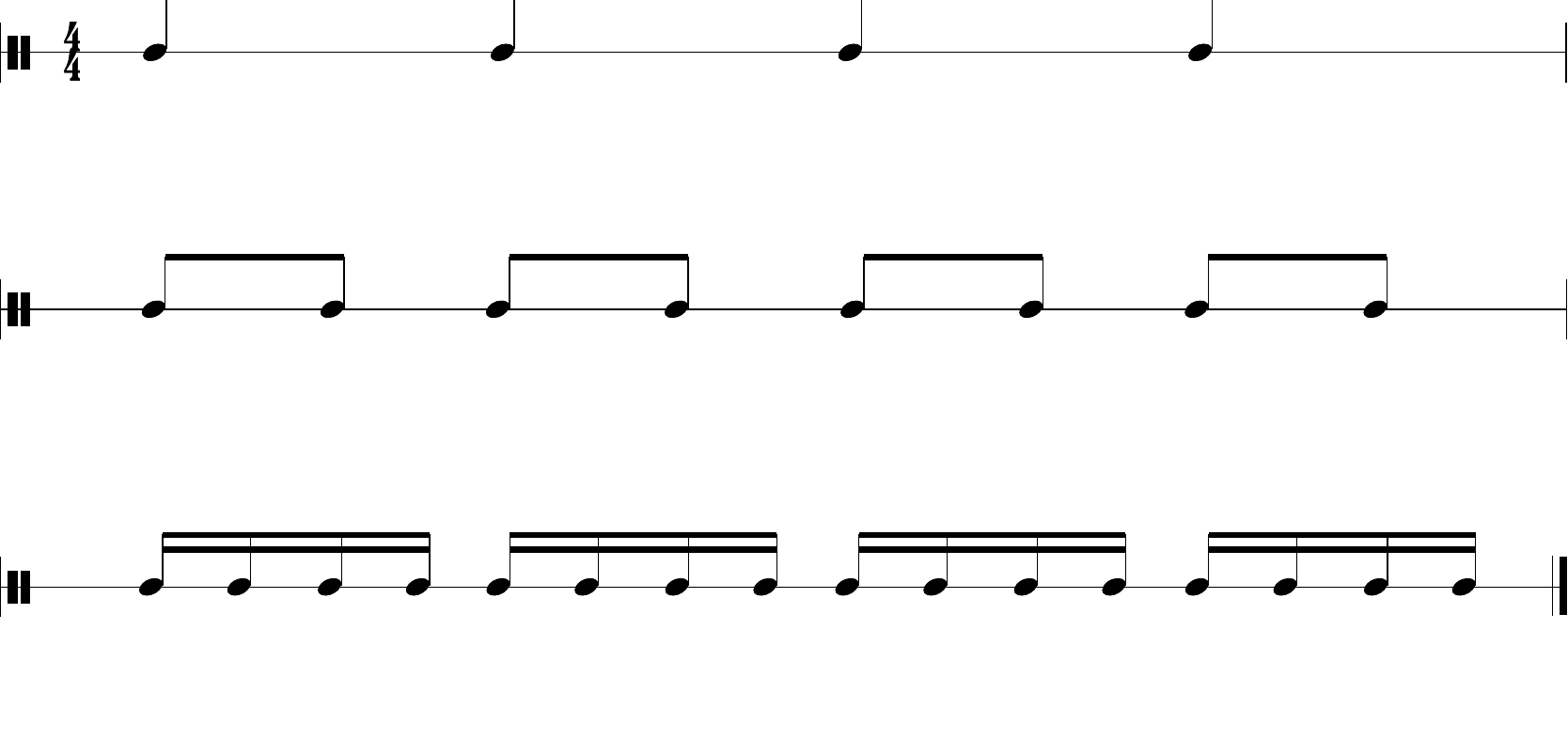}
  \caption{Counting 16\textsuperscript{th} notes with $\frac{4}{4}$ time signature.}
  \label{fig:binary-counting}
  \end{subfigure}
  
  \begin{subfigure}{.9\textwidth}
  \center
  \resizebox{.9\textwidth}{!}{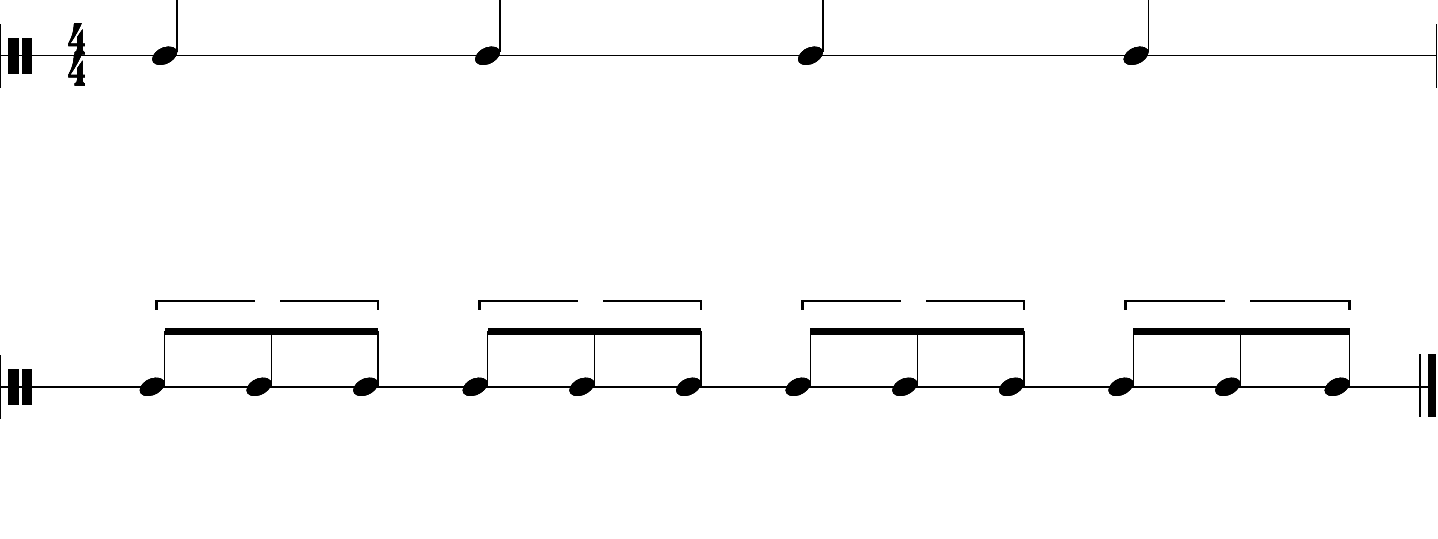}
  \caption{Counting 8\textsuperscript{th} triples with $\frac{4}{4}$ time signature.}
  \label{fig:ternary-counting}
  \end{subfigure}
  \hfil
  \caption[Counting $\frac{4}{4}$ bars]{Counting $\frac{4}{4}$ bars. The first lines of text are the linguistic way of counting, while the second lines are the numerical way.}
  \label{fig:counting}
\end{figure}

In this work, \emph{count} can either refer to a linguistic or a numerical representation of the position in a bar. Both representations share that successive integers, stating by 1, are assigned to the \emph{beats} of each \emph{bar}. While the numerical representation is then created by linear interpolation (see Figure \ref{fig:binary-counting}), the linguistic one is a little more complex. ``and''s are inserted between the beats when the beat duration (see time signature in Section~\ref{subsec:beats-bars-time-signature}) is split in half. If split in half again, ``a''s are inserted between the beats and the ``and''s. If split into 3 (using triples), the new notes are called ``trip'' and ``let'' (see Figure \ref{fig:ternary-counting}).

When linguistically referring to a count a simplification is made for shortness. Instead of calling the 8\textsuperscript{th} 16\textsuperscript{th} note in Figure \ref{fig:binary-counting} (row 3) ``1 a and a 2 a and a'' all syllables before the last beat (``2'') are dropped. Thus, it's called ``2 a and a''. Commonly syllables that aren't necessary for uniqueness are dropped as well, i.e. it's ``2 and a''. More generally, binary counts match the regular expression ``\verb|\d+( and)?( a)?|'', while ternary counts match the regular expression ``\verb|\d+( trip( let)?)?|''.

Though it usually refers to something different, we will use the term \emph{off-beat} throughout the work as follows: The off-beat is the position in the bar, relative to the previous beat. Let $ n $ be the numerical representation of a count. Then the off-beat $ o $ is defined as $ n \bmod 1 $. The count ``4 and a'' (numerically 4.75) has an off-beat of .75.

\section{Proposed Method}
\label{sec:proposed_method}
\begin{figure}[htb]
  \centering
  \def\svgwidth{\textwidth}
  \begin{footnotesize}
    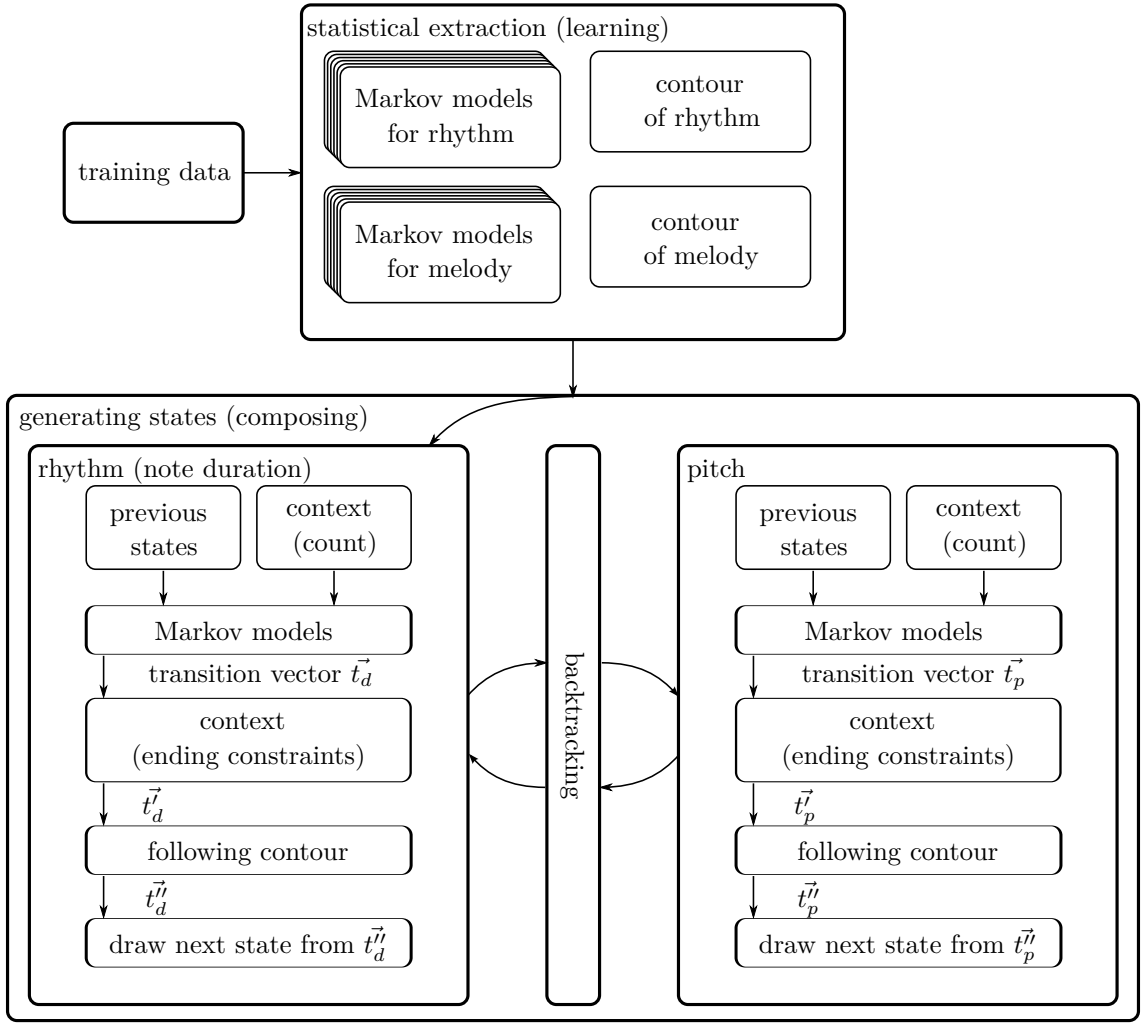
  \end{footnotesize}
  \caption[Workflow of proposed method]{
    Workflow of the proposed method: An overview is given in Section \ref{sec:proposed_method}. For details on the blocks see their respective sections. Training data: \ref{subsubsec:mtc-dataset}, Markov models: \ref{subsubsec:gmMarkovChains} and \ref{subsubsec:markov-models}, Contour learning: \ref{subsubsec:contours}, Ending constraints: \ref{subsubsec:ending-constraints}, Following Contours and drawing next state: \ref{subsubsec:following-contours}, Backtracking: \ref{subsubsec:backtracking}.
    %As training data we use the MTC-Dataset as explained in Section \ref{subsubsec:mtc-dataset}. The Markov models are explained in Section \ref{subsubsec:gmMarkovChains} and \ref{subsubsec:markov-models}. More information about the contours can be found in Section \ref{subsubsec:contours}. In Section \ref{subsubsec:ending-constraints} we'll explain the ending constraints. Section \ref{subsubsec:following-contours} explains how we are following the learned contours and draw the following state.
  }
  \label{fig:processing-pipeline}
\end{figure}

In the following, we will give an overview of the proposed method. The source code can be found on GitHub (\url{https://github.com/roba91/melody-composer}). We will explain the process more in detail in the next subsections. The proposed method is a hybrid approach combining Markov models (see Section \ref{subsubsec:gmMarkovChains}) with a further statistical extraction method aiming for compensating the lack of global context awareness of Markov models. Also, further modifications are made to give more context to the melody generation process. Our processing pipeline is shown in Figure \ref{fig:processing-pipeline}.

The pipeline starts with the training dataset. In a pre-processing step, the melodies of the MTC-Dataset (see Section \ref{subsubsec:mtc-dataset}) are filtered (keeping only melodies in \emph{major} mode with a $\frac{4}{4}$ time signature) and transposed\footnote{Transposing a melody technically means to increase or decrease the pitch of all notes by a constant (e.g. one semi-tone).} to the key of \emph{C major}. This simplification was made to avoid having to handle different keys or modes and to increase the number of training melodies per key as all melodies are in \emph{C major}. After that pre-processing the actual algorithm starts its work. It first reads in the melodies and splits them into phrases as indicated by phrase separator marks that came with the MTC-Dataset (see Section \ref{subsubsec:mtc-dataset}). The algorithm then extracts two statistical characteristics of rhythm and pitches for each melody:
\begin{itemize}
\item For note pitches and note durations each, a parametric Markov model of higher order is trained. Instead of using a transition matrix modeling a conventional Markov model we use a transition tensor. The added axis describes the current off-beat. It's defined within $[0,\,1)$. So for a note being played at count ``\emph{3 and a}'' the value would be $ 3.75 \bmod 1 = 0.75 $.

The lower the order of a Markov model is, the less context sensitive will the results be. However, if the order is very high the results will be very close to the training data and the memory consumption increases exponentially with the order (see Section \ref{subsubsec:markov-models}). The available RAM\footnote{Random access memory: Temporal memory of a computer used for calculations.} determined our upper limit to Markov models of fourth order while a lower order makes less sense from a musical perspective as four notes often form a small musical pattern.

\item For both pitches and note durations we extract a contour (see Section \ref{subsubsec:contours}) for each phrase, which is basically the low-passed and x-normalized function of the feature (duration or pitch) over time. The contours are then clustered and one of the clusters is selected. A mean of all contours in it is calculated and used as contour to follow while generating melodies.
\end{itemize}
After that learning process, phrases can be generated. In contrast to the algorithm proposed by Kathiresan (see Section \ref{subsubsec:master-thesis}), where all pitches are generated first and the rhythm is generated afterwards, we generate a note duration, then the pitch for it and repeat that process until we are done. If the process reaches a point where the training data is insufficient (i.e. the combination of previous states couldn't be observed in the training data---at least not for the current count) we use backtracking to go back and truncate the search tree (see Section \ref{subsubsec:backtracking}).

Now that a basic overview of the algorithm is given we will explain the details of it.

\subsection{Learning}
\label{subsec:learning}
\subsubsection{MTC-Dataset}
\label{subsubsec:mtc-dataset}
Before learning anything, a dataset to learn from is needed. Hence, we will give an introduction to the MTC-Dataset that is used by the proposed algorithm.

As will explained in Section \ref{subsubsec:contours} we develop a method to extract melody and rhythm contours from our training data. Treating each song as a whole yields the problem that contours become very diverse and don't form nice clusters. However, considering phrase-wise contours works much better. Thus, we need a phrase annotated dataset. The \emph{Meertens Tune Collections} (MTC) \cite{Kranenburg2014} consists of several datasets of music pieces. The datasets vary in their representations and the meta data provided. The MTC-FS dataset is one of them. It consists of \emph{digitally encoded} (many formats like MIDI\footnote{Musical Instrument Digital Interface: Its a multi-track (polyphinic) file format that represents musical events on a time scale. E.g. when and how long which note is to be played.\\See \url{https://www.midi.org/}.}, LilyPond\footnote{A file format that encodes sheet music in ASCII. See \url{https://www.lilypond.org/}.}, and PDF\footnote{Portable Document Format: Its a cross-platform file format to encode documents.\\See \url{https://www.adobe.com/devnet/pdf/}.} are available) dutch vocal folk song \emph{melodies} that are manually \emph{phrase annotated}\footnote{The phrase annotations are encoded as text events on a separate track in the MIDI files.}. Thereby, it fulfills all our requirements.

Handling a dataset with all possible keys, modes and time signatures of melodies mixed together would clearly go beyond the temporal constraints of this work. Therefore, a pre-processing step that filters and normalizes the melodies to one key (\emph{C major}) and one time signature ($\frac{4}{4}$) is introduced. It also removes polyphonic\footnote{Monophony means that at most one note is played at the same time. Polyphony makes no restrictions to number of simultaneous notes.} input as our definition of melodies includes monophony.

The pre-processing tags the MIDI files with the difference of their key to \emph{C major}\footnote{The information about the key the melodies are in is taken from the LilyPond files that came with the MTC-Dataset. LilyPond files are designed to represent sheet music and therefore contain the key information. MIDI files in contrast, don't include information about the key, and guessing the key of a musical piece is error prone as it can be ambiguous.}. A melody in \emph{D major} e.g. would be tagged with the infix ``\emph{\_m2}'' (for minus two semitones) between the file name and the file ending. This infix is considered by our MIDI module when reading the files. It then transposes the melodies according to the file name infix.

The whole dataset has 4,120 songs with a total of 23,936 phrases (on average 5.81 phrases per song, average phrase length of 1.82 $\frac{4}{4}$-bars\footnote{Evaluated on only 4,084 songs because 36 weren't usable as they were polyphonic, which our MIDI module isn't capable of handling.}) From those melodies we only use the ones in a \emph{major} key with a $\frac{4}{4}$ time signature (find the distributions in Tables \ref{tab:mode-distribution} and \ref{tab:signature-distribution}). After filtering the data, there were 378 songs with a total of 2147 phrases (on average 5.68 phrases per song, average phrase length of 1.92 $\frac{4}{4}$-bars) left. % a total of 1303 songs would have been possible -.- damn bug!

%It first reads in the MIDI files, transposes them to \emph{C major} as indicated by the infix of the pre-processing and finally splits them into phrases.

\begin{table}[htb]
  \begin{center}
    \begin{tabular}{|*{6}{c|}}
      \hline
      major & minor & dorian & phrygian & lydian & mixolydian\\ \hline
      3,879 &   161 &     69 &        4 &      2 &          5\\ \hline
    \end{tabular}
    \caption[Mode distribution of MTC-FS]{The mode distribution among the songs in the MTC-FS dataset.}
    \label{tab:mode-distribution}
  \end{center}    
\end{table}

\begin{table}[htb]
  \begin{center}
    \begin{tabu} to 0.9 \textwidth {|c| *{10}{|X[c]}|}
      \hhline{-||----------|}
      \backslashbox{k}{n} &
             1 &    2 &    3 &    4 &    5 &    6 &    7 &    8 &    9 &   12 \\                  
                                 \hhline{=::==========|}
      2 &   -- &  191 &   44 &   12 &   -- &   -- &   -- &   -- &   -- &   -- \\
                                 \hhline{-||----------|}
      4 &    6 &  382 &  593 & 1,379&   10 &   94 &    1 &    3 &    2 &   -- \\
                                 \hhline{-||----------|}
      8 &   -- &    1 &   66 &   13 &   12 & 1,222&    1 &    1 &   71 &   16 \\
                                 \hhline{-||----------|}
    \end{tabu}
    \caption[Time signature distribution of MTC-FS]{The time signature ($\frac{n}{k}$) distribution among the melodies in the MTC-FS dataset.}
    \label{tab:signature-distribution}
  \end{center}
\end{table}

\subsubsection{Parametric Markov models}
\label{subsubsec:markov-models}
From a musical perspective it is intuitive that the probabilities of the note durations are different depending on their position in a musical phrase. Examples are phrase ending rhythms and even more prominent the fact that it's very unlikely (at least for folk songs) that after two 8\textsuperscript{th} triples (on count ``\emph{1 trip let}'') something else follows than another 8\textsuperscript{8} triplet or similar (e.g. two 16\textsuperscript{th} triplets) because this would cause an offset by one 8\textsuperscript{th} triplet that wouldn't be resolved until the next 8\textsuperscript{th} triplet or similar. We will now present an approach to make Markov models aware of that fact.

Parametric Markov models can be thought of as several Markov models. Instead of a transition matrix a transition tensor of $ \rank 3 $ is implemented. The first idea could be to use the added third dimension for the current position in a bar (i.e. the count) while composing. This causes the training process to need much more training data as transitions are separately observed for all counts in this case. From a musical perspective it can be argued that rhythm considerations are similar for same off-beats regardless of the count (one exception might be the count ``\emph{1}''). Thus, a reasonable simplification to this approach is to reduce the count to its off-beat. So, for a note being played at count ``\emph{3 and a}'', the value would be $3.75 \bmod 1 = 0.75$. In other words, the third dimension does not represent the count (time relative to current bar), but the off-beat (time relative to current beat).

To get an idea of the memory consumption of our algorithm which is growing exponentially with the order chosen for the Markov model, here is an example. Let's assume we are looking at a parametric Markov model of fourth order for pitches and we have a total of 29 different pitches in our training set (which is the case for the MTC-FS dataset after filtering as described in Section \ref{subsubsec:mtc-dataset}). This would result in $ \sum\limits_{i=1}^4 29^i = 732,540 $ rows each holding 29 probabilities (floats take up 32 bits = 4 bytes). So it uses $ \sum\limits_{i=2}^5 29^i \cdot 4\,\text{B} = 84,974,640\,\text{B} \approx 85.0\,\text{MB} $ per count. Assuming the training data contains a 32\textsuperscript{nd} as shortest note duration it means we have 8 of those transition matrices in our tensor. The storage needed is then $ 84,974,640\,\text{B}\,\cdot\,8 = 679,797,120\,\text{B} \approx 679.8\,\text{MB} $. Repeating the same approximation with a Markov model of fifth order results in a memory usage of about $ 19.7\,\text{GB} $ which isn't feasible any more for our computational infrastructure.

\subsubsection{Contours}
\label{subsubsec:contours}
Markov models are known to not take a global context into account when used as a generative statistical model. To compensate for that, we will now introduce a contour concept. A contour can be thought of as the rough course of the melody (note pitches) or the rhythm (note durations) over time. The process to learn those contours for melody is very similar to the one for rhythm. Hence, we will focus on the melody contour in the following. The rhythm contour can be learned analogously.

Several approaches to melodic contours have been proposed. One of the earliest was described by Huron \cite{Huron1996}. In his work he only considers the relations between the first, last, and the mean pitch of a phrase. Thus, it is only capable to identify nine types of contours. It is not possible to use the model in a generative way, as it does not make use of absolute values, but ternary relations (higher, lower, same) of consecutive notes, while their duration (i.e. rhythm) is ignored. Schmuckler \cite{Schmuckler1999} proposed a more suitable description of melodic contour. He uses the Fourier transform, which is an advantage in our use case, due to easier implementation of smoothing and comparison of contours, compared to Müllensiefen and Wiggens \cite{Mullensiefen2011} who describe a polynomial approximation.

\begin{figure}[htbp]
  \center
  \begin{subfigure}{.5\textwidth}
  \center
  \resizebox{.85\textwidth}{!}{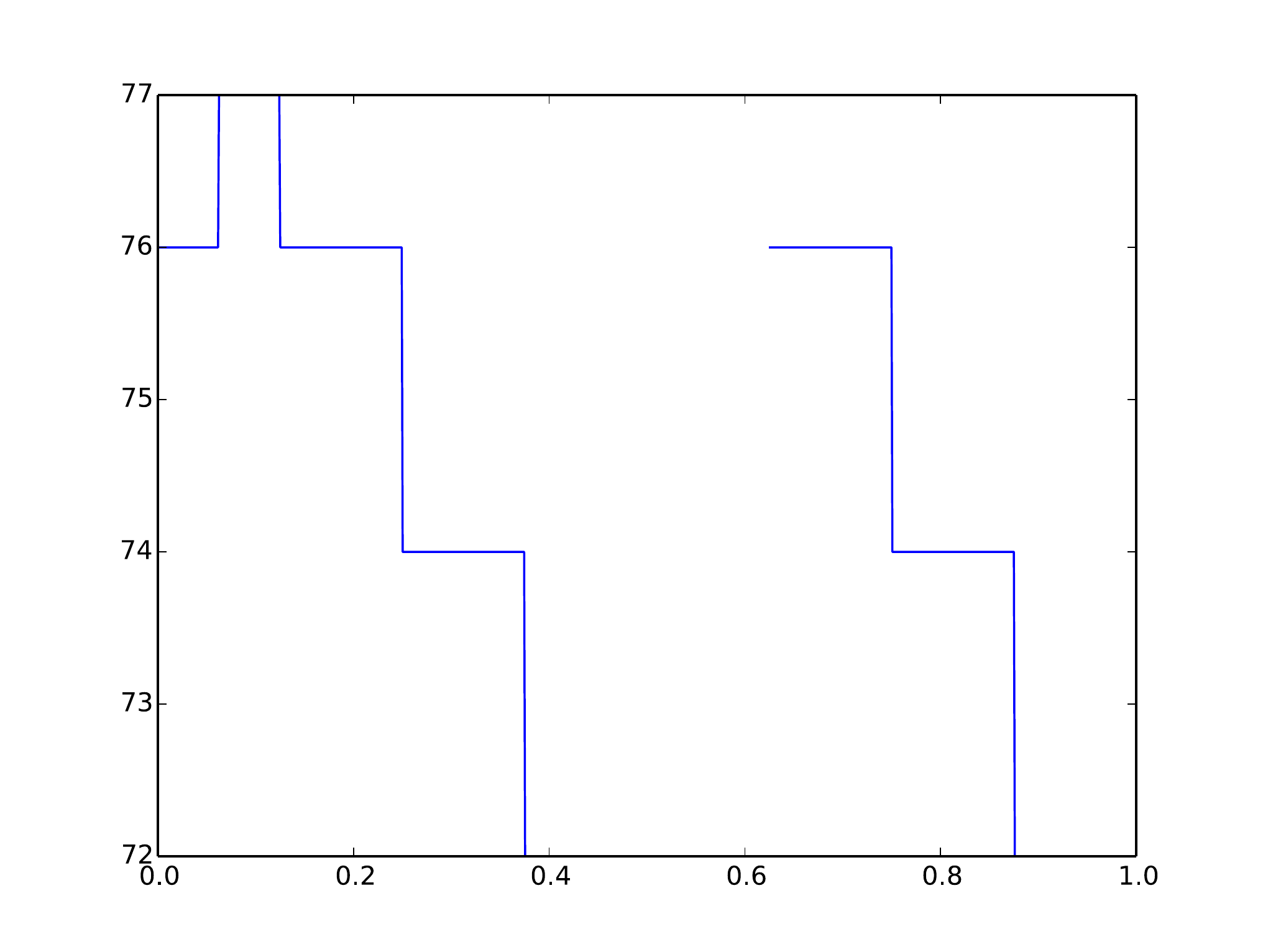}
  \caption{A phrase's melody.}
  \end{subfigure}%
  \begin{subfigure}{.5\textwidth}
  \center
  \resizebox{.85\textwidth}{!}{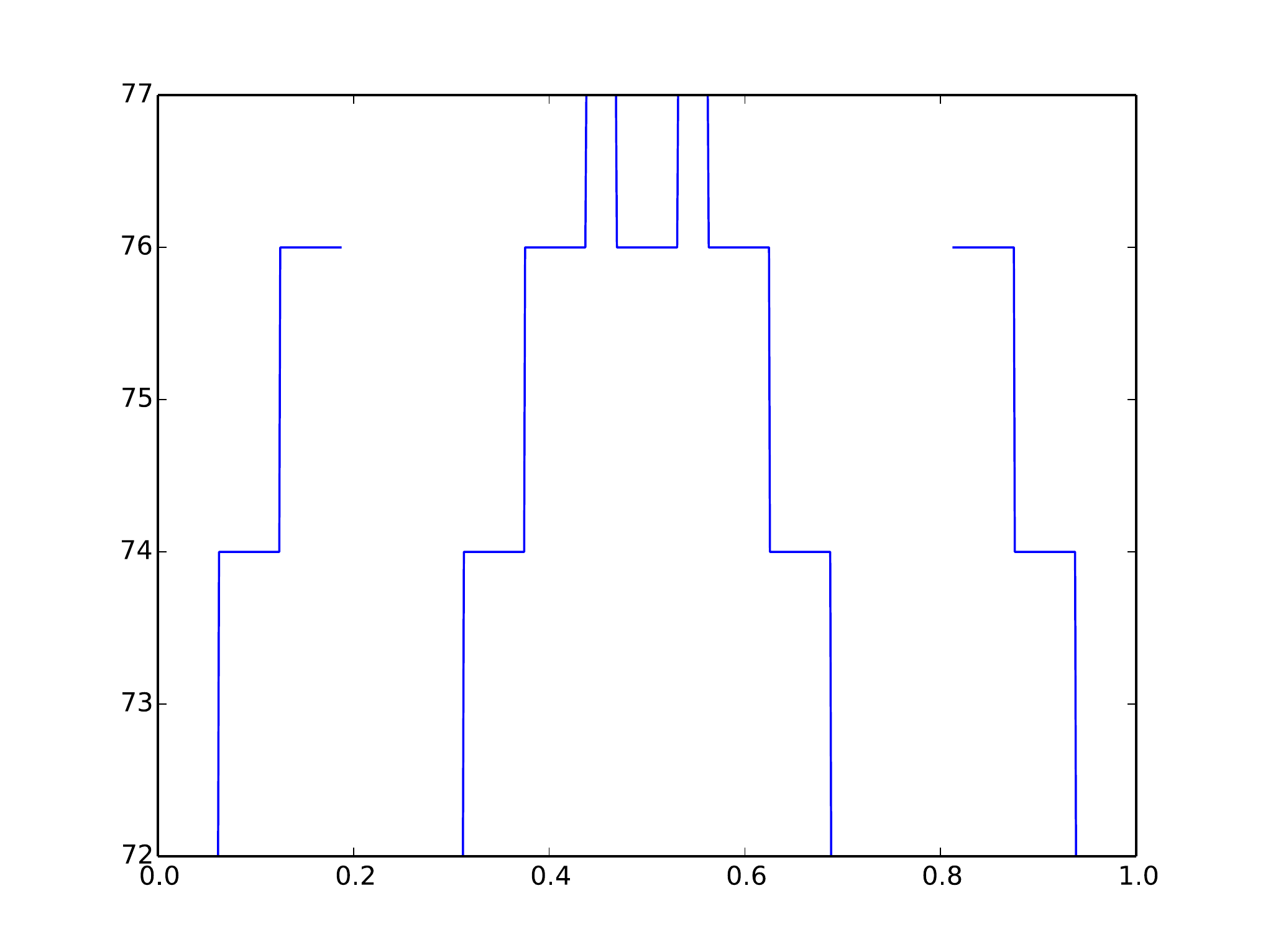}
  \caption{Mirroring the phrase (Step \ref{clustering:mirror}).}
  \end{subfigure}\\%
  \begin{subfigure}{.5\textwidth}
  \center
  \resizebox{.85\textwidth}{!}{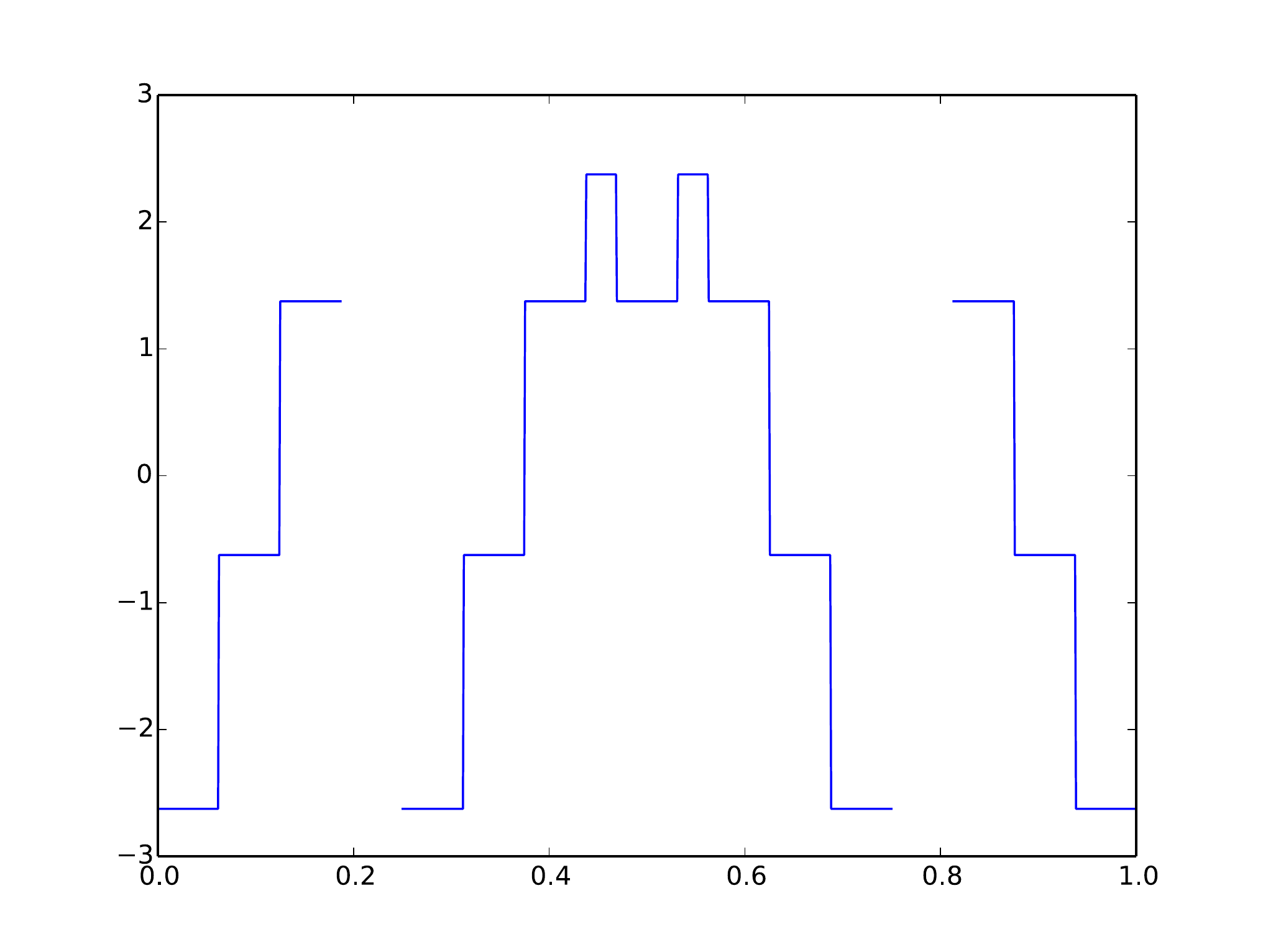}
  \caption{Transposing the phrase (Step \ref{clustering:transpose}).}
  \end{subfigure}%
  \begin{subfigure}{.5\textwidth}
  \center
  \resizebox{.85\textwidth}{!}{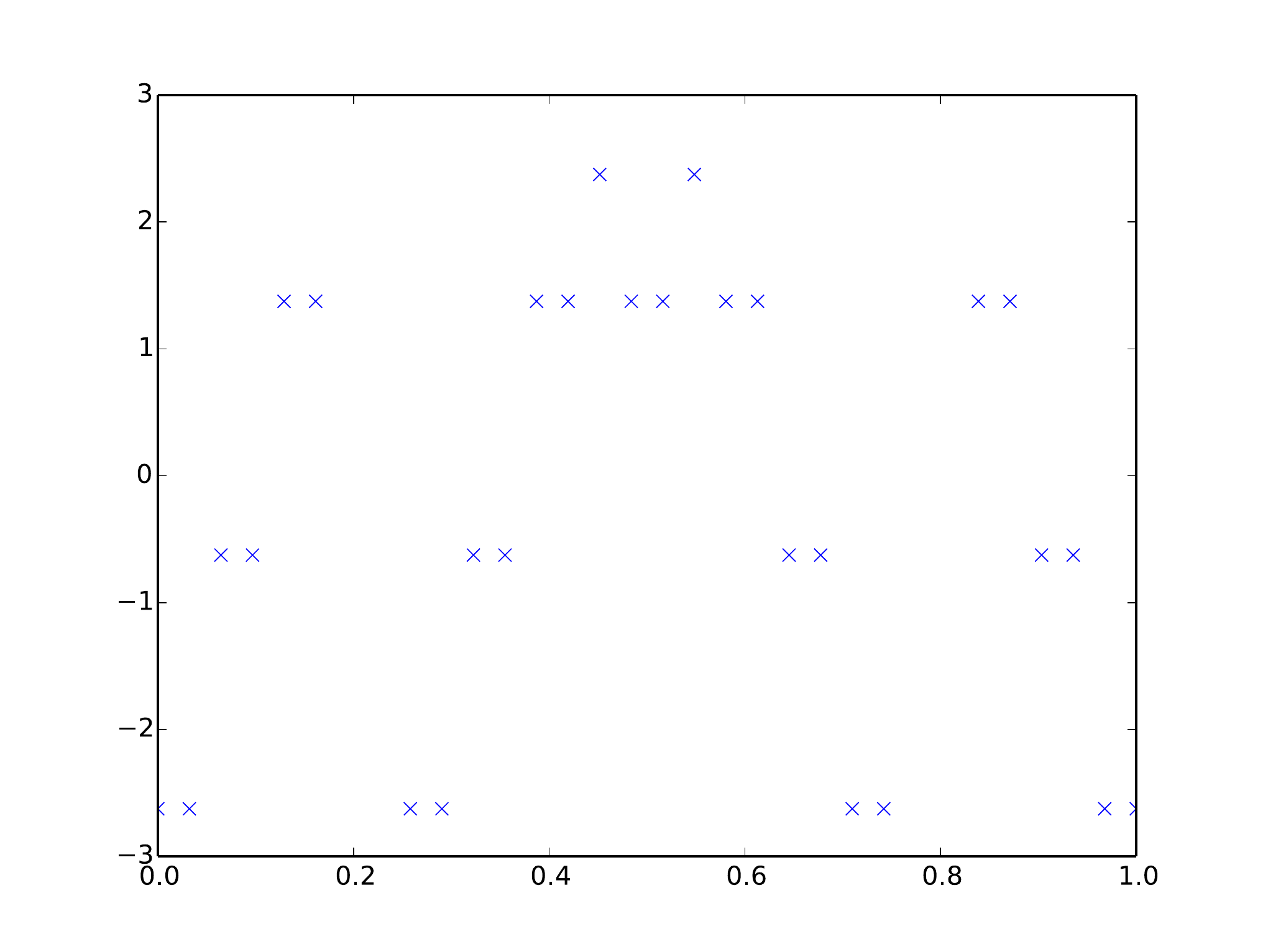}
  \caption{Time discretize the phrase (Step \ref{clustering:sample}).}
  \end{subfigure}\\%
  \begin{subfigure}{.5\textwidth}
  \center
  \resizebox{.85\textwidth}{!}{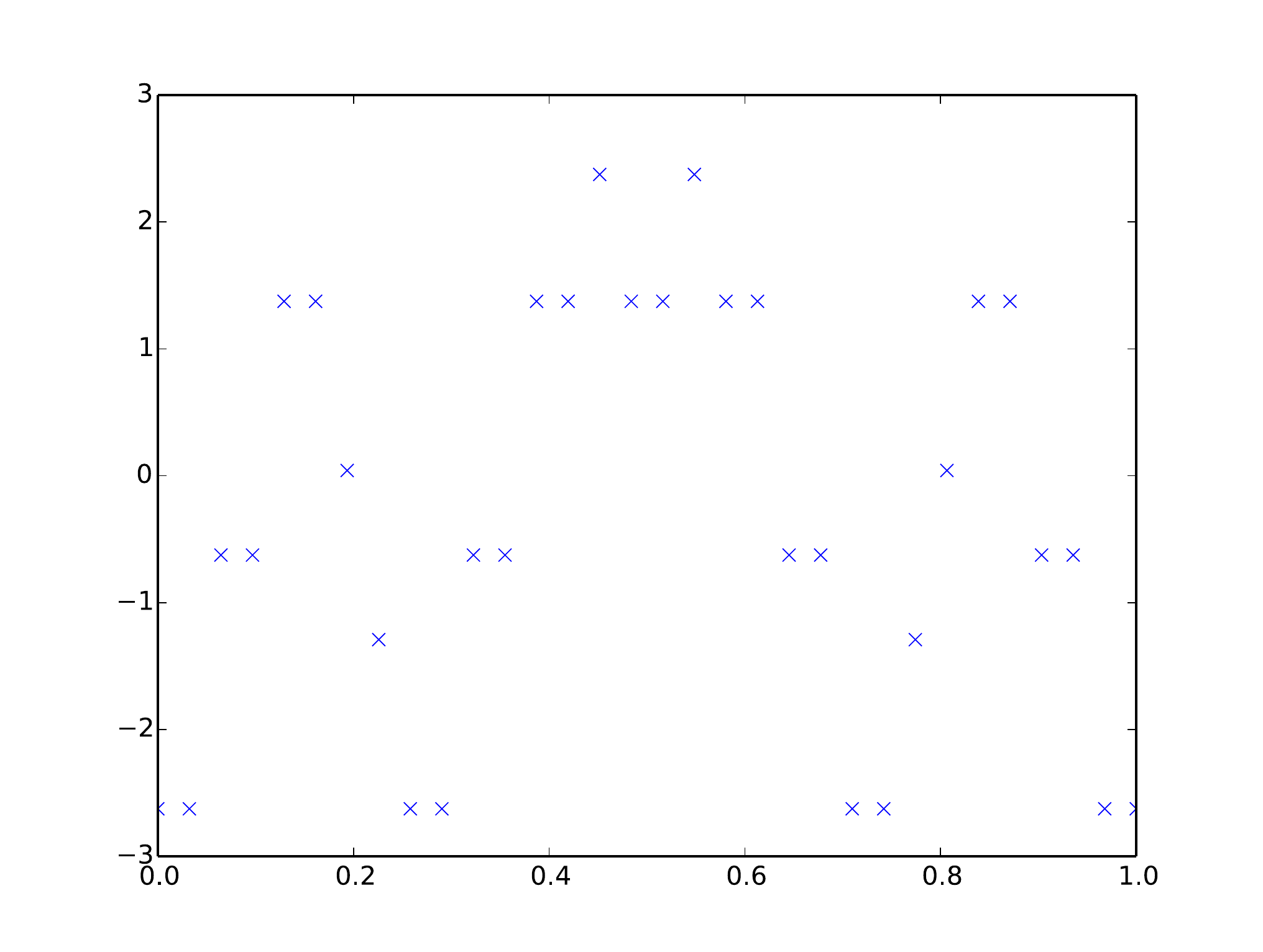}
  \caption{Interpolate the rests (Step \ref{clustering:interpolate-rests}).}
  \end{subfigure}%
  \begin{subfigure}{.5\textwidth}
  \center
  \resizebox{.85\textwidth}{!}{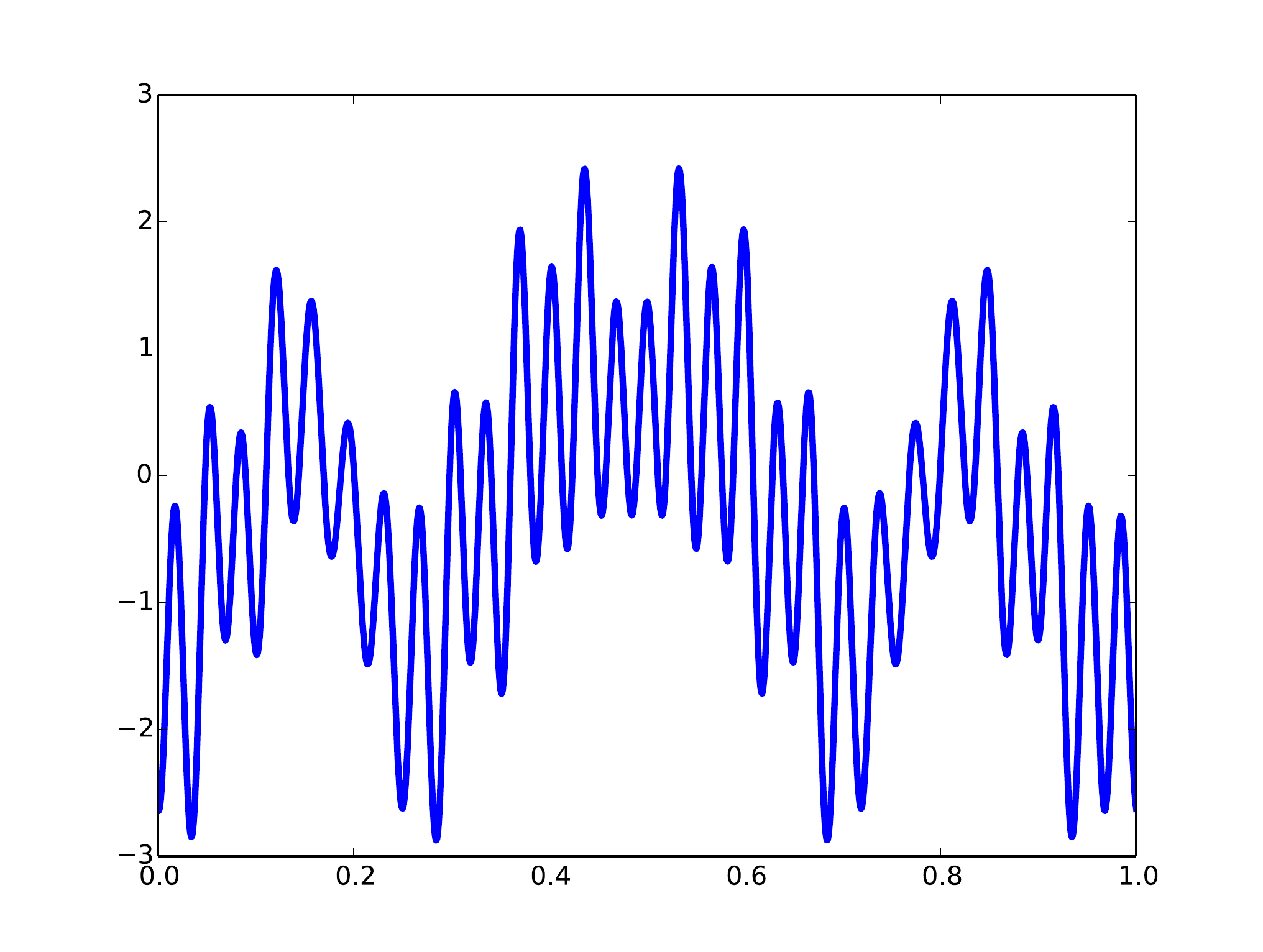}
  \caption{Fourier transform the phrase$^\dagger$ (Step \ref{clustering:fft})}
  \end{subfigure}\\%
  \begin{subfigure}{.5\textwidth}
  \center
  \resizebox{.85\textwidth}{!}{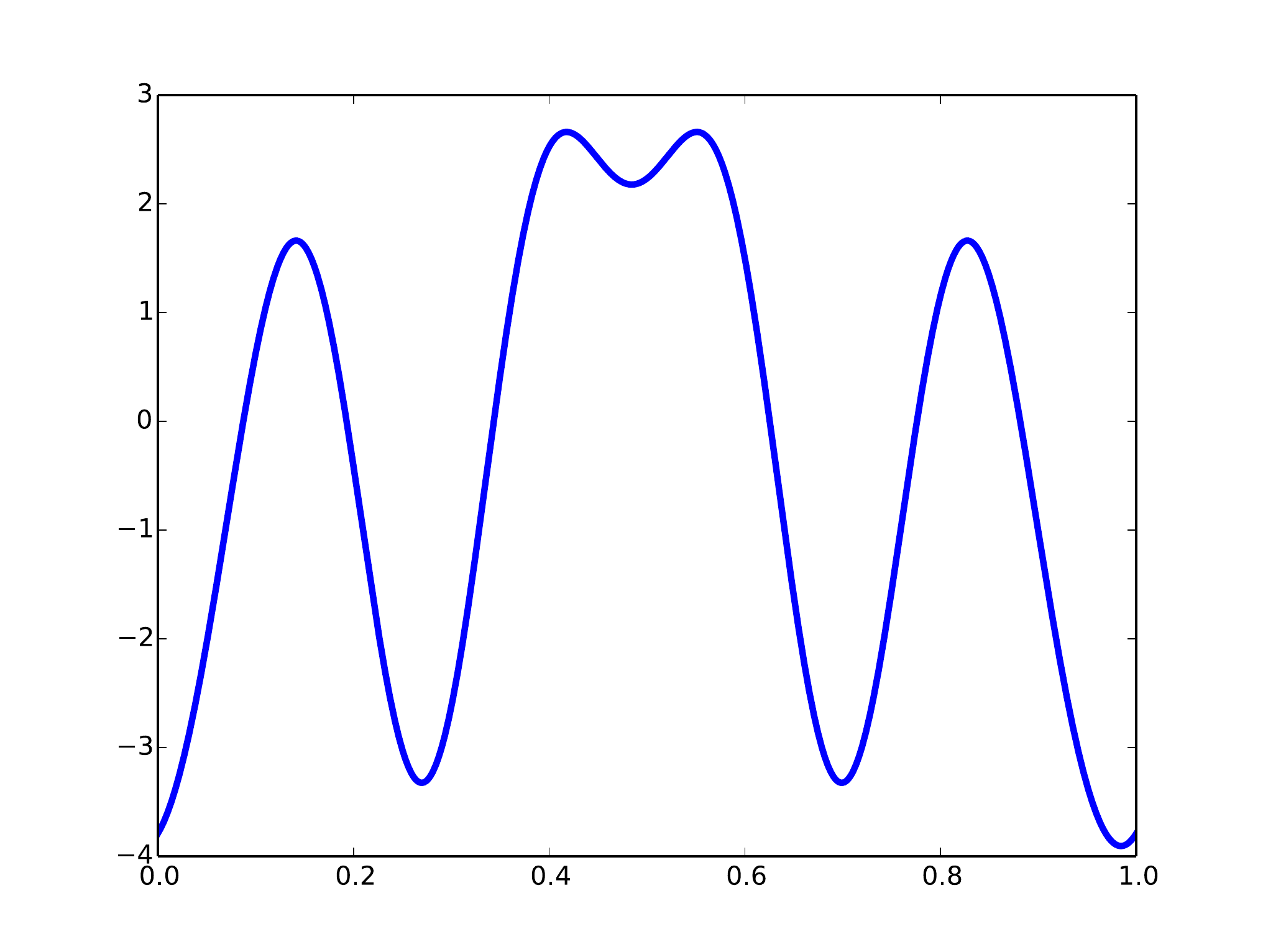}
  \caption{Apply low pass filter$^\dagger$ (Step \ref{clustering:fft}).}
  \end{subfigure}%
  \begin{subfigure}{.5\textwidth}
  \center
  \resizebox{.85\textwidth}{!}{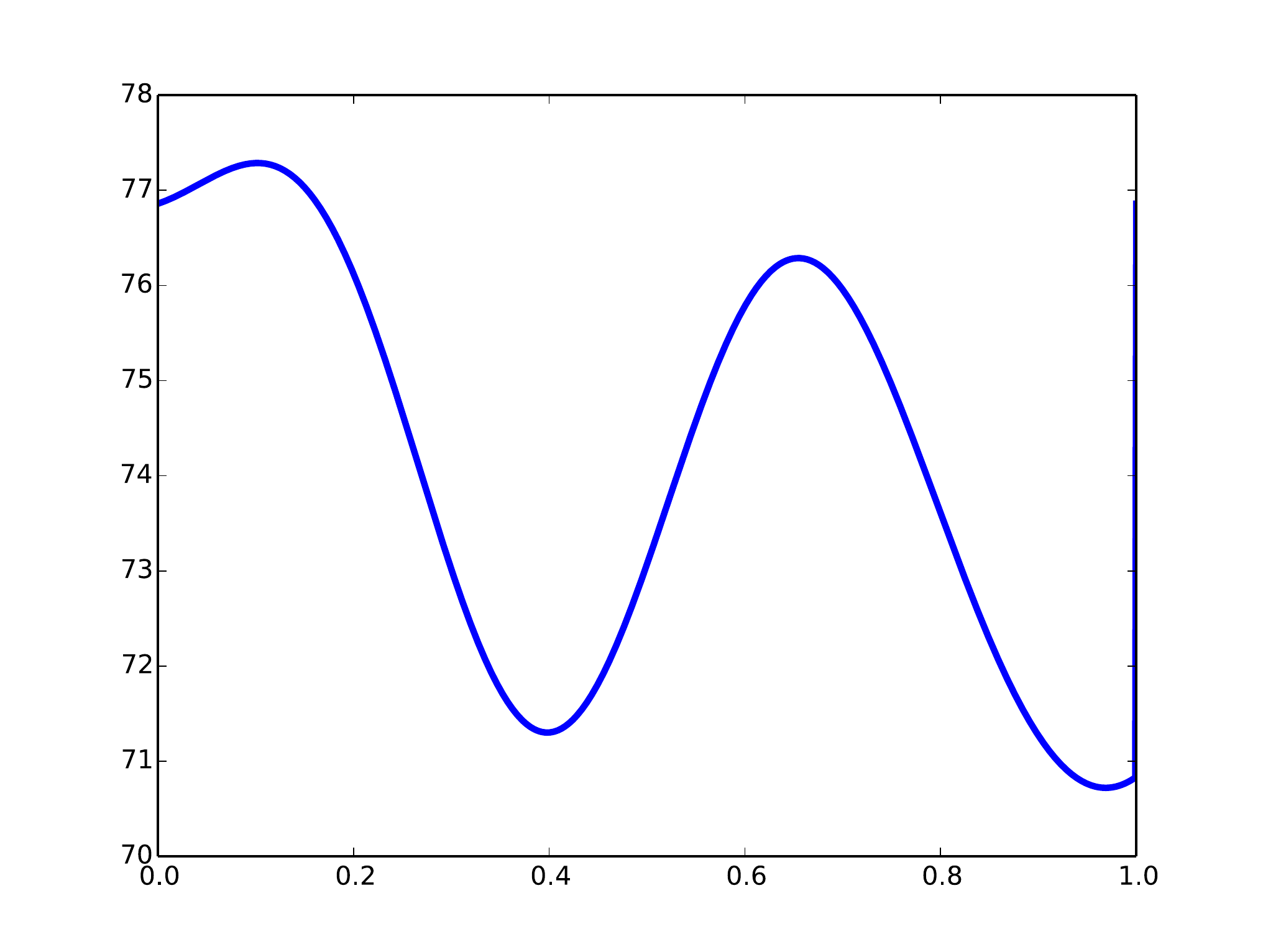}
  \caption{Unmirror the result (Step \ref{clustering:unmirror}).}
  \end{subfigure}%
\caption[Workflow of contour learning]{A minimal example of learning a melody contour. For simplicity we only use one melody. Because of this, the whole clustering part (steps 7--9) is skipped.\\
\footnotesize{$^\dagger$While these steps are actually performed in the frequency domain (after applying the $ \fft $) we chose to plot them in the time domain (applying $ \fft^{-1} $) for better illustration here.}}
\label{fig:contour_workflow}
%\footnotetext{\label{footnote:timedomain}This step is done on the frequency domain (after applying the $ FFT $). But for better illustration we plotted the time domain (applying $ FFT^{-1}) $)}
\end{figure}

An illustration of the major steps of our method can be found in Figure~\ref{fig:contour_workflow}. For each phrase we extracted a contour-feature vector like this:
\begin{enumerate}
\item\label{clustering:mirror}Mirror the phrase. In other words: Prepend the reversed phrase to the original phrase. This avoids a possible wide jump from the last to the first note as the signal is assumed to be repetitive when calculating a discrete Fourier Transform.
\item Transform the phrase into a step function $ m\colon  [0,\,1) \mapsto \{0,\,\ldots,\,127\} $. Mapping a point of time to a MIDI pitch. (For the rhythm the function was defined as $ r\colon [0,\,1) \mapsto \mathbb{R}_{>0} $ where $ 1 $ was a quarter note, $ 0.5 $ a quaver and so on.)
\item\label{clustering:transpose} Center the phrase to be around 0. This step helps to learn contours, regardless of their register. This step is omitted for rhythm. Several methods for transposing were tested: Transposing so that the first note (that isn't a rest) is 0, transposing so that the mean of the melody is 0 and transposing by the mean of all notes (ignoring their duration). The latter method lead to the best clustering behavior for our dataset as indicated by first experiments. A final evaluation of the results is left as future work.
\item\label{clustering:sample} Sample the function in the time axis. 
\item\label{clustering:interpolate-rests} Interpolate rests. Rests don't have a pitch but we needed $ m $ to be a function ($ \forall t \in [0,\,1) \colon \exists m(t) $) for the next steps. This interpolation step is only necessary for the melody, since rests define a rhythm as well. For rests at the beginning of the phrase the first note with a ,pitch is expanded to the beginning of the phrase. For those at the end the last note is expanded. For rests between pitches the interpolation is done linearly.
\item\label{clustering:fft} Apply the Fast Fourier Transform ($\fft$) to the function. Then set all frequencies above a threshold to zero. First experiments indicate that keeping the lowest 6 frequencies is a good choice. An evaluation of the impact of different choices is left for future work (see Section \ref{sec:conclusion}). This acts as a low-pass filter (smoothing of the curve). It is important to beware of the energy-loss that such a operation causes. So we multiplied the result with the ration of energy lost to not lose tonal or rhythmic range (i.e. the width of the function when transformed back into the time domain).  For implementation details see \emph{MelodyContour.low\_pass} in \emph{contour.py}.
\newcounter{contourCounter}
\setcounter{contourCounter}{\theenumi}
\end{enumerate}

\noindent These steps result in a feature vector $ \vec{x} \in \mathbb{C}^6 $ for each phrase. Unsupervised learning techniques can then be applied to find a representative contour for composition:

\begin{enumerate}
\setcounter{enumi}{\value{contourCounter}}
\item\label{clustering:cluster} Clustering of the vectors. You can think of the clusters as families of contours. Using the Ward variance minimization algorithm\cite{Ward1963} yields the best looking results in our use-case (see Figure \ref{fig:contour_clusters}).
Ward's algorithm is a hierarchical clustering algorithm that, in each step, joins the cluster pair that causes the least increase of in-cluster variance. Different algorithms like K-Means and hierarchical clustering with all distance measures provided by SciPy\footnote{For more detail see the SciPy-Documentation: \url{http://docs.scipy.org/doc/scipy/reference/generated/%
scipy.cluster.hierarchy.linkage.html\#scipy.cluster.hierarchy.linkage}} including min-, max-, and average-distance result in more diverse clusters. For this work a number of up to 17 clusters turned out to be a good choice as indicated by small experiments. However, in future work a more refined method of choosing the clustering threshold depending on the presented training data is possible (see Section \ref{sec:conclusion}).
\item\label{clustering:untranspose} Replace the cluster members with their not transposed matches (see Step \ref{clustering:transpose}).
\item\label{clustering:choose-cluster} After clustering one of the resulting clusters is chosen. A first trivial way to do so is to select the largest cluster. This however, often results in a cluster with a mean curve (see Step \ref{clustering:unmirror}) that has little variance. A more promising approach is to use a quality measure incorporating size and variance of the mean curve to choose a cluster. For each cluster $ C_i $ let\\
$\begin{aligned}
r_{C_i}(t) &= \fft^{-1}\left(\frac{1}{\abs{C_i}} \sum\limits_{\vec{x_i} \in C_i}\vec{x_i}\right)& &\text{be the mean contour,}\\
s(C_i) &= \frac{\norm{C_i}}{\max\limits_j \abs{C_j}}& &\text{the normalized size,}\\
mean(r_{C_i}) &= \int\limits_0^1 r_{C_i}(t)\,dt& &\text{the mean of}\ r_{C_i}\  \text{and}\\
w(C_i) &= \int\limits_0^1 \abs{mean(r_{C_i}) - r_{C_i}(t)} dt& &\text{the width of the cluster.}
\end{aligned}$\\
The quality measure is then defined as $ q(C_i) = \frac{s(C_i)}{\max\limits_j s(C_j)} + \frac{w(C_i)^\frac{1}{\gamma}}{\max\limits_k w(C_k)} $ where $ \gamma $ is the weighting parameter, determining the ratio of importance between a cluster's size and the width of its mean contour. $ \gamma = 3 $ is chosen due to preliminary experiments. An exact evaluation of the best choice for $\gamma$ is left as future work (see Section \ref{sec:conclusion}). We then select the cluster $ C_i $ by $ \argmin\limits_i q(C_i) $.
\item\label{clustering:unmirror} The learned contour is then defined as the inverse $\fft$ of the mean contour of the selected Cluster after a mapping to invert the mirroring (see Step \ref{clustering:mirror}):\\
$ contour\colon [0,\,1) \mapsto \{0,\,...,\,127\}, t \mapsto r_{C_i}\left(\frac{1+t}{2}\right) $
\end{enumerate}

\begin{figure}[htbp]
  \begin{subfigure}{.33\textwidth}
  \resizebox{\textwidth}{!}{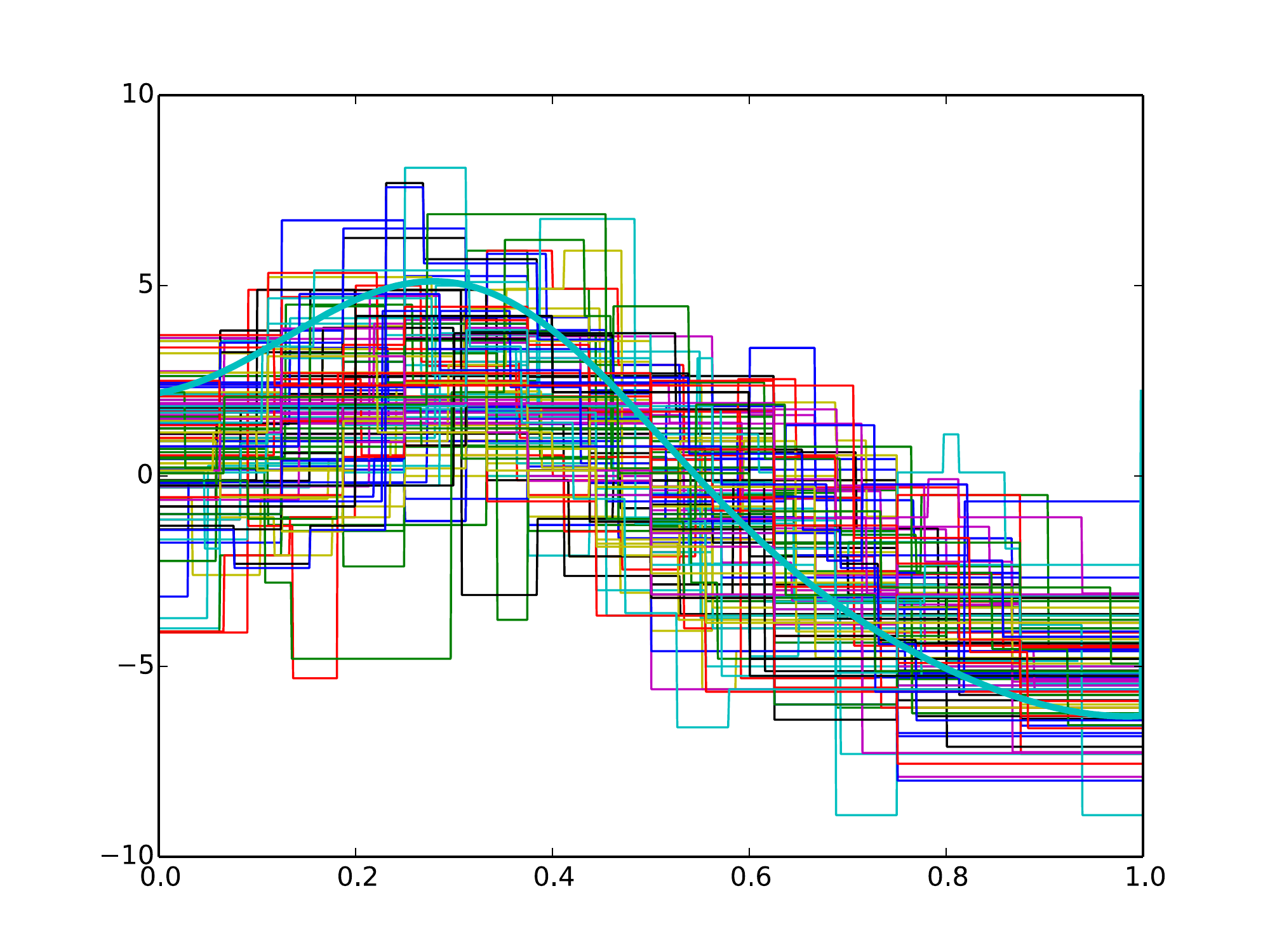}
  \caption{$ \abs{C_1} = 115 $; $ q(C_1) \approx 1.26 $}
  \end{subfigure}%
  \begin{subfigure}{.33\textwidth}
  \resizebox{\textwidth}{!}{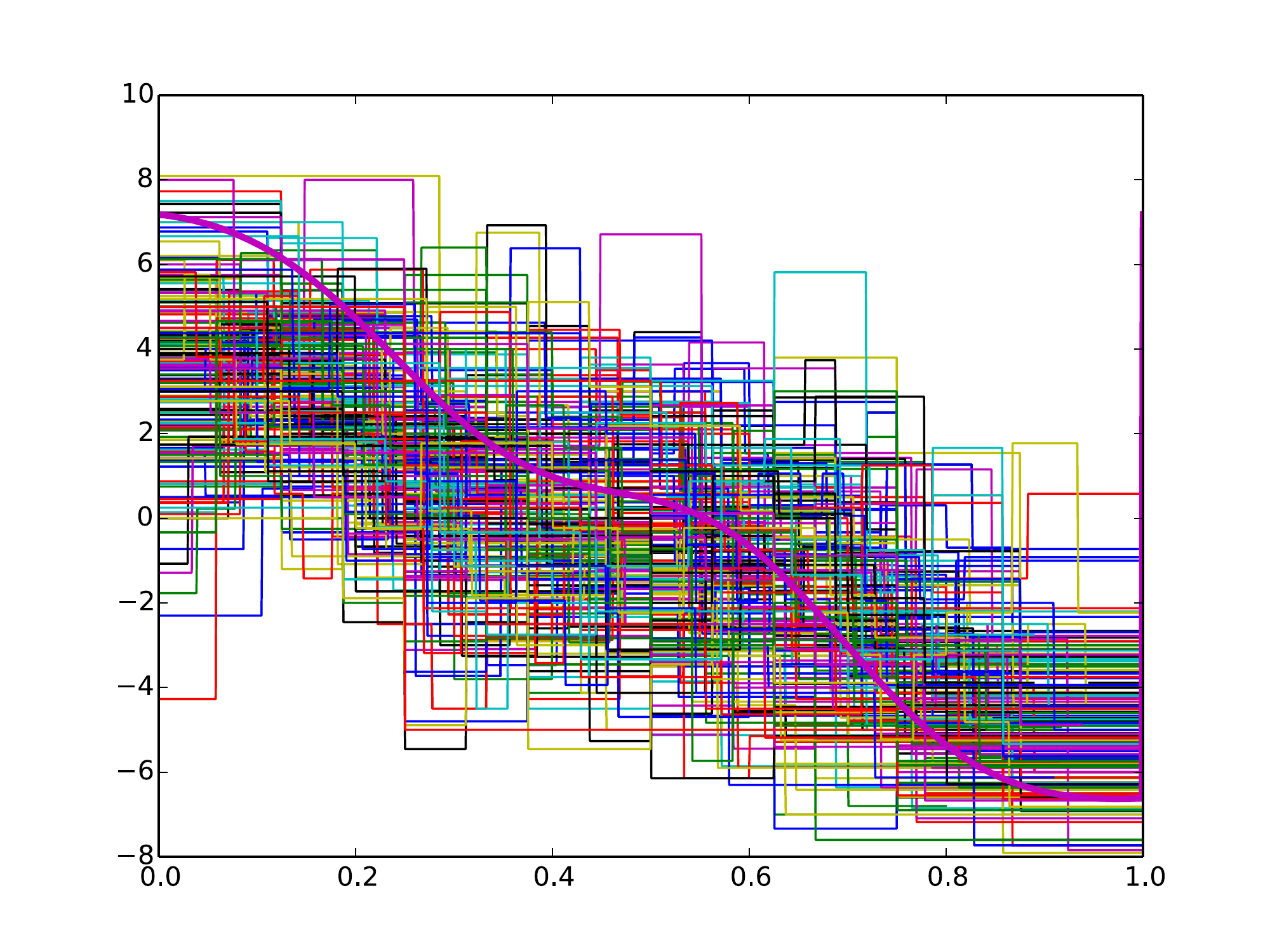}
  \caption{$ \abs{C_2} = 235 $; $ q(C_2) \approx 1.68 $}
  \end{subfigure}%
  \begin{subfigure}{.33\textwidth}
  \resizebox{\textwidth}{!}{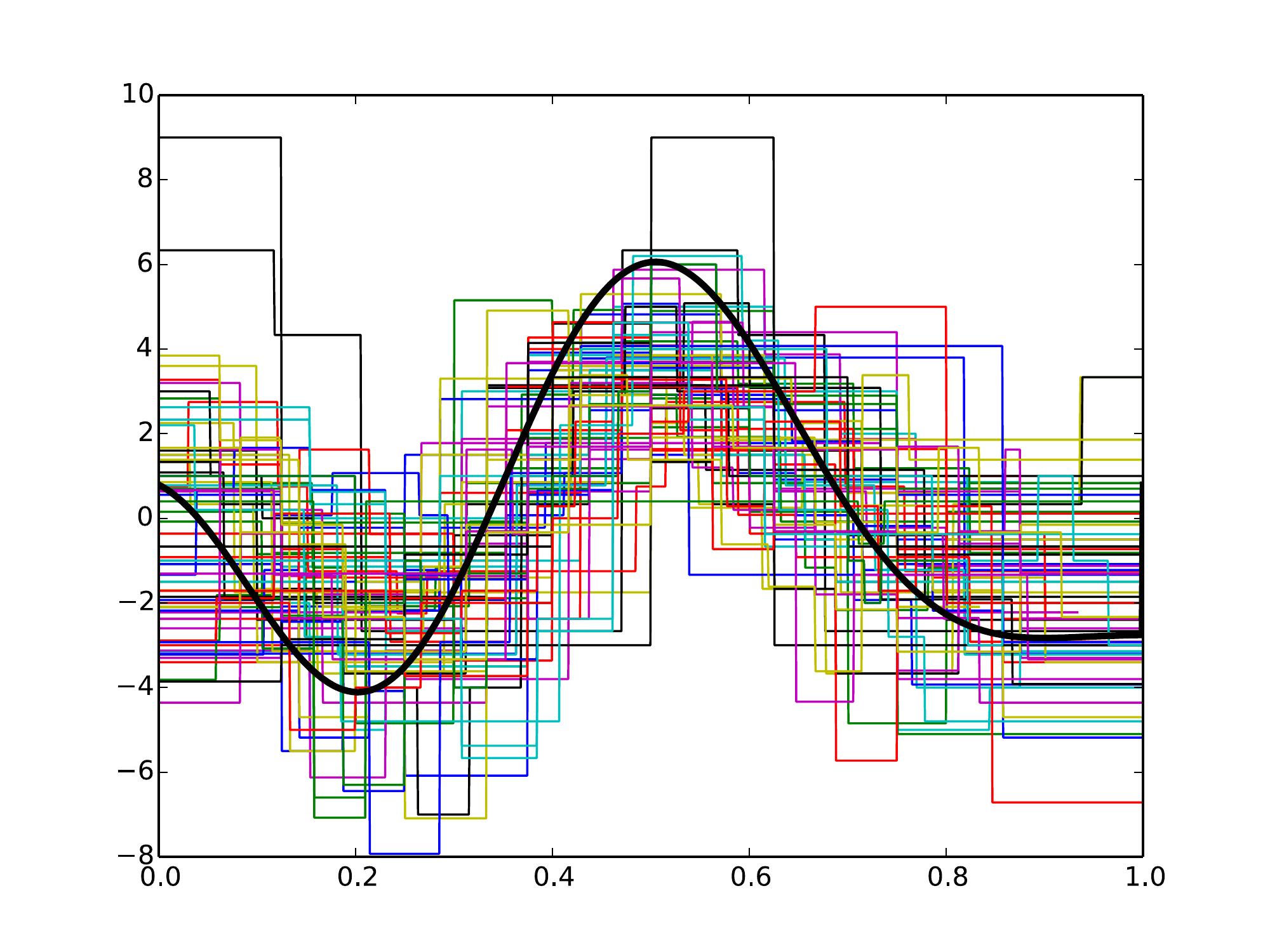}
  \caption{$ \abs{C_3} = 69 $; $ q(C_3) \approx 1.01 $}
  \end{subfigure}\\%
  \begin{subfigure}{.33\textwidth}
  \resizebox{\textwidth}{!}{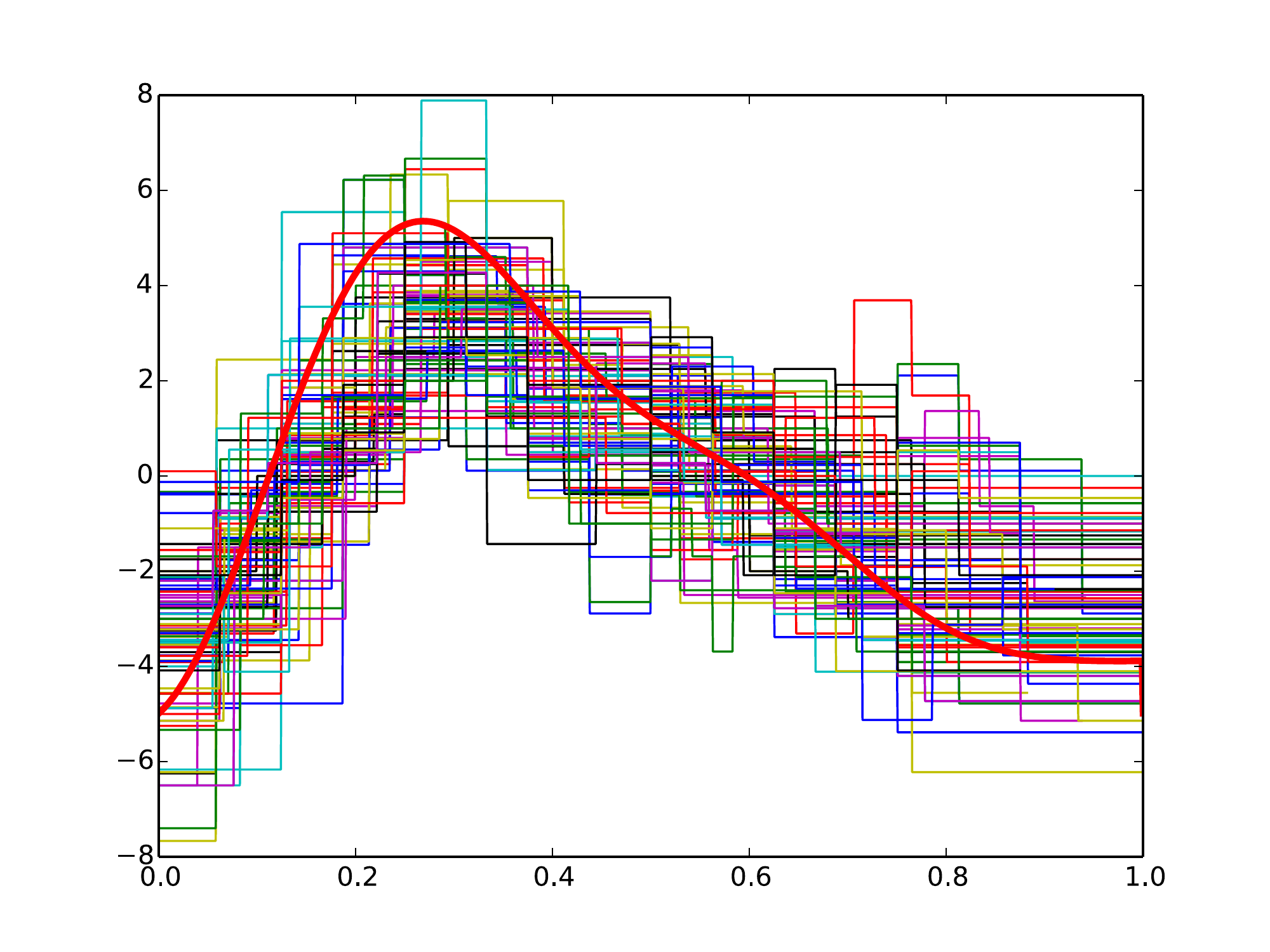}
  \caption{$ \abs{C_4} = 93 $; $ q(C_4) \approx 1.10 $}
  \end{subfigure}%
  \begin{subfigure}{.33\textwidth}
  \resizebox{\textwidth}{!}{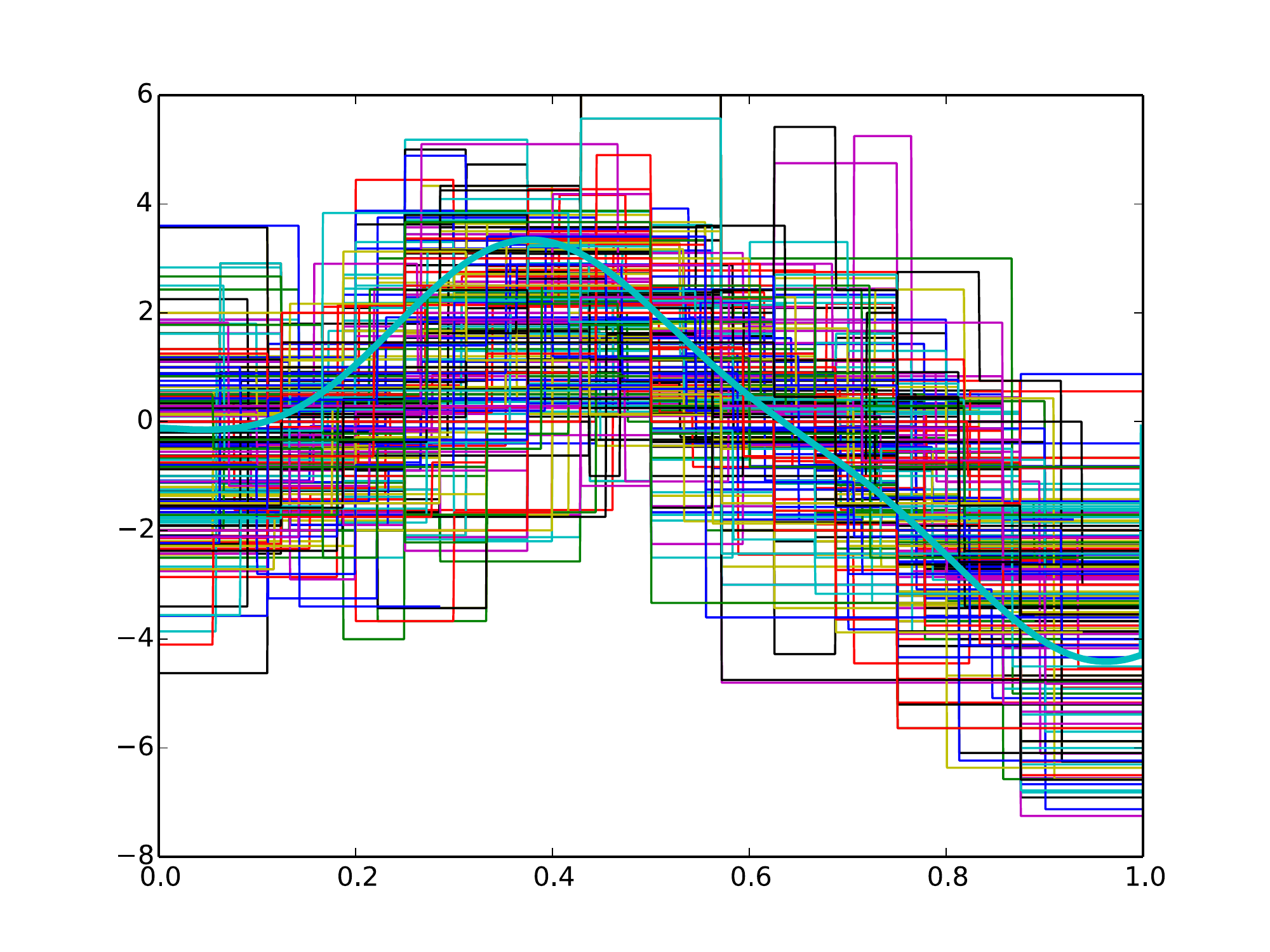}
  \caption{$ \abs{C_5} = 283 $; $ q(C_5) \approx 1.68 $}
  \end{subfigure}%
  \begin{subfigure}{.33\textwidth}
  \resizebox{\textwidth}{!}{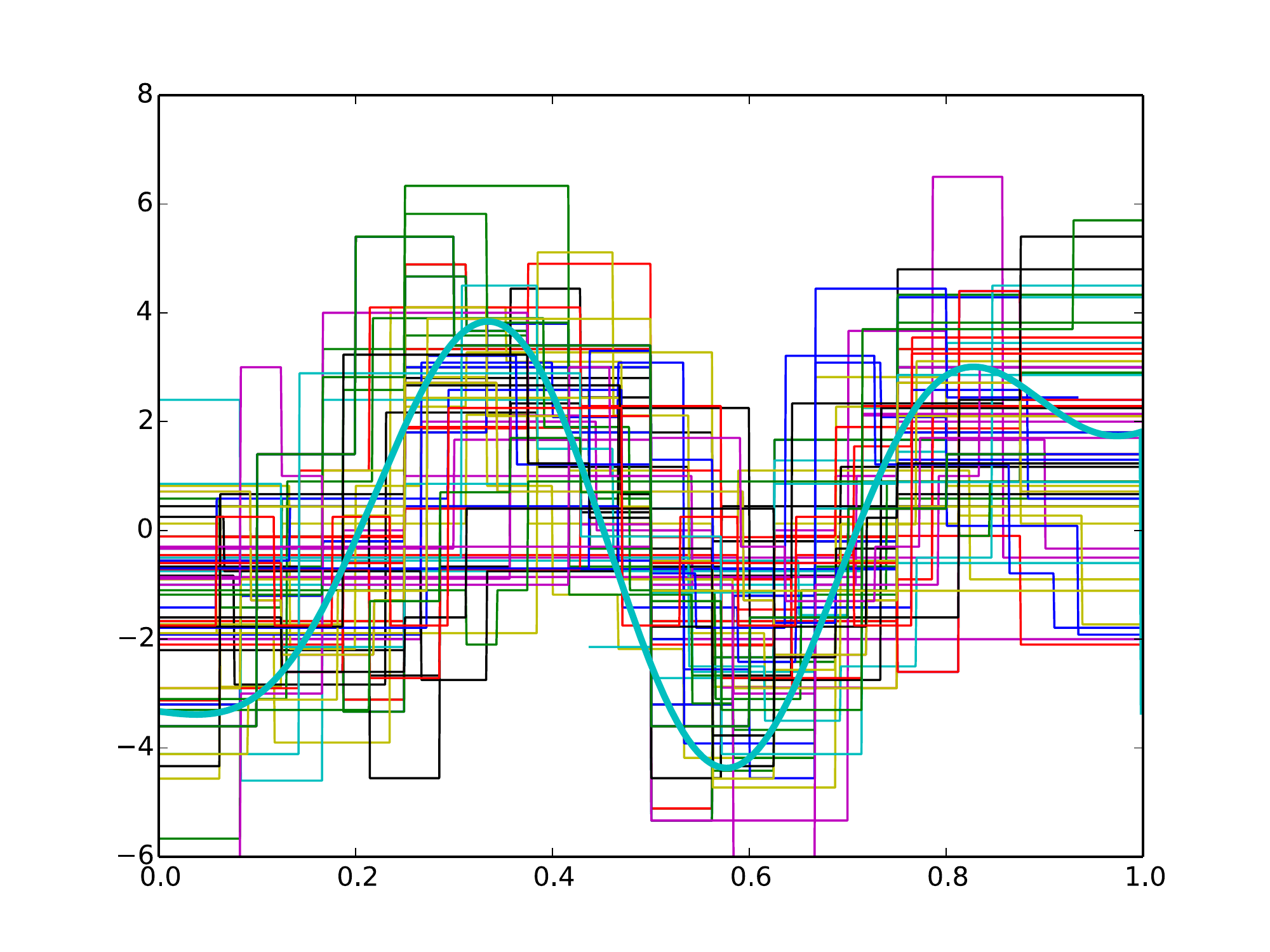}
  \caption{$ \abs{C_6} = 73 $; $ q(C_6) \approx 0.99 $}
  \end{subfigure}\\%
  \begin{subfigure}{.33\textwidth}
  \resizebox{\textwidth}{!}{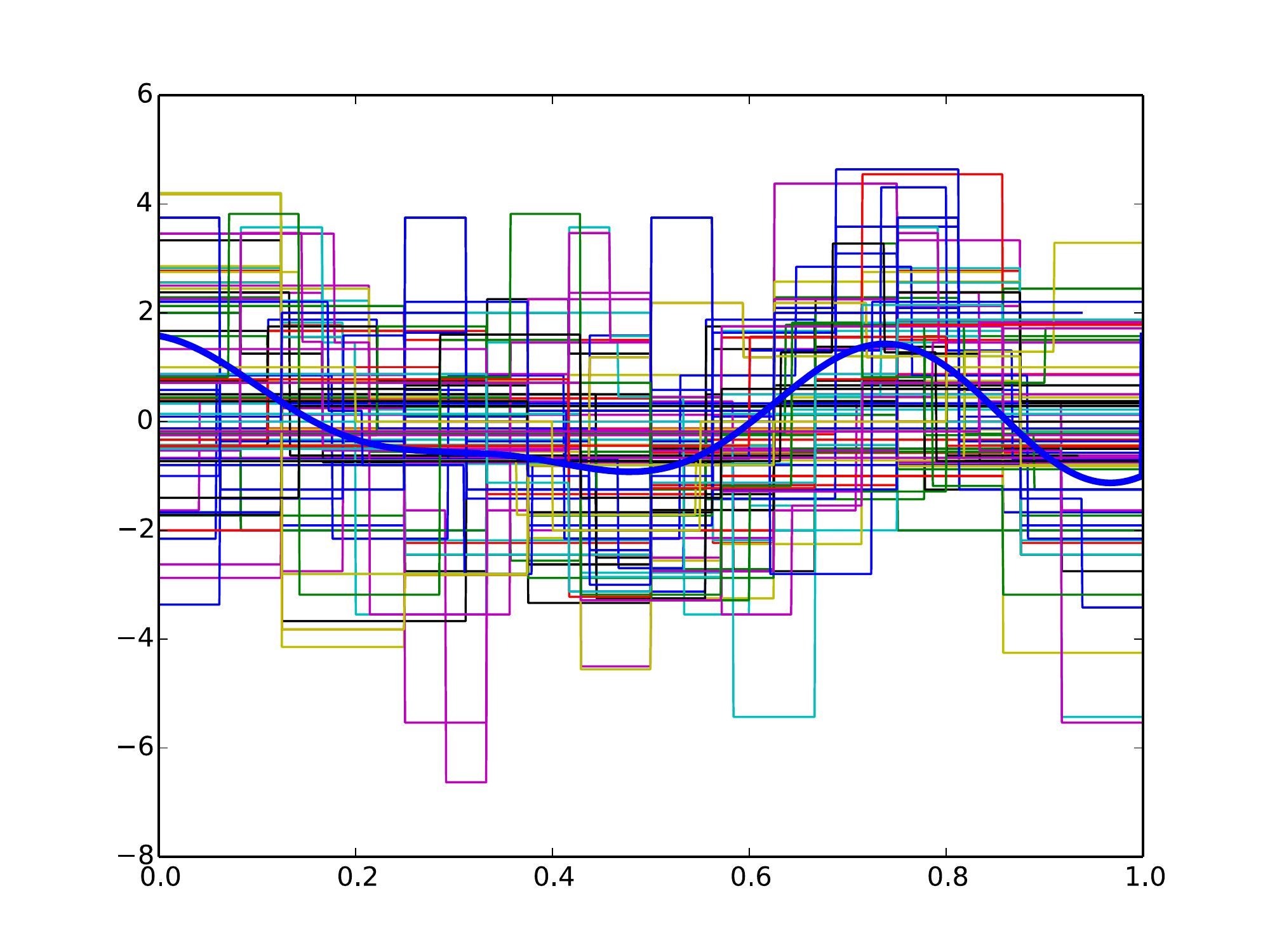}
  \caption{$ \abs{C_7} = 126 $; $ q(C_7) \approx 0.94 $}
  \end{subfigure}%
  \begin{subfigure}{.33\textwidth}
  \resizebox{\textwidth}{!}{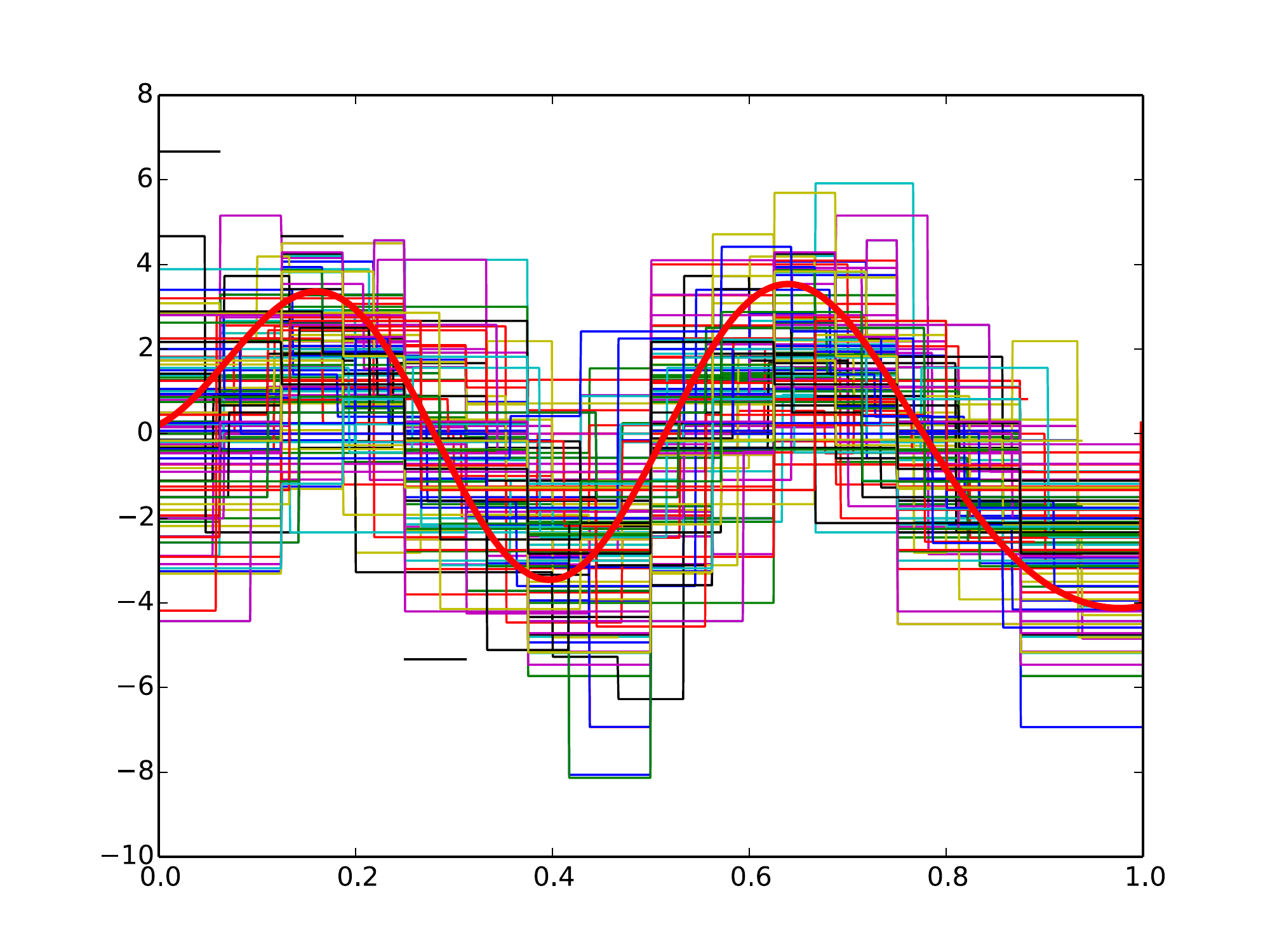}
  \caption{$ \abs{C_8} = 128 $; $ q(C_8) \approx 1.17 $}
  \end{subfigure}%
  \begin{subfigure}{.33\textwidth}
  \resizebox{\textwidth}{!}{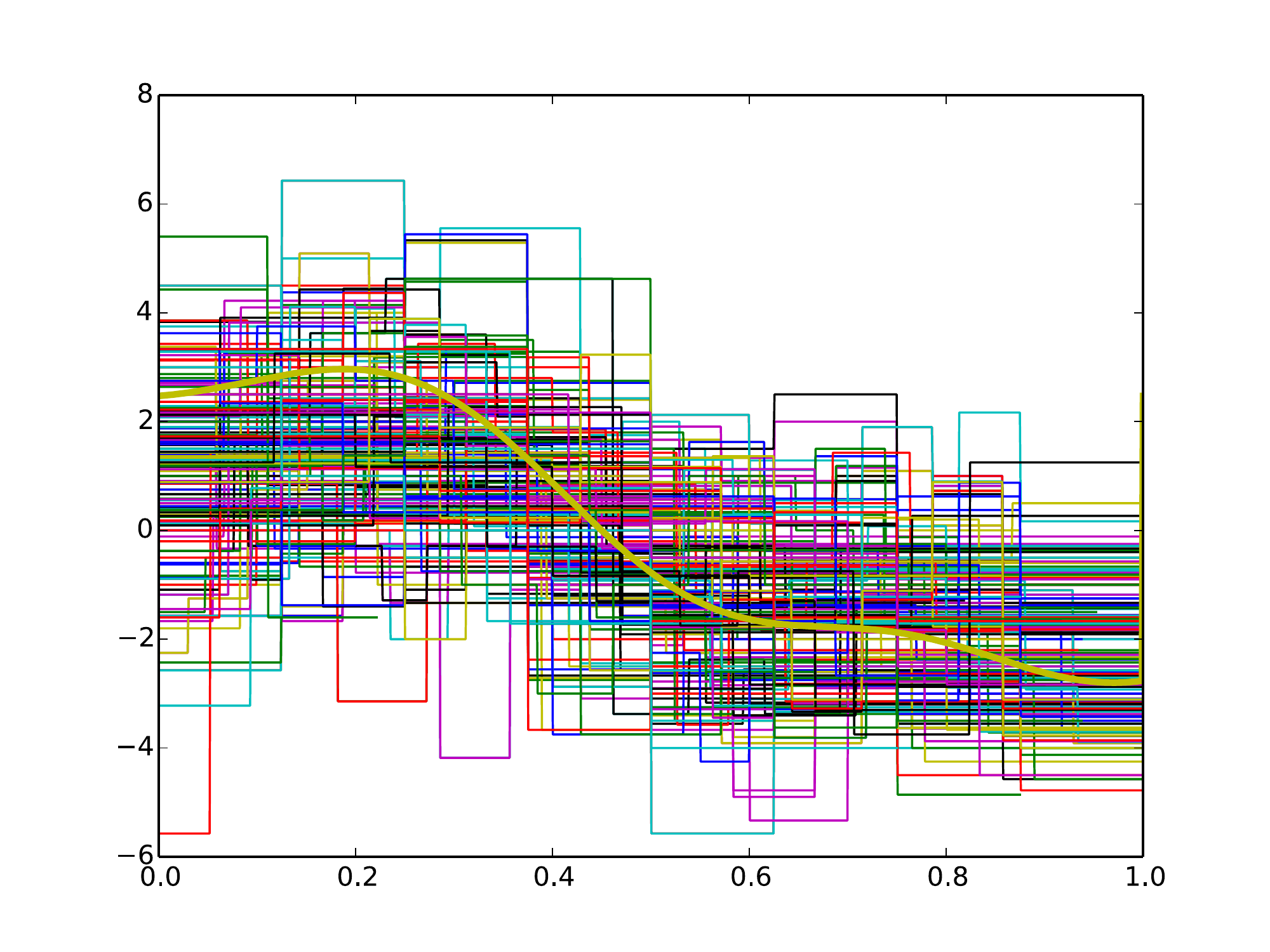}
  \caption{$ \abs{C_9} = 236 $; $ q(C_9) \approx 1.53 $}
  \end{subfigure}\\%
  \begin{subfigure}{.33\textwidth}
  \resizebox{\textwidth}{!}{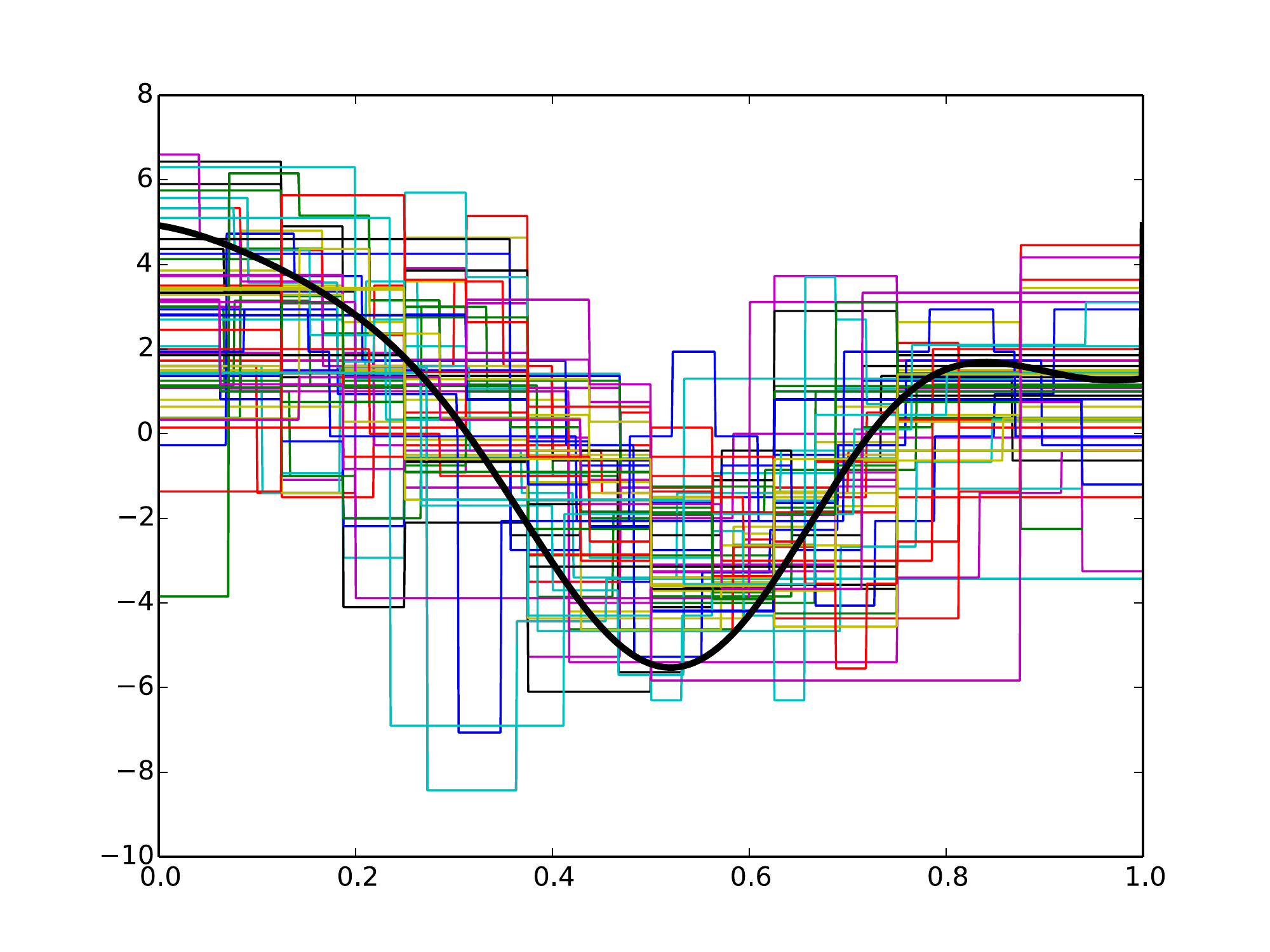}
  \caption{$ \abs{C_{10}} = 62 $; $ q(C_{10}) \approx 0.98 $}
  \end{subfigure}%
  \begin{subfigure}{.33\textwidth}
  \resizebox{\textwidth}{!}{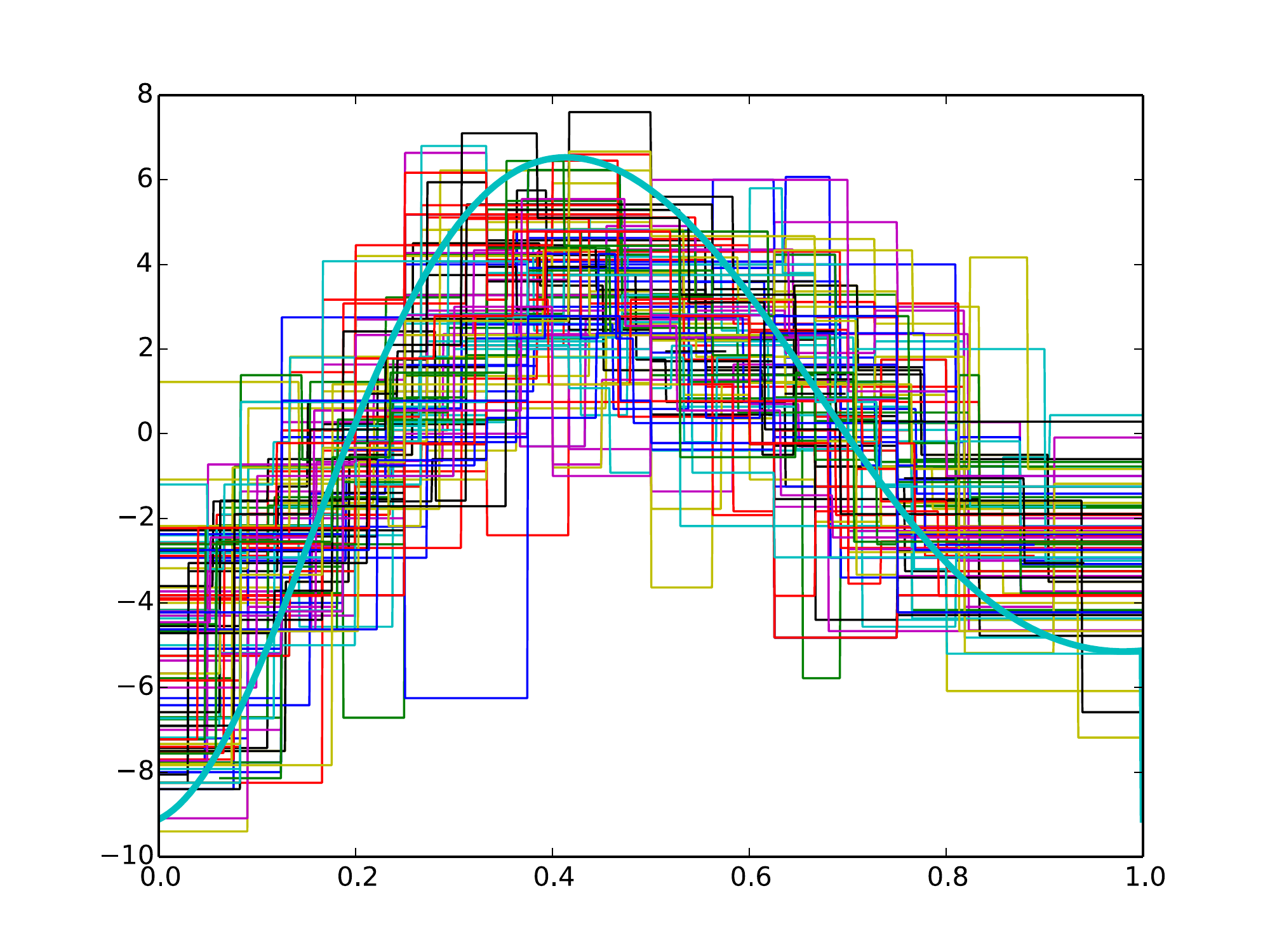}
  \caption{$ \abs{C_{11}} = 94 $; $ q(C_{11}) \approx 1.22 $}
  \end{subfigure}%
  \begin{subfigure}{.33\textwidth}
  \resizebox{\textwidth}{!}{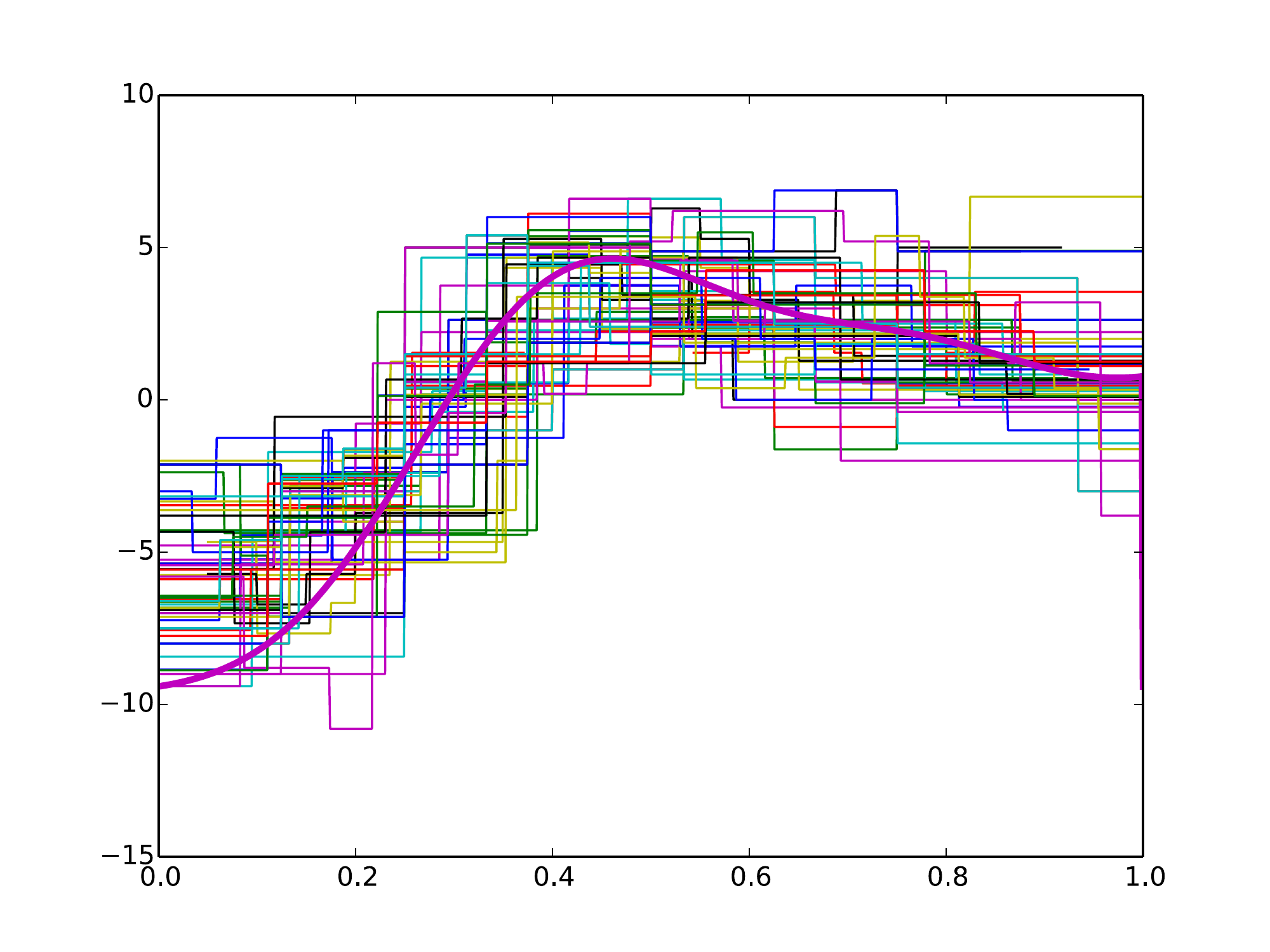}
  \caption{$ \abs{C_{12}} = 60 $; $ q(C_{12}) \approx 1.05 $}
  \end{subfigure}\\%
  \begin{subfigure}{.33\textwidth}
  \resizebox{\textwidth}{!}{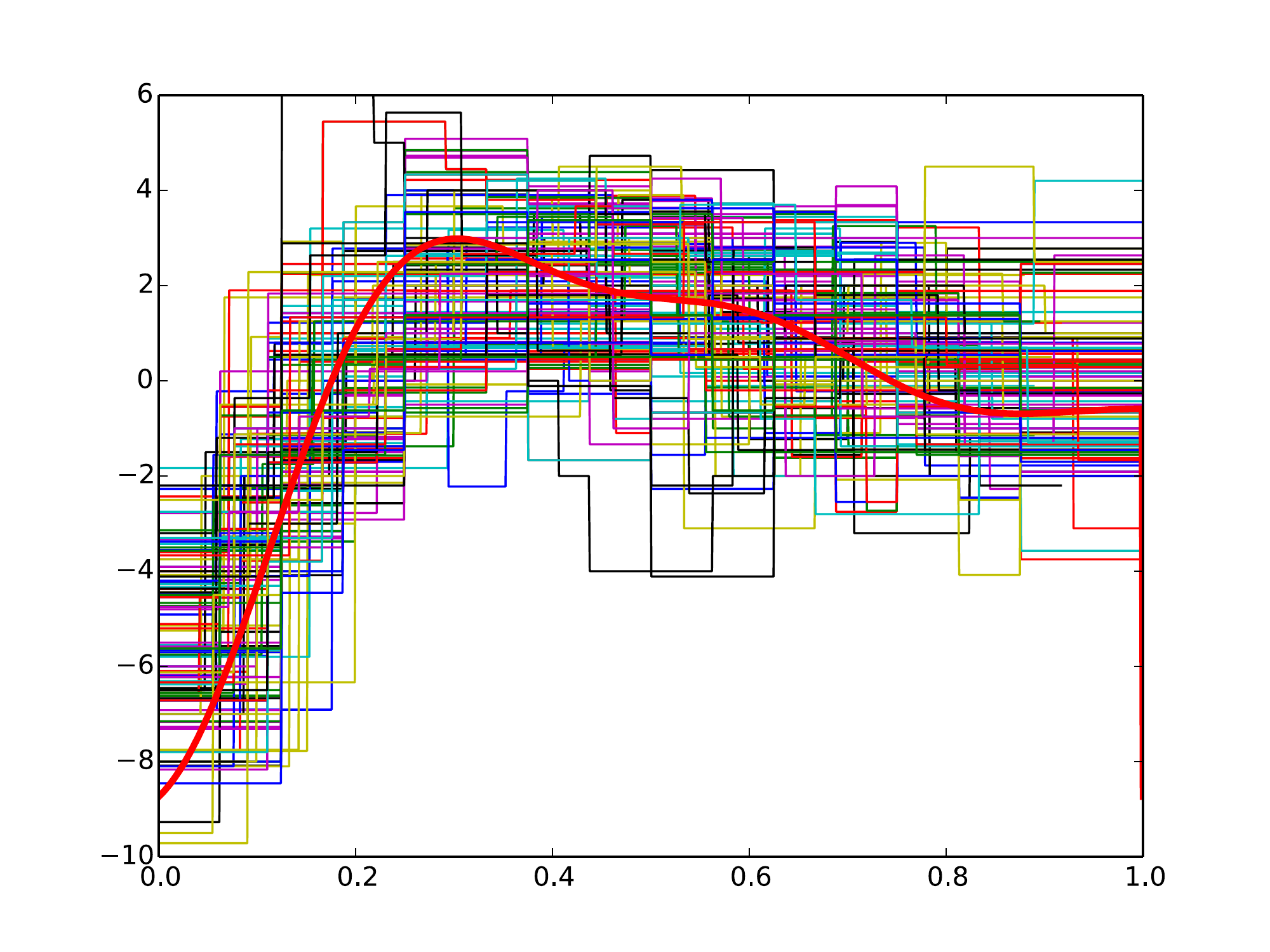}
  \caption{$ \abs{C_{13}} = 135 $; $ q(C_{13}) \approx 1.16 $}
  \end{subfigure}%
  \begin{subfigure}{.33\textwidth}
  \resizebox{\textwidth}{!}{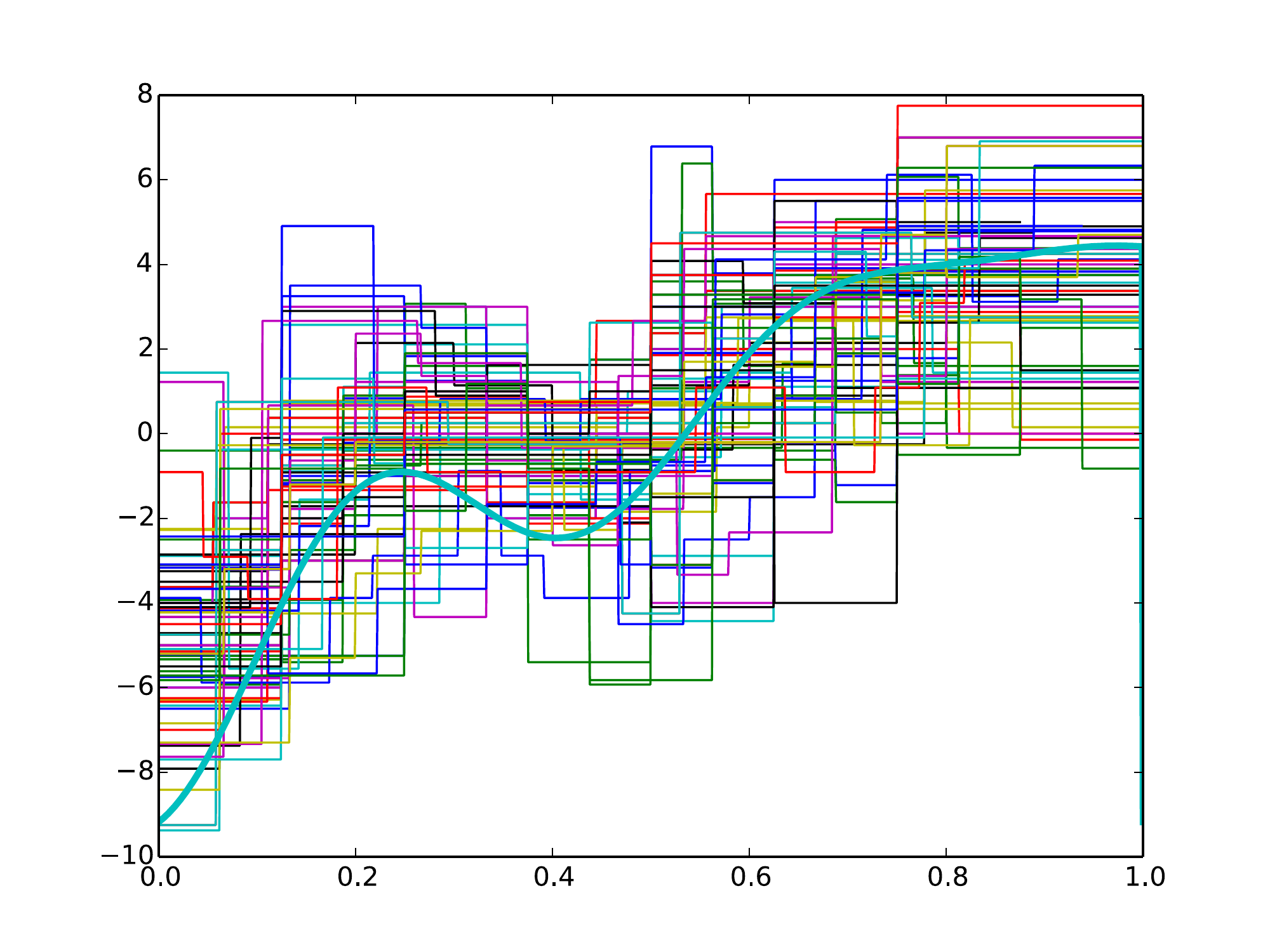}
  \caption{$ \abs{C_{14}} = 73 $; $ q(C_{14}) \approx 1.06 $}
  \end{subfigure}%
  \begin{subfigure}{.33\textwidth}
  \resizebox{\textwidth}{!}{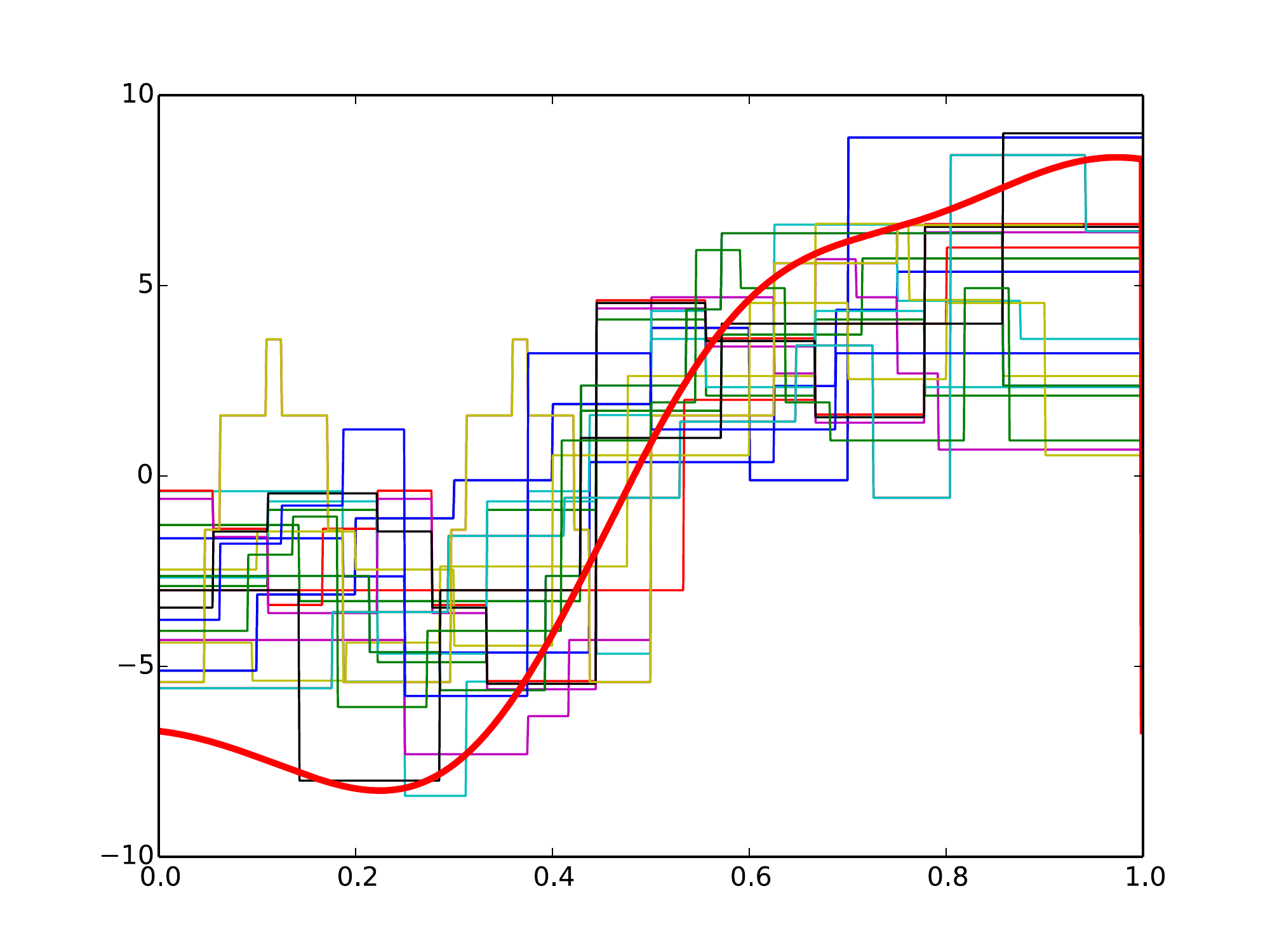}
  \caption{$ \abs{C_{15}} = 23 $; $ q(C_{15}) \approx 1.08 $}
  \end{subfigure}\\%
  \captionsetup{labelformat=empty}
  \caption[Clustering results of contour learning]{}
  \label{fig:contour_clusters}
\end{figure}
%\clearpage
\begin{figure}
  \ContinuedFloat
  \begin{subfigure}{.33\textwidth}
  \resizebox{\textwidth}{!}{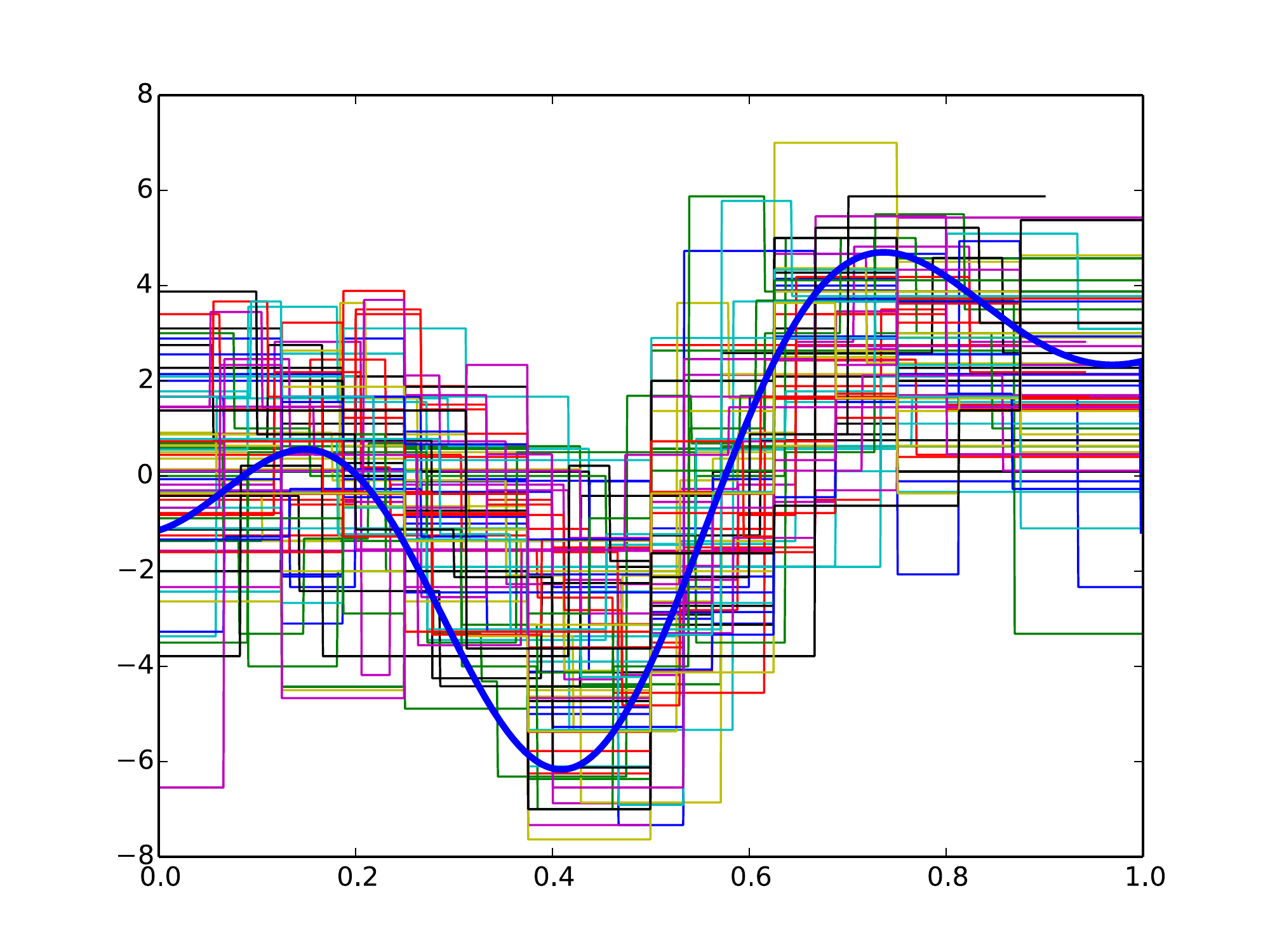}
  \caption{$ \abs{C_{16}} = 77 $; $ q(C_{16}) \approx 1.04 $}
  \end{subfigure}%
  \begin{subfigure}{.33\textwidth}
  \resizebox{\textwidth}{!}{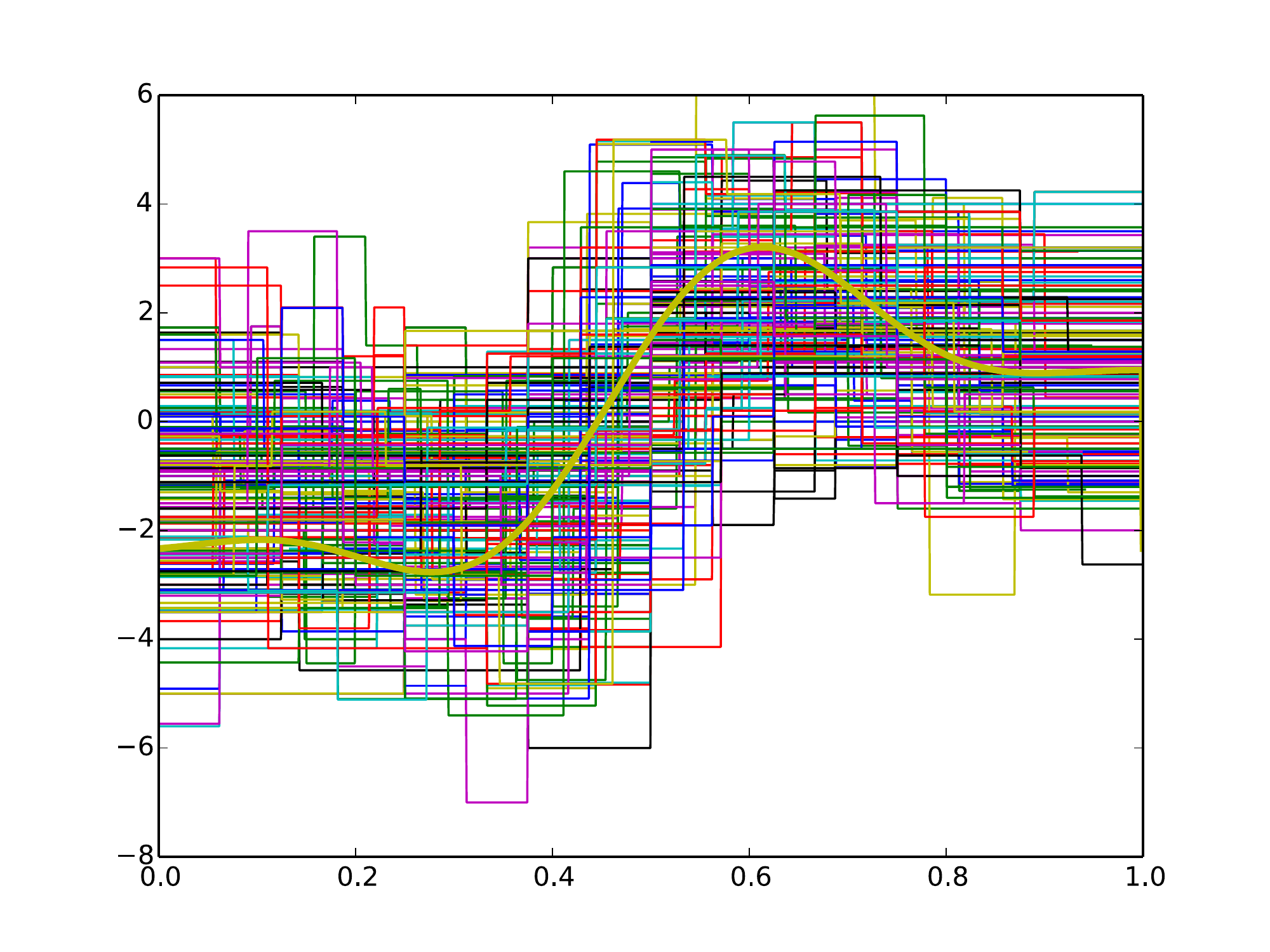}
  \caption{$ \abs{C_{17}} = 229 $; $ q(C_{17}) \approx 1.49 $}
  \end{subfigure}
  \captionsetup{list=off}
  \caption{All phrases of the MTC-FS data set clustered by their melody's contour. $ \abs{C_i} $ is the number of phrases in cluster $ i $ while $ q(C_i) $ is its quality measure as defined in Section \ref{subsubsec:contours} Step 9. In this example, cluster $C_5$ will be selected as $ \argmax\limits_i q(C_i) = 5 $.}
\end{figure}

\subsection{Phrase Composition}
\label{subsec:composing}
\subsubsection{Joining Pitch and Rhythm Generation}
\label{subsubsec:joining_pitch_and_rhythm_generation}
Following the human way of composing, we change KATH's method of composition (see Section \ref{subsubsec:master-thesis}). KATH is only capable of generating a melody with a given number of notes but not a given number $n$ of bars or beats, it composes melodies by first generating $n$ pitches and then $n$ note durations. By changing this strategy to iteratively generate a single duration and then a single pitch, better control over the duration of the melody as well as more context for the composition process is gained. The new algorithm has the information of where in a bar or the piece it is all the time.

With this gain of control over the duration of the melody -- in terms of not needing to know the number of notes we want to use for the composition before knowing the rhythm or the melody -- it is a straight forward modification to the algorithm to make it work with a number of $\frac{4}{4}$-bars as parameter instead of number of notes.

The knowledge of \emph{where} or better \emph{when} the algorithm is within the composition is used to select a Markov model according to the current off-beat. As explained in Section \ref{subsubsec:markov-models} the learning process incorporates training parametric Markov models. The off-beat is used as parameter for the models. The off-beat is calculated as $ c \bmod 1 $, where c is the count in the current bar. The knowledge of the position relative to the whole composition is also used to follow the learned contours (see Section \ref{subsubsec:contours}) as will be described in Section \ref{subsubsec:following-contours}.

After selecting the appropriate Markov model according to the current off-beat it defines a transition vector $ \vec{t} $ for each state (or sequence of states if using higher order Markov models). $ \vec{t} $ contains the probability distribution\footnote{As it's a distribution $ \norm{\vec{t}} = 1 $. It may happen though, that $ \norm{\vec{t}} = 0 $. This is the case, when the training data does not contain any transition from the corresponding (sequence of) state(s).} of transition for each possible state. We call the transition vector which depends on the off-beat and the previous sequence of states $ \vec{t_p} $ if we refer to pitches as states, $ \vec{t_d} $ if we refer to durations as states, and $ \vec{t} $ if we talk about both kinds of transition vectors.

Let $ \vec{s} $ be the vector of all sates of the Markov model. Note that states are numerical in our case (durations as well as pitches)\footnote{The MIDI format encodes pitches with integers. We make use of that here. Durations are normalized to quarter notes. So a quarter note has the value $1$ while a eight note has the value $0.5$ etc.}. We will use $ \vec{s_p} $ when talking about pitches as states and $ \vec{s_d} $ for durations as states.

\subsubsection{Ending Constraints}
\label{subsubsec:ending-constraints}
Assume $ \norm{\vec{t}} = 1 $ as all cases where $ \norm{\vec{t}} = 0 $ are caught before as will be described in Section \ref{subsubsec:backtracking}. In the next step, we apply ending constraints to $ \vec{t_p} $ to obtain $ \vec{t_p'} $ and to $ \vec{t_d} $ to obtain $ \vec{t_d'} $.
\begin{itemize}
\item The duration transition vector $ \vec{t_d} $ is manipulated in a way to make sure that no longer duration is generated than time (counts) is left in the composition. Let $ c $ be the current count in the composition and $ c_{total} $ the wanted length of our composition in counts. We now define a duration manipulation vector $ \vec{m_d} \in \mathbb{R}^{dim(\vec{s_d})} $ element-wise as
$\vec{m_{d_i}} =
\begin{cases}
  1 & \vec{s_{d_i}} \leq c_{total}-c\\
  0 & \text{else}
\end{cases}
$. We then calculate $ \vec{t_d'} = \normalized{\vec{t_d}\bigodot\vec{m_d}} $.
\item The pitch vector is manipulated in a way, that the last pitch in a melody is one of the tonic triad. As we are working only in \emph{C major} this is \emph{c}, \emph{e} and \emph{g} in any octave. Since the rhythm is generated first, the algorithm can easily determine which pitch is the last one. This is calculated as\\
$ \vec{t_p'} = 
\begin{cases}
  \normalized{\vec{t_p}\bigodot\vec{m_p}} & \text{if last pitch}\\
  \vec{t_p}                               & \text{else}
\end{cases}
$, with
$\vec{m_{p_i}} =
\begin{cases}
  1 & \vec{s_{p_i}} \in \{\text{c},\,\text{e},\,\text{g}\}\\
  0 & \text{else}
\end{cases}
\in \mathbb{R}^{\dim(\vec{s_p})}
$
\end{itemize}

\subsubsection{Following Contours}
\label{subsubsec:following-contours}
Now that we have two contour functions over time that describe the course of the composition, one for its melody and one for its rhythm (see Section \ref{subsubsec:contours}). We have to manipulate the trained Markov models so that their output follows the respective contour. The basic idea here is to manipulate the transition probabilities with a Gaussian filter and let its center follow the learned contours. This will make the melody and the rhythm loosely follow the contours as well.

We therefore define a filter vector $ \vec{f} \in \mathbb{R}^{\vert\vec{s}\vert} $ element-wise as $ \vec{f}_i = \varphi_{\mu, \sigma^2}(\vec{s}_i) $ where $ \mu $ is the value of the contour at the current time in the composition and $ \sigma^2 $ a tweaking parameter. $ \sigma^2 = 4 $ for melody contours (which means a standard deviation is four semi-tones) and $ \sigma^2 = 0.33 $ for the rhythm contours (which means a standard deviation of an eighth triplet) are chosen as preliminary experiments showed they work well. A full evaluation of the effects and the best choices is left as future work (see Section \ref{sec:conclusion}).

Let $ \vec{t'} $ be the transition vector as described in Section \ref{subsubsec:ending-constraints}. We then calculate the filtered transition vector $ \vec{t'} = \normalized{\vec{t}\bigodot\vec{f}} $ and use this instead to draw the next state\footnote{States are, like in the original work \cite{Kathiresan2015}, drawn by choosing a random number $ r \in [0, 1] $,  calculating the accumulative sum $ \vec{s} = accsum(\vec{t'}) $ and returning $ \vec{s_i} $ with the smallest $ i $, $ r \leq \vec{s_i} $.}.

% A Markov model defines a transition vector $ \vec{t} $ for each state (or sequences of states if using higher order Markov models). That vector represents the transition probabilities to each state. Since states are numerical in our case (durations and pitches) the vector can also be represented as a discrete function $ t(state) = P_{transition}(state) $. To make the composition follow the contour loosely we calculated $ t' = \varphi_{\mu, \sigma} \cdot t $ where $ \varphi $ is a Gaussian curve, $ \mu $ being the value of the contour at the current time in the composition and $ \sigma $ a tweaking parameter. We decided to choose $ \sigma = 4 $ for melody contours (which means a standard deviation is four semi-tones) and $ \mu = 0.33 $ for the rhythm contours (which means a standard deviation of a eighth triplet). We then interpreted $ t' $ as a vector $ \vec{t'} $ and normalized it to $ \vert\vec{t'}\vert = 1 $. Using this manipulated transition vector we were able to make the melody and the rhythm follow the learned contours.

\subsubsection{Backtracking}
\label{subsubsec:backtracking}
As the algorithm composes a melody, it might find itself in a situation where the previous states do not appear in the training data. In that case, $ \vec{t} = \vec{0} $. All the manipulations in Sections \ref{subsubsec:ending-constraints} and \ref{subsubsec:contours} would fail as $ \normalized{\vec{0}} $ implies a division by zero. Skipping the manipulations wouldn't help either as there is no information on which state could or should be generated next. A way around that problem is to use backtracking. Since the parametric Markov model's transition probabilities depend on the count, it might well happen that the problem of insufficient training data can be resolved by just drawing another duration state while keeping the pitch. Therefore, our algorithm sticks to one pitch after failing and will not change that pitch until all possible durations for it have failed. This can be quite memory consuming if the training data is small. For every leaf in the search tree (which corresponds to an instance of insufficient data) the algorithm stores the current pitch (the one we decided to try all durations for) and all durations tried yet for each pitch tried yet in its parent node. For the exact implementation, see the code at \emph{melody\_generation.py:MelodyTreeTruncate}.

\section{Results}
\label{sec:results}
We now want to evaluate the results of the algorithm proposed in Section \ref{sec:proposed_method}. The results will be evaluated on the basis of a human benchmark incorporating aesthetics and ambiguity error. Therefore, we will first create a baseline from KATH and the training data (see Section \ref{subsec:baseline}). We will then explain the online survey set up (see Section \ref{subsec:listenig-test}), and finally, we will present and interpret the results of the survey (see Sections \ref{subsec:results-listening-test} and \ref{subsec:interpretation-listening-test}).

\subsection{Generating the Baseline}
\label{subsec:baseline}
The algorithm proposed by Kathiresan \cite{Kathiresan2015} is only capable of learning from one MIDI file and composes a melody with a given number of notes. To be comparable to our algorithm a set of ten melodies was created with our algorithm. It was run five times with only one melody as training data, composing two  melodies with a length of four $ \frac{4}{4} $-bars each time. For Kathiresan's algorithm a thin wrapper is needed to force it to compose melodies with four $ \frac{4}{4} $-bars. The wrapper reads the training data and calculates from the longest and the shortest note in the data the minimal and maximal number of notes that fit into four $ \frac{4}{4} $-bars. In a loop a random number $ n $ in that range is drawn. Kathiresan's algorithm is then called to compose a melody with $ n $ notes. This is repeated until the composed melody has a length of four $ \frac{4}{4} $-bars. The same training melodies used for our algorithm are used for Kathiresan's as well to achieve comparability of the results.

\subsection{Conducted Listening Test}
\label{subsec:listenig-test}
The conducted listening test was set up as an online survey. This is possible as there are no requirements to be met by the participants except for the ability to hear. Nor are requirements to be met by the subject's environment expect the ability to hear the presented stimuli. To guarantee that a test stimulus was presented to all subjects, before they started the survey. They were asked if they could hear it. If not, we excluded them from our study.

One of the main advantages of an online survey for our case is the fact that it is much easier to acquire a higher number of participants than with a face-to-face test. Knowing that online surveys will most likely not result in a representative sample of the population \cite{Wright2005} we prefer having a big sample over a small but representative one. To avoid costs, the server software is implemented by ourselves and hosted on already existing infrastructure.

To prevent over-tiring of the test subjects, the test time was limited. Therefore, the subjects were randomly split into five groups. Each subject was first asked to answer a few personal questions. A melody was then presented and the subjects were asked three questions about it: 1) if the melody was known to the them, 2) if the subjects think the melody is human or computer composed, and 3) how much the subjects like the melody). This was repeated for twelve different melodies. The twelve melodies presented to each subject were different for each group and in random order for each subject, but the distribution of types were the same for all groups. See Table \ref{tab:melody-types}. The exact questions asked can be found in Appendix \ref{appendix:poll}.

\begin{table}[H]
  \begin{center}
    \begin{tabu} to \textwidth { r | r *{3}{| X[l]} }
      type  & qty.  & source (type) & parameters & comment \\
      \hline
      1 & 2 & Randomly picked from MTC dataset & Phrases of length four $\frac{4}{4}$-bars & \\
      \hline
      2 & 2 & Manually composed by us for this test & Phrases of length four $\frac{4}{4}$-bars in a similar style to phrases of the MTC dataset & \\      
      \hline
      3 & 2 & Generated from Kathiresan's work \cite{Kathiresan2015} & Length forced to be four $\frac{4}{4}$-bars & see Section \ref{subsec:baseline} \\
      \hline
      4 & 2 & Generated by our algorithm & Learn from one mtc-melody only & To be comparable to Kathiresan's work \\
      \hline
      5 & 1 & \multicolumn{1}{c|}{\ditto} & from 10 melodies & \\
      \hline
      6 & 1 & \multicolumn{1}{c|}{\ditto} & \ditto\ 20 \ditto & \\
      \hline
      7 & 1 & \multicolumn{1}{c|}{\ditto} & \ditto\ 50 \ditto & \\
      \hline
      8 & 1 & \multicolumn{1}{c|}{\ditto} & \ditto\ all (378) \ditto & \\
    \end{tabu}
    \caption[Melody types presented to the subjects]{This table shows the distribution of melody types presented to each subject regardless of the group assigned.\\    
    All melodies composed by our algorithm implicitly share the following parameters: length of four $\frac{4}{4}$-bars and Markov model order four.\\
    Please take note of the difference between the terms ``melody'' and ``phrase'', especially when talking about the MTC dataset. A melody can be composed of multiple phrases.}
    \label{tab:melody-types}
  \end{center}
\end{table}

\subsection{Results of the Listening Test}
\label{subsec:results-listening-test}

The online survey was open for about 6 weeks. A total of 248 subjects participated. In the following the answers are visualized.

\begin{figure}[H]
  \center
  \resizebox{0.9\textwidth}{!}{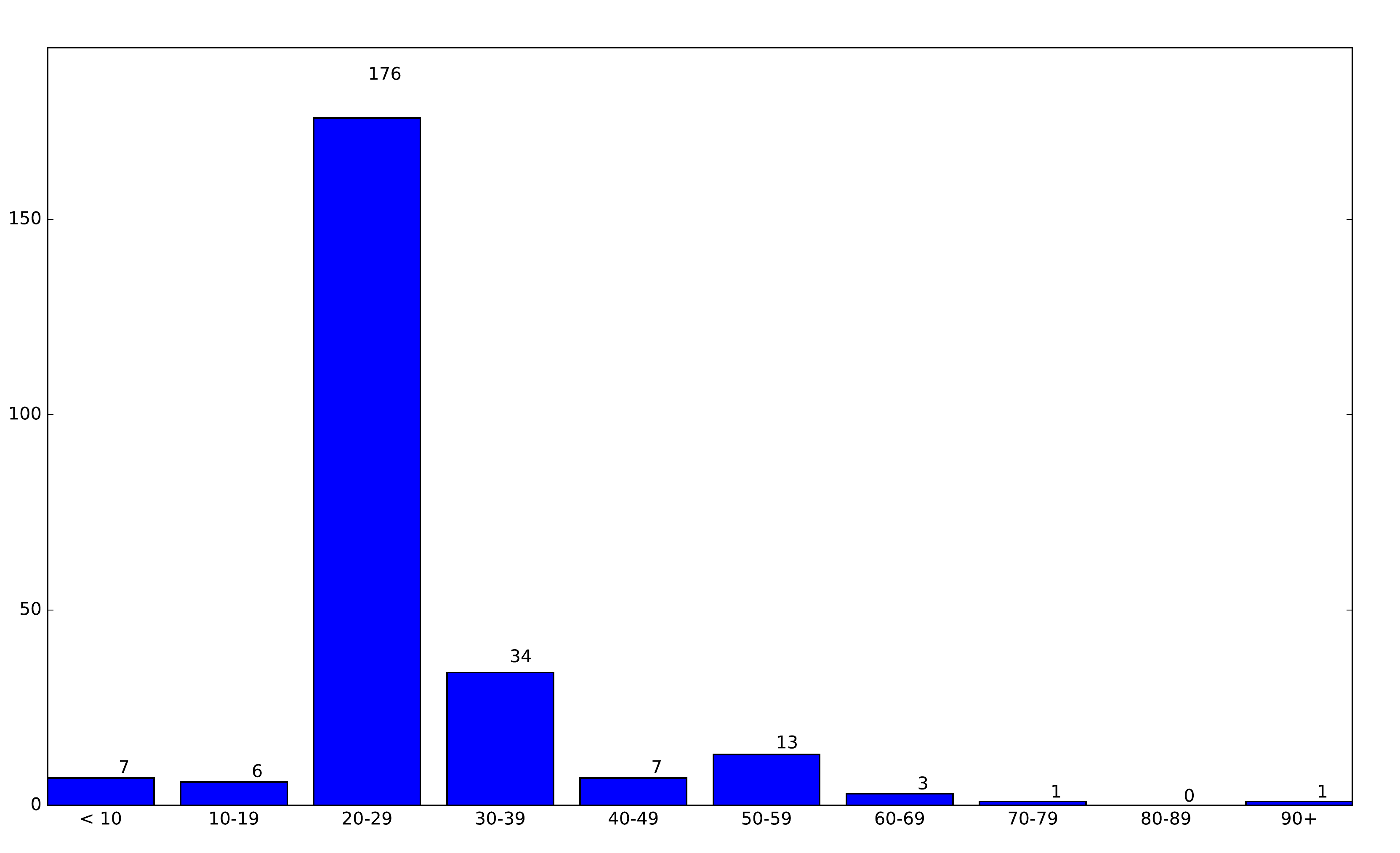}
  \caption[Survey result: age]{Age of participants.}
  \label{fig:poll-result-age}
\end{figure}

\begin{figure}[H]
  \begin{subfigure}{.33\textwidth}
  \resizebox{\textwidth}{!}{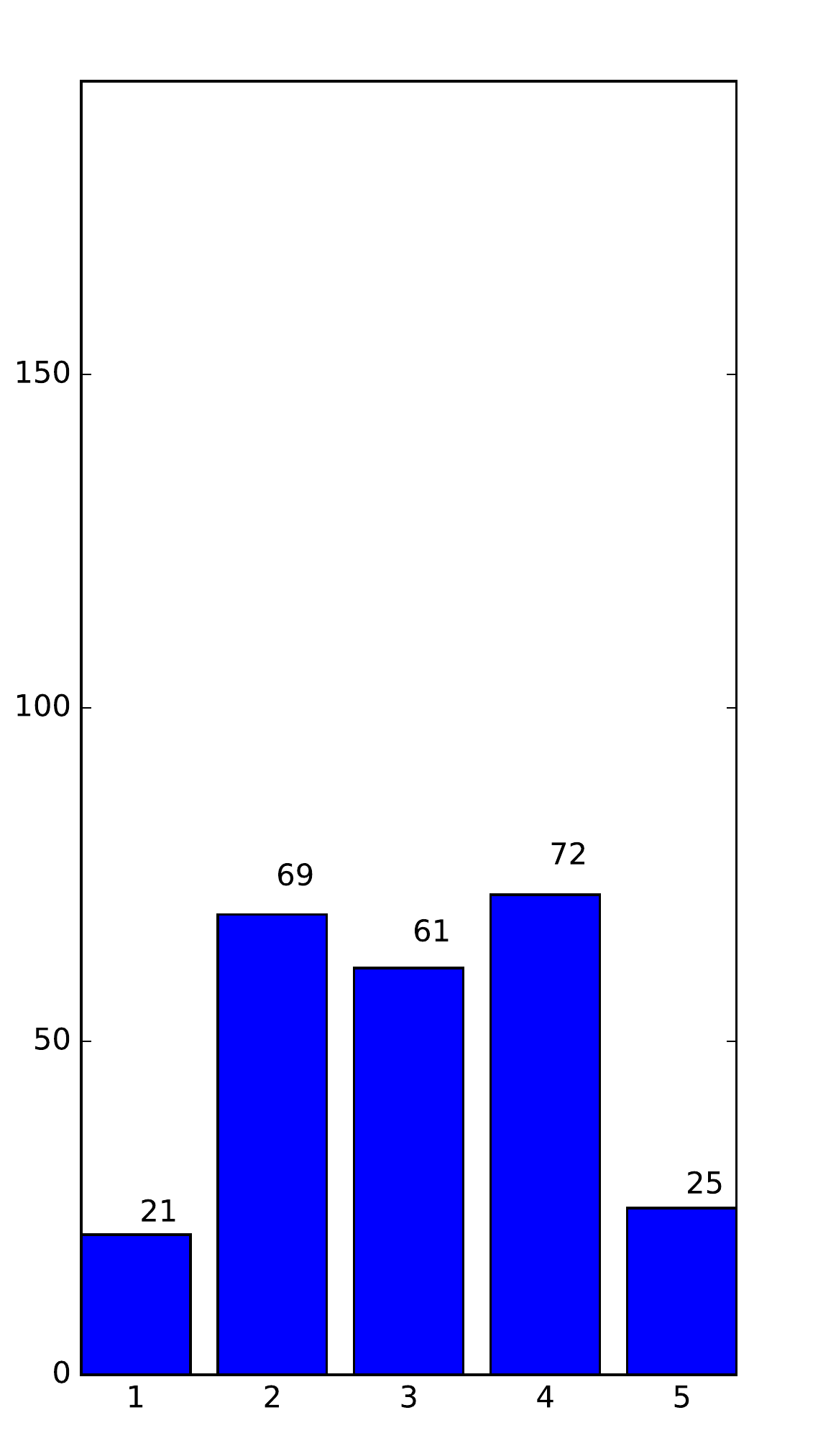}
  \caption{all\\$ \mu = 3.04,\ \sigma^2 = 1.14 $}
  \end{subfigure}%
  \begin{subfigure}{.33\textwidth}
  \resizebox{\textwidth}{!}{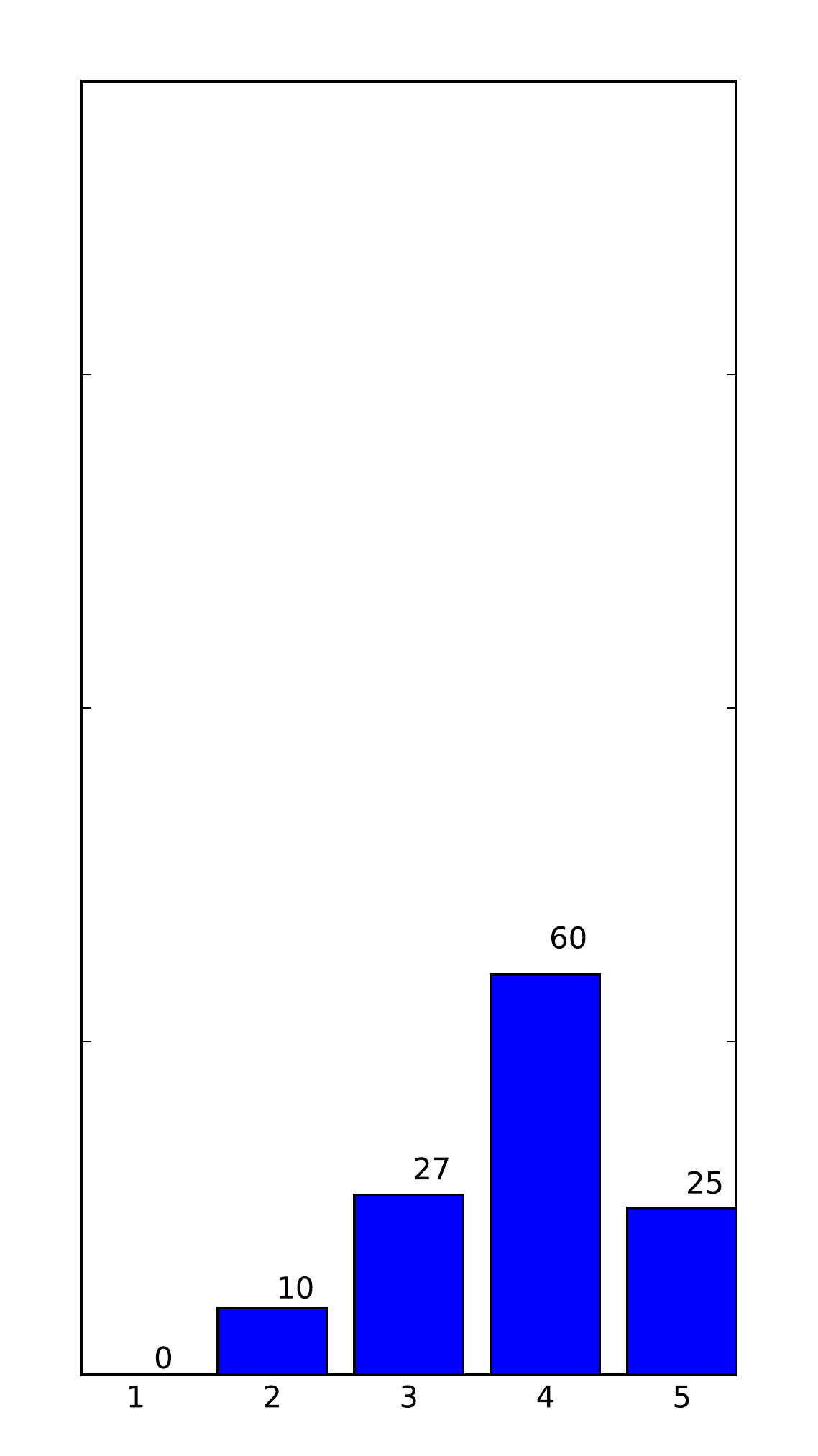}
  \caption{playing an instrument\\$ \mu = 3.82,\ \sigma^2 = 0.85 $}
  \end{subfigure}%
  \begin{subfigure}{.33\textwidth}
  \resizebox{\textwidth}{!}{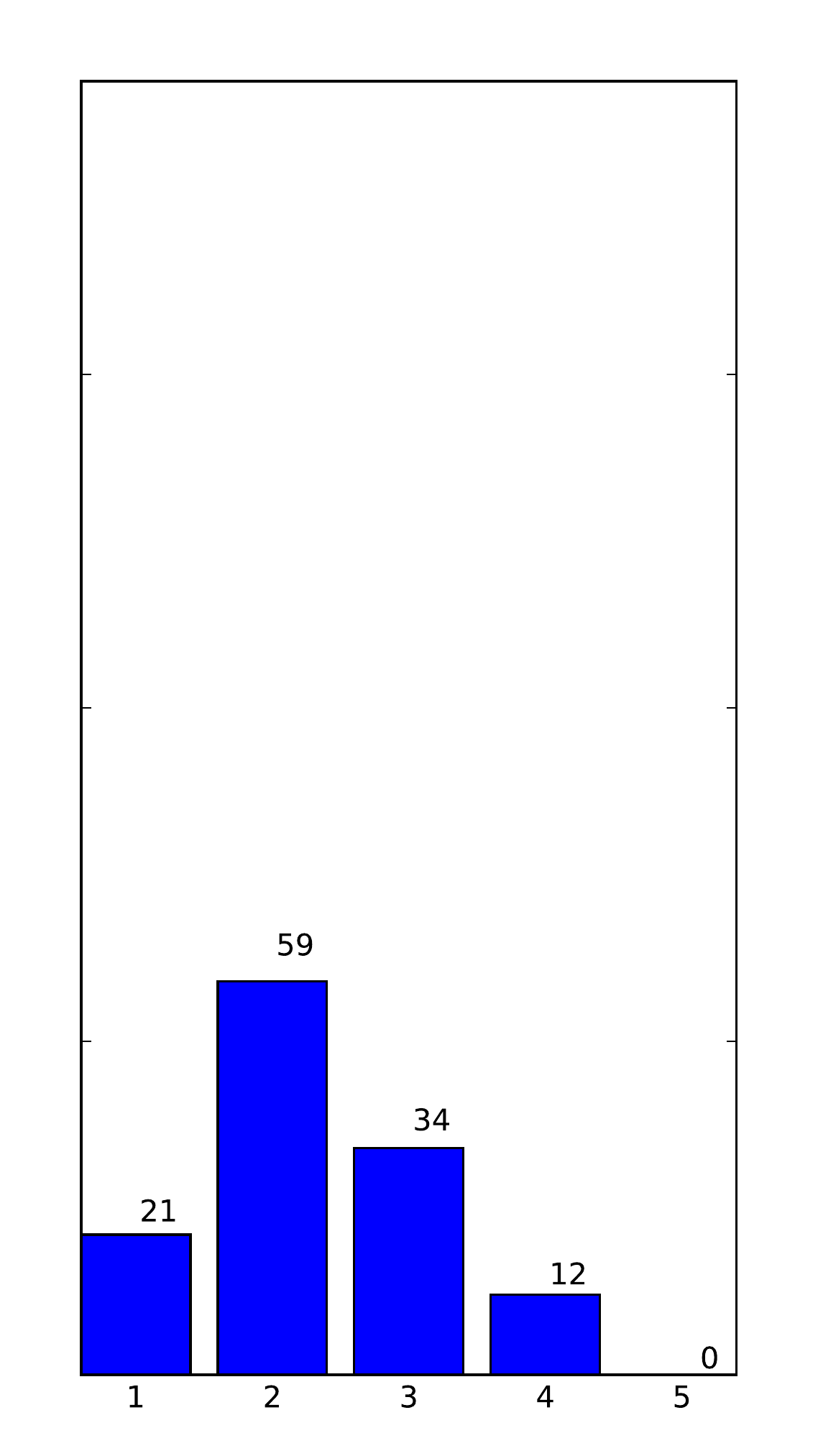}
  \caption{playing no instrument\\$ \mu = 2.29,\ \sigma^2 = 0.86 $}
  \end{subfigure}\\%
  \caption[Survey result: self-estimated musicality]{Self-estimated musicality of the subjects. The subjects were asked to self-estimate their musicality on this scale:\\
  1 -- ``not at all'', 2 -- ``a bit'', 3 -- ``moderate'', 4 -- ``good'', 5 -- ``excellent''}
  \label{fig:poll-result-musicality}
\end{figure}

\begin{figure}[H]
  % pie chart made with https://live.amcharts.com
  \resizebox{\textwidth}{!}{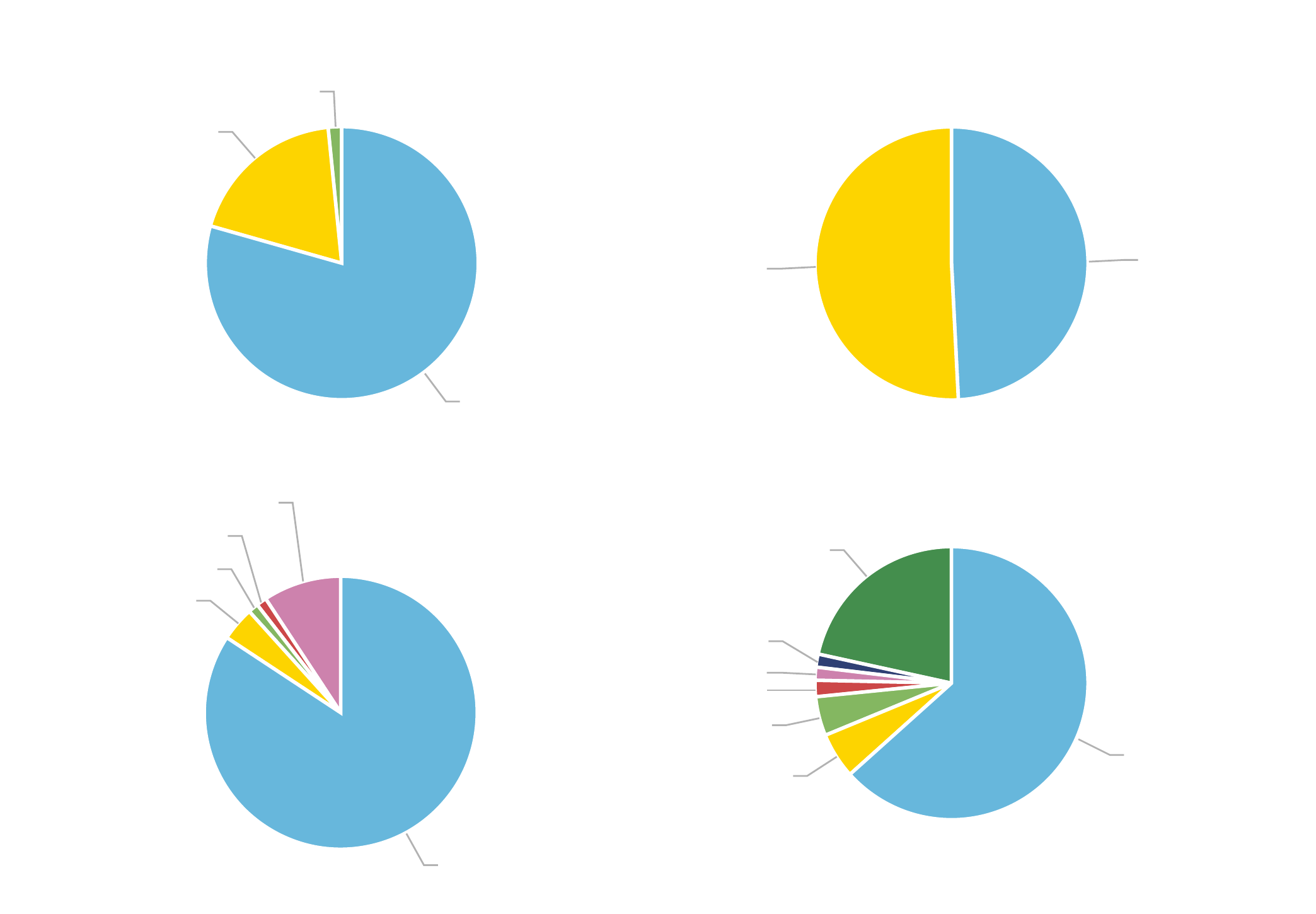}
	
%    \begin{tikzpicture}
%      \node at (0, 4) {\large{'Gender'}};
%      \pie[scale font, after number=\,\%]{0/, 1.6/Other, 19.0/Female, 79.4/Male};
%      
%      \node at (8, 4) {\large{'Do you play an instrument?'}};
%      \pie[scale font, pos={8, 0}]{0/, 0/, 0/, 0/, 49.2/yes, 50.8/no};
%    \end{tikzpicture}  
%    \begin{tikzpicture}[every legend entry/.append style={text width=2cm}]
%      \node at (0, 3.6) {\large{'Field of activity'}};
%      \pie[scale font, text=legend, before number=\ (, after number=\,\%), number in legend, pos={0, 0}]{66.13/IT, 5.65/mathematics, 4.84/electronics, 2.02/audio technology, 1.61/music, 1.61/social work, 22.52/other};
%    \end{tikzpicture}
%    \begin{tikzpicture}[every legend entry/.append style={text width=2cm}]
%      \node at (0, 3.6) {\large{'The country you grew up in'}};
%      \pie[scale font, text=legend, before number=\ (, after number=\,\%), number in legend, pos={0, 0}]{84.27/Germany, 4.03/India, 1.21/Iran, 1.21/Egypt, 9.27/other};
%    \end{tikzpicture}    
  \caption[Survey result: gender, instrument playing, origin and field of activity]{Distribution of gender, instrument playing, ``field of activity'' and origin among subjects.}
  \label{fig:poll-result-gender-instrument}
\end{figure}

\begin{table}[H]
  \begin{center}
    \begin{tabu}{ r | c || *{6}{c|} c }
       \multicolumn{2}{c ||}{\textbf{ground truth}} & \multicolumn{7}{c}{\textbf{answers}}\\
       type & human & $ \Sigma $ & $ k $ & $ h $ & $ k \land h $ & $ \lnot k \land h $ & $ k \land \lnot h $ & $ \lnot k \land \lnot h $ \\
       \hhline{==::=======}
       1 & yes & 496 &  76 & 343 &  74 & 269 &   2 & 151 \\
       \hhline{--||-------}
       2 & yes & 496 &  52 & 348 &  46 & 302 &   6 & 142 \\
       \hhline{--||-------}
       3 & no  & 496 &  11 &  97 &   4 &  93 &   7 & 392 \\
       \hhline{--||-------}
       4 & no  & 496 &  67 & 261 &  60 & 201 &   7 & 228 \\
       \hhline{--||-------}
       5 & no  & 248 &  20 & 151 &  20 & 131 &   0 &  97 \\
       \hhline{--||-------}
       6 & no  & 248 &  11 & 123 &   7 & 116 &   4 & 121 \\
       \hhline{--||-------}
       7 & no  & 248 &   8 & 124 &   6 & 118 &   2 & 122 \\
       \hhline{--||-------}
       8 & no  & 248 &  16 & 133 &  14 & 119 &   2 & 113
    \end{tabu}
    \caption[Raw melody responses]{$ \Sigma $ is the number of responses for the given type of melody. $ k $ is the number of those responses that stated to know the melody. $ h $ is the number that stated the melody to be composed by a human.\label{tab:raw-melody-responses}}
  \end{center}
\end{table}

\begin{figure}[H]
  \resizebox{\textwidth}{!}{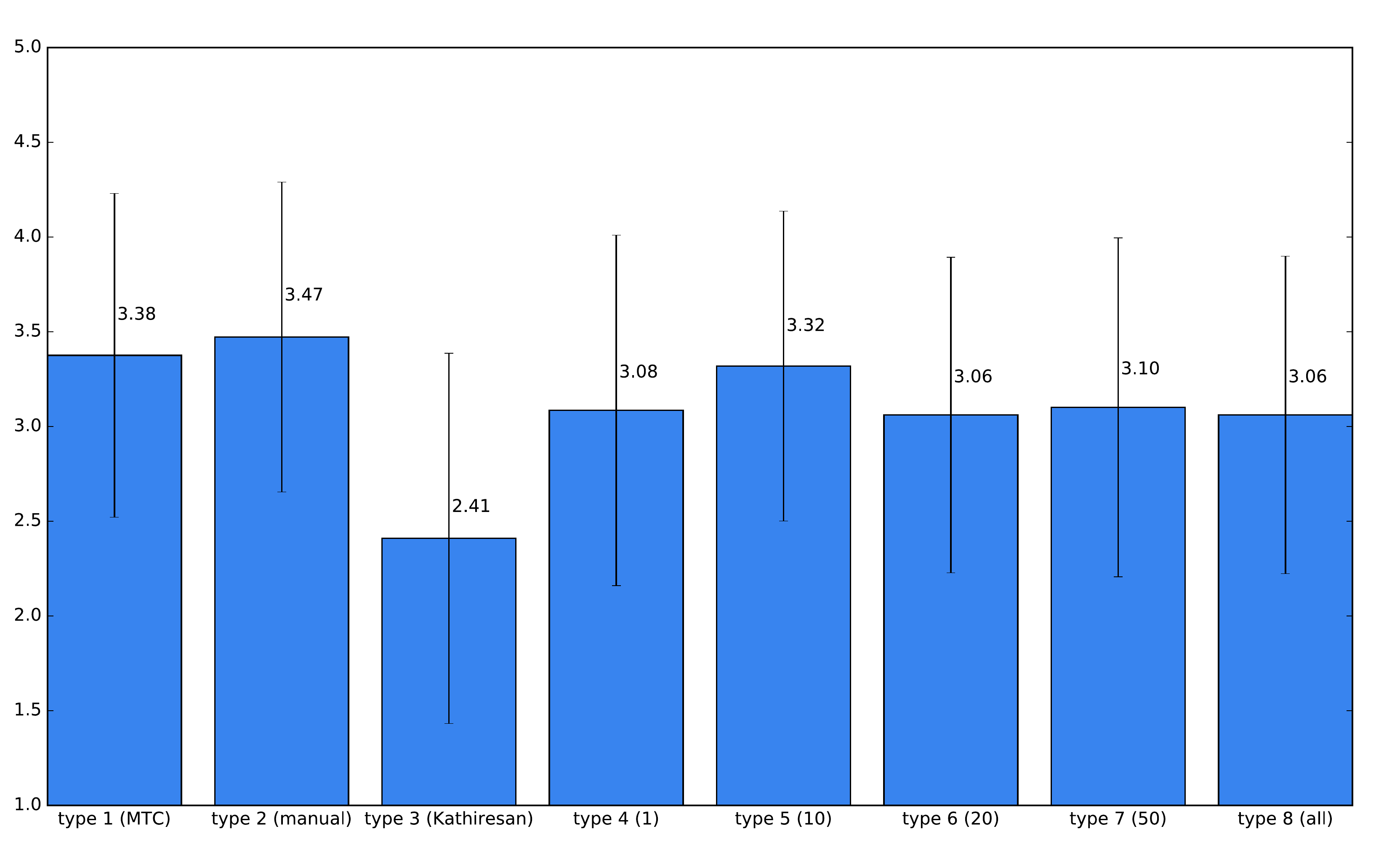}
  \caption[Survey result: liking of melody types]{Subjects had to rate how much they like every melody on a scale from 1 to 5 (``strongly dislike'',``dislike'',``neither like nor dislike'',``like'',``strongly like'').}
  \label{fig:poll-result-pleasure}
\end{figure}

%\begin{table}[h]
%  \begin{center}
%    \begin{subtable}{.4 \textwidth}
%      \begin{tabu}{|c|*{2}{|X[c]}|}
%        \hhline{-||--|}
%        \backslashbox{GT}{Pred.} & Human &      CGM\\ \hhline{=::==|}
%                           Human &   127 &      841\\ \hhline{-||--|}
%                             CGM &   119 &     1333\\ \hhline{-||--|}
%      \end{tabu}
%      \caption{For melodies of our algorithm (types 4\,--\,8) \label{tab:confusion-matrix-our-algorithm}}
%    \end{subtable}
%    \begin{subtable}{.4 \textwidth}
%      \begin{tabu}{|c|*{2}{|X[c]}|}
%        \hhline{-||--|}
%        \backslashbox{GT}{Pred.} & Human &      CGM\\ \hhline{=::==|}
%                           Human &   127 &      841\\ \hhline{-||--|}
%                             CGM &    11 &      473\\ \hhline{-||--|}
%      \end{tabu}
%      \caption{For melodies of Kathiresan's algorithm (type 3) \label{tab:confusion-matrix-original-algorithm}}
%    \end{subtable}  
%  \caption[confusion matrices evaluating results of listening test]{Confusion matrices visualizing the results of the conducted listening test. 'GT' stands for 'ground truth' while 'Pred.' stands for 'prediction'. GT is what the presented melody's class actually was, while Pred. is what the participants thought it's class would be.}    
%  \end{center}
%\end{table}

\subsection{Interpretation of the Results}
\label{subsec:interpretation-listening-test}
As mentioned in Section \ref{subsec:listenig-test} it was expected that the probe of people would not be representative. Viewing the age and gender distribution (Figures \ref{fig:poll-result-gender-instrument} and \ref{fig:poll-result-age}) compared to the German 'Zensus 2011' \cite{Zensus2011} proves that. Our subjects are younger and there are more males among them than the German average shows. This is probably due to the fact that the listening test was mainly advertised at the local university with a stress on the computer science department (clearly to see in Figure \ref{fig:poll-result-gender-instrument} ``field of activity''). However, the number of subjects playing a music instrument is more than 30\,\% higher than the German average of 17\,\% \cite{miz2014}. It can be assumed that subjects with musical experience are better in detecting non-human composed melodies. Thus, one should keep in mind that our subjects were probably more skeptical towards the presented melodies when classifying them as human or computer composed.
Another aspect becomes visible when analyzing the \emph{age} and the \emph{field of activity} responses: When opening the access to the internet without restrictions, the responses will end up having some noise. It does not seem plausible that we reached seven people younger than ten years old, nor people older than 90 that are able to operate a computer. This becomes even more clear when ``sdfgh'' is answered when asking for the \emph{field of activity}.

\begin{table}[t]
  \begin{center}
    \begin{tabu}{ r | c || *{6}{c|} c }
       \multicolumn{2}{c ||}{\textbf{ground truth}} & \multicolumn{7}{c}{\textbf{answers}}\\
       type & human & $ \frac{k}{\Sigma} $ & $ \frac{h}{\Sigma} $ & $ \frac{\lnot\text{k}\land\text{h}}{\lnot\text{k}} $ \\
       \hhline{==::=======}
       1 & yes & 15.32\,\% &  69.15\,\% &  64.05\,\% \\
       \hhline{--||-------}
       2 & yes & 10.48\,\% &  70.16\,\% &  68.02\,\% \\
       \hhline{--||-------}
       3 & no  &  2.22\,\% &  19.56\,\% &  19.18\,\% \\
       \hhline{--||-------}
       4 & no  & 13.51\,\% &  52.62\,\% &  46.85\,\% \\
       \hhline{--||-------}
       5 & no  &  8.06\,\% &  60.89\,\% &  57.46\,\% \\
       \hhline{--||-------}
       6 & no  &  4.44\,\% &  49.60\,\% &  48.95\,\% \\
       \hhline{--||-------}
       7 & no  &  3.23\,\% &  50.00\,\% &  49.17\,\% \\
       \hhline{--||-------}
       8 & no  &  6.45\,\% &  53.63\,\% &  51.29\,\%
    \end{tabu}
    \caption[Percental melody responses]{Percental values relevant for evaluation of the results. They were calculated from Table \ref{tab:raw-melody-responses}.}
    \label{tab:calc-melody-responses}
  \end{center}
\end{table}

Looking at Table \ref{tab:calc-melody-responses} we first want to see how well our subjects were able to distinguish human versus computer composed melodies. The row $ \frac{\lnot\text{k}\land\text{h}}{\lnot\text{k}} $ of Table \ref{tab:calc-melody-responses} shows how often a phrase was classified as ``human composed'', excluding all cases where the subject knew the phrase. These cases are excluded because the subjects can deduce from knowing the phrase that it must be human composed. However, we only want to look at deduction from the phrase itself. The results for types 1 to 3 match the results Kathiresan found (see Table \ref{tab:kath-confusion}). While KATH could only reach about 20\,\%, our proposed algorithm reached about 50\,\% of subjects thinking the phrase was human composed. This is as good as random guessing. So the subjects were not able to decide if the presented phrases were human or computer composed. Comparing this number with those of types one and two, it is interesting to see that the participants did not do much better for the human composed phrases.

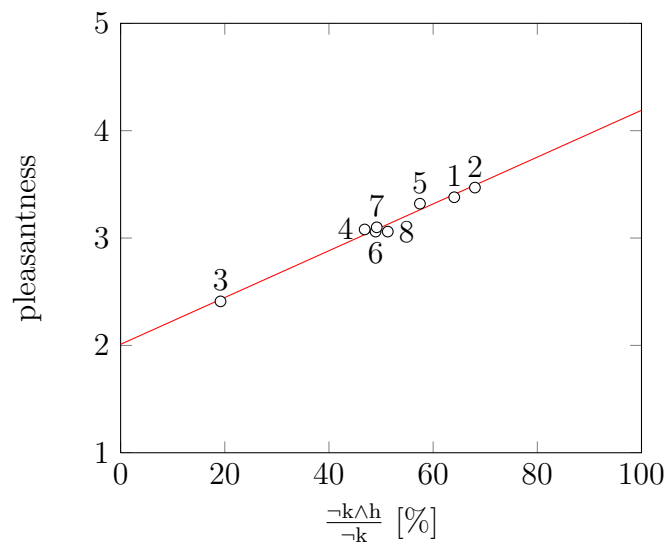
\begin{figure}[ht]
  \center
  \begin{tikzpicture}
    \begin{axis}[xmin=0,ymin=1,xmax=100, ymax=5,
                 xlabel=$ \frac{\lnot\text{k}\land\text{h}}{\lnot\text{k}} $\ {[\%]},
                 ylabel=pleasantness]
      \addplot+[only marks][
        black,
        mark=*,
        mark options={fill=white},
        visualization depends on=\thisrow{alignment} \as \alignment,
        nodes near coords, % Place nodes near each coordinate
        point meta=explicit symbolic, % The meta data used in the nodes is not explicitly provided and not numeric
        every node near coord/.style={anchor=\alignment} % Align each coordinate at the anchor 40 degrees clockwise from the right edge
        ] table [% Provide data as a table
         meta index=2 % the meta data is found in the third column
        ] {
x       y       label  alignment
64.05   3.38    1       -90
68.02   3.47    2       -90
19.18   2.41    3       -90
46.85   3.08    4         0
57.46   3.32    5       -90
48.95   3.06    6        90
49.17   3.10    7       -90
51.29   3.06    8       180
      };
    \addplot[mark=none, red] table % compute a linear regression from the input table
    {
x       y   
0       2.01
100     4.19
    };
    \end{axis}
  \end{tikzpicture}
  \caption[Correlation of phrase liking and knowing it]{Showing the correlation between how the subjects liked a set of phrases and how often they thought the melody was human composed. The red line shows the linear regression of the data points. For the null hypothesis that the error is 0 the two-sided p-value is $ 1.5 \cdot 10^{-6} $; the standard error is 0.12.}
  \label{fig:correlation-melody-results}
\end{figure}

The answers visualized in Figure \ref{fig:poll-result-pleasure} yield similar results. In fact the data is nearly perfectly linear correlated as shown in Figure \ref{fig:correlation-melody-results}. 

Finally, a side note on the high number of subjects stating to know the phrases of type four presented to them, even though they were composed by our algorithm: Since this type of phrases has only one melody as reference (training data), it isn't surprising that the results are very close to the input. Looking at Appendix \ref{appendix:close-results}, one can see that this is true indeed but they are not simple replicates of the input. This is theoretically possible as this is not checked explicitly by the proposed algorithm.

\section{Conclusion and Future Work}
\label{sec:conclusion}
In this work, we have shown that our system produces reasonable results, significantly outperforming the given baselines. Even though Markov models alone are seen as no proper method for algorithmic composition, we successfully showed that when combined with further methods they can yield much better results in terms of being closer to human composed melodies. This can be seen when comparing our results with the ones of Kathiresan \cite{Kathiresan2015}, whose basic algorithm solely relies on Markov models. Apart from the previous works, our algorithm outperforms a random guessing baseline, meaning that humans are not able to clearly distinguish its compositions from humans anymore.

Other aspects of melody composition are still open and could improve the results. Due to time restrictions, we had to leave those other aspects as future work. The most promising areas of future study could be the following:

\begin{itemize}
\item Generate more but shorter phrases to compose a livelier and longer melody. One could incorporate concepts like variation, repetition and similar in this process to imitate the human composition processes.
\item A method for finding proper phrase endings in terms of rhythm. As is, our algorithm might well end a phrase with a very atypical rhythmical pattern, because the transition probability vector of the duration Markov model is not modified except for the last note to make sure the composition does not get too long. A possible solution to this could be to compose the melody from beginning to end and from end to beginning (with a separately trained Markov model) simultaneously and meet somewhere in the middle of the composition.
\item Enrich musical diversity by supporting different keys, modes (major, minor, etc.) and time-signatures.
\item The contour clustering uses a fixed number of clusters. However, in future work a more refined method of choosing the clustering threshold depending on the presented training data could improve the algorithm.
\item Utilize a similarity function to avoid plain rehashing or replicating of the input data.
\item Several tweaking parameters were chosen by small experiments because the resources for an intense study to find the best choices were missing. Namely the low-pass filter threshold in Section \ref{subsubsec:contours}, Step \ref{clustering:fft}, the weighting parameter $\gamma$ in Section \ref{subsubsec:contours}, Step \ref{clustering:choose-cluster}, and the standard deviation values in section \ref{subsubsec:following-contours} that determine how accurately the algorithm follows the learned contours.
\end{itemize}

The interested reader can find the source code of the proposed method on GitHub (\url{https://github.com/roba91/melody-composer}).

%\section{Future Work}
%\label{sec:future_work}
%\subsection{Phrase Endings}
%\label{subsec:phrase_endings}
%\subsection{Concatenating Phrases}
%\label{subsec:concatenating_phrases}
%\subsection{Musical Diversity}
%\label{subsec:musical_diversity}
%\subsubsection{Time-Signature}
%\label{subsubsec:time-signature}
%\subsubsection{Keys and Modi}
%\label{subsubsec:keys_and_modi}

\onecolumn
% einfacher Zeilenabstand
\singlespacing
% Literaturliste soll im Inhaltsverzeichnis auftauchen
\newpage
\phantomsection
\addcontentsline{toc}{section}{Bibliography}
% Literaturverzeichnis anzeigen
\renewcommand\refname{Bibliography}
% Festlegung Art der Zitierung - Havardmethode: Abkuerzung Autor + Jahr
\bibliographystyle{alpha}
\bibliography{../Sosse/sources}

%% Index soll Stichwortverzeichnis heissen
% \newpage
% % Stichwortverzeichnis soll im Inhaltsverzeichnis auftauchen
% \addcontentsline{toc}{section}{Stichwortverzeichnis}
% \renewcommand{\indexname}{Stichwortverzeichnis}
% % Stichwortverzeichnis endgueltig anzeigen
% \printindex

\onehalfspacing
% evtl. Anhang
\newpage
\phantomsection
\addcontentsline{toc}{section}{Appendix}
\fancyhead[L]{Appendix} %Kopfzeile links
%------------------ !!! RE-INCLUDE !!!------------------------
\appendix
\newcommand{\nocontentsline}[3]{}
\newcommand{\tocless}[2]{\bgroup\let\addcontentsline=\nocontentsline#1{#2}\egroup}
\section*{Appendices}\label{appendix}
\tocless\section{Screen shots of the conducted listening test}
\label{appendix:poll}
\begin{figure}[H]
\begin{center}
\includegraphics[width=\textwidth]{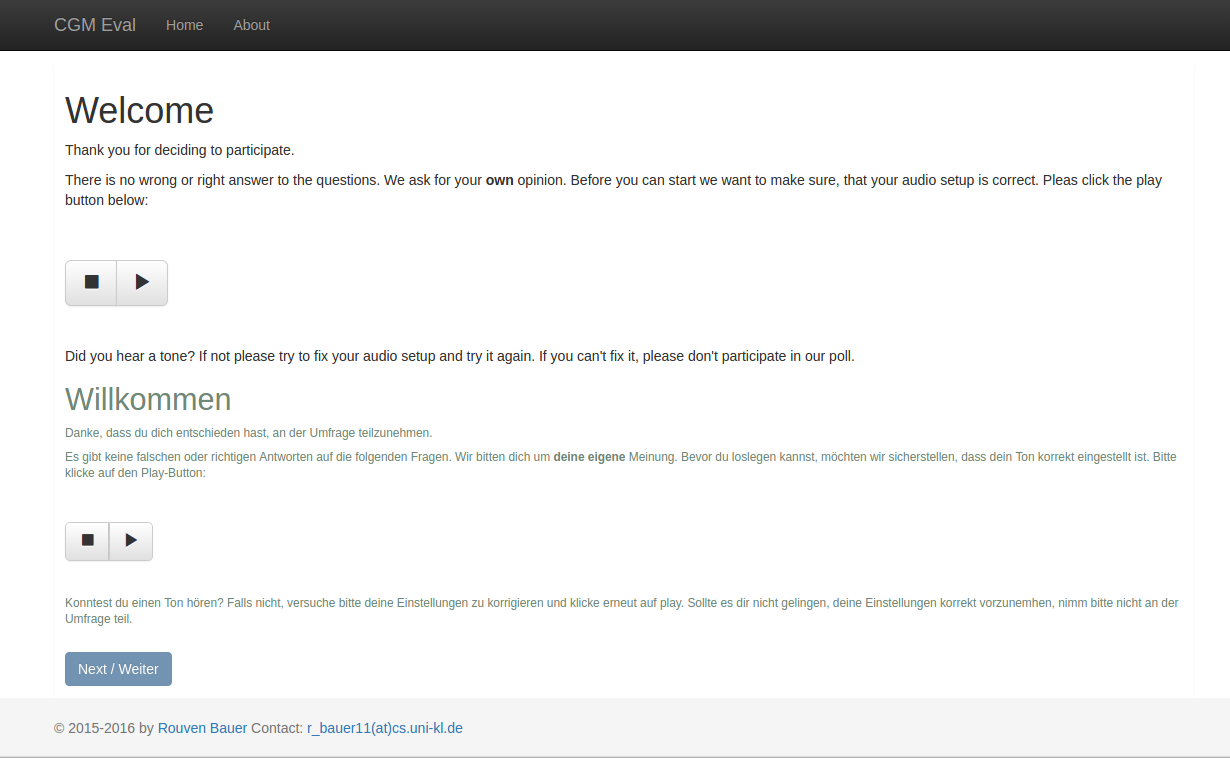}
\caption*{Screen shot of test stimulus of conducted listening test.}
\end{center}
\end{figure}

\newpage

\begin{figure}[H]
\begin{center}
\includegraphics[width=\textwidth]{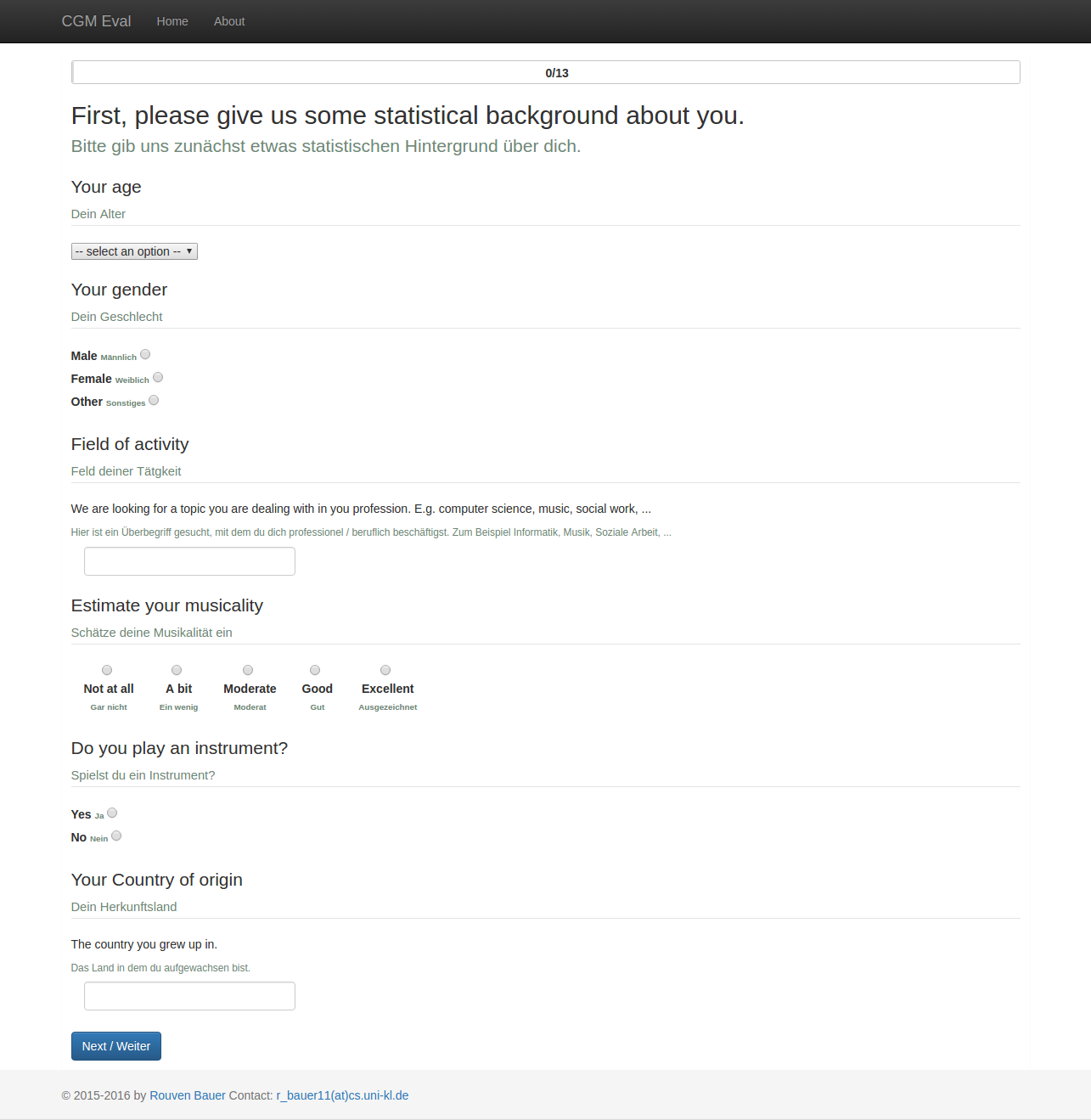}
\caption*{Screen shot of personal questions of conducted listening test.}
\end{center}
\end{figure}

\newpage

\begin{figure}[H]
\begin{center}
\includegraphics[width=\textwidth]{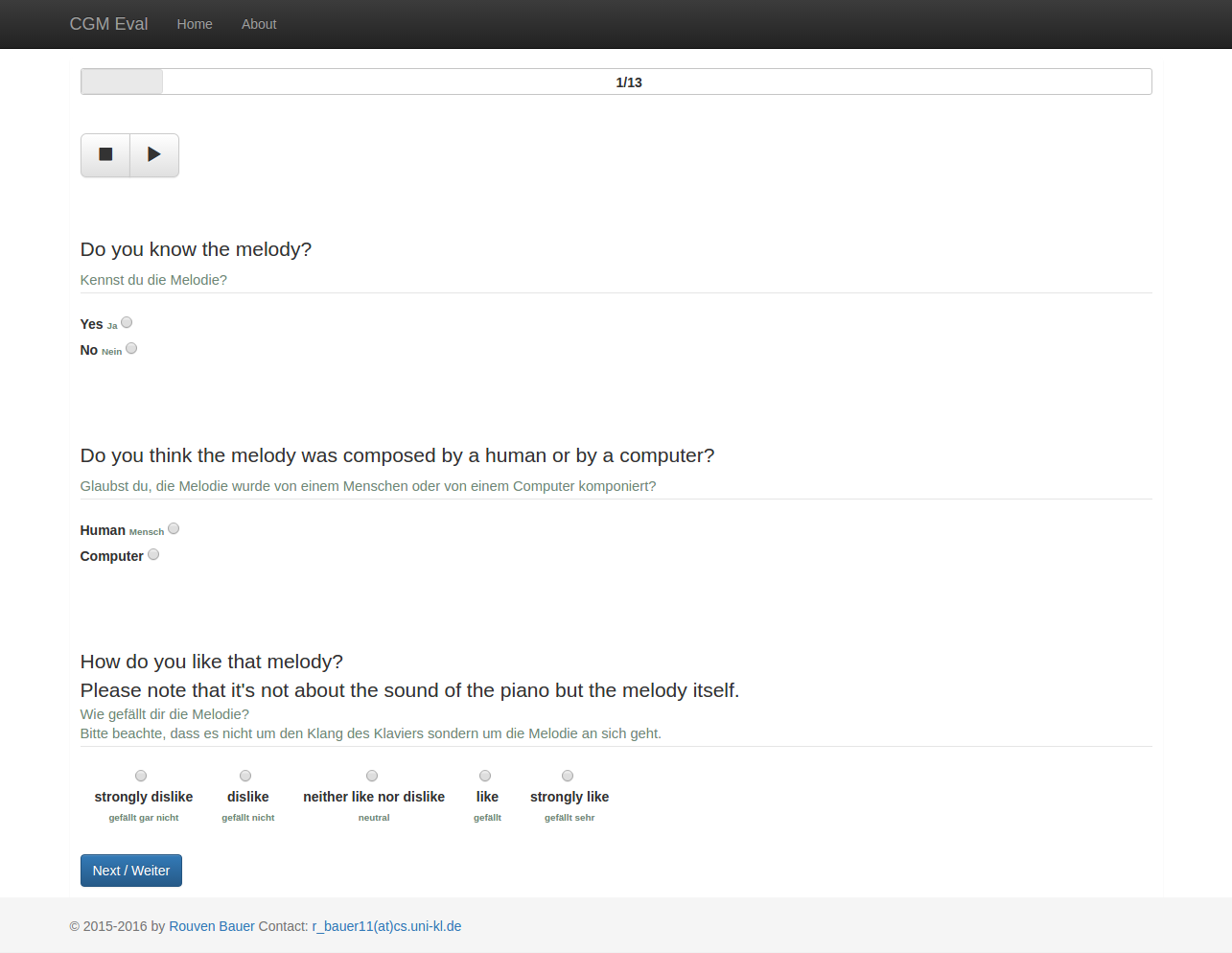}
\caption*{Screen shot of melody questions of conducted listening test. Each participant answered twelve of them.}
\end{center}
\end{figure}

\newpage

\noindent\begin{minipage}{\textwidth}
\tocless\section{Composition results with one training melody}
\label{appendix:close-results}
\noindent{}Note that the training melody was transposed to c major before learning from it and composing new melodies. Keep that in mind when comparing the input with the outputs. KATH on the other hand, preserves the key of the training melodies.
\end{minipage}

\newpage

\tocless\subsection{Melody 1}
\begin{figure}[H]
  \begin{subfigure}{\textwidth}
    \center
    \resizebox{0.9\textwidth}{!}{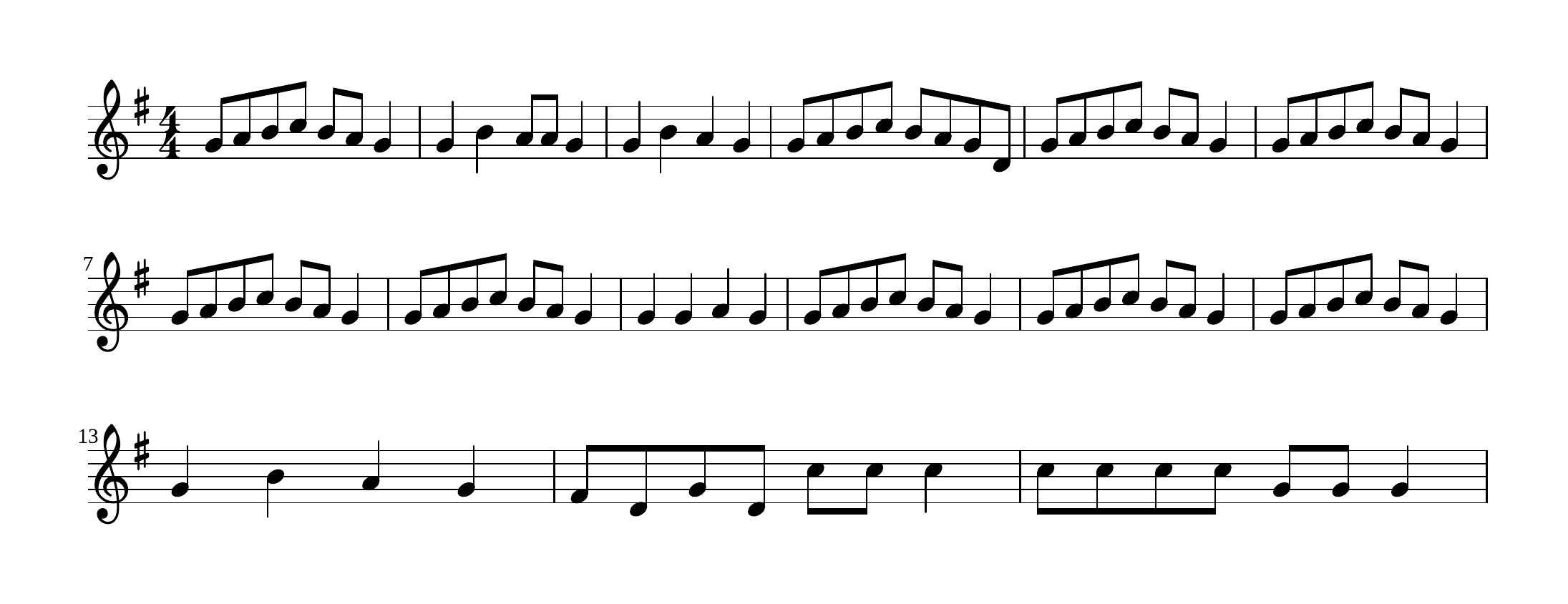}
    \caption{Training melody}
  \end{subfigure}
  \begin{subfigure}{\textwidth}
    \center
    \resizebox{0.9\textwidth}{!}{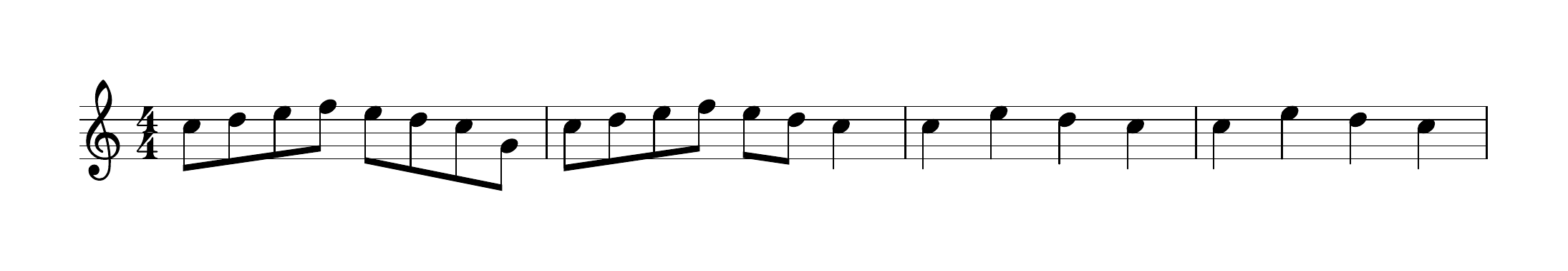}
    \caption{composed by our algorithm}
  \end{subfigure}
  \begin{subfigure}{\textwidth}
    \center
    \resizebox{0.9\textwidth}{!}{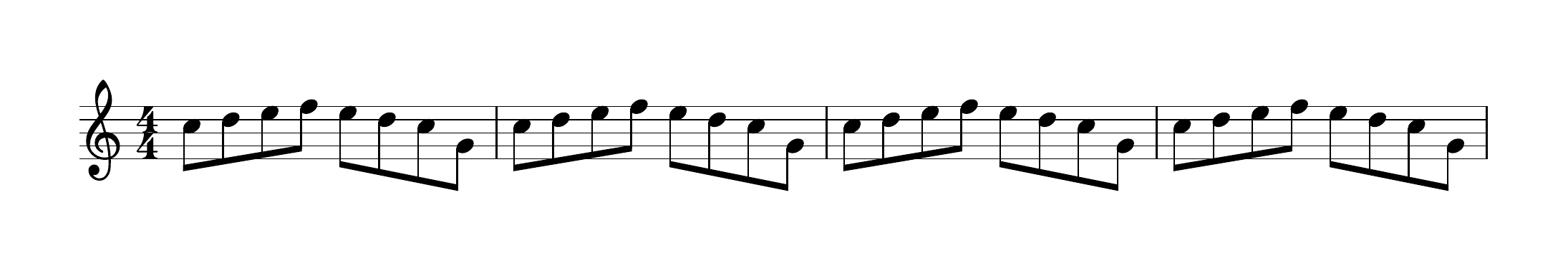}
    \caption{composed by our algorithm}
  \end{subfigure}
  \begin{subfigure}{\textwidth}
    \center
    \resizebox{0.9\textwidth}{!}{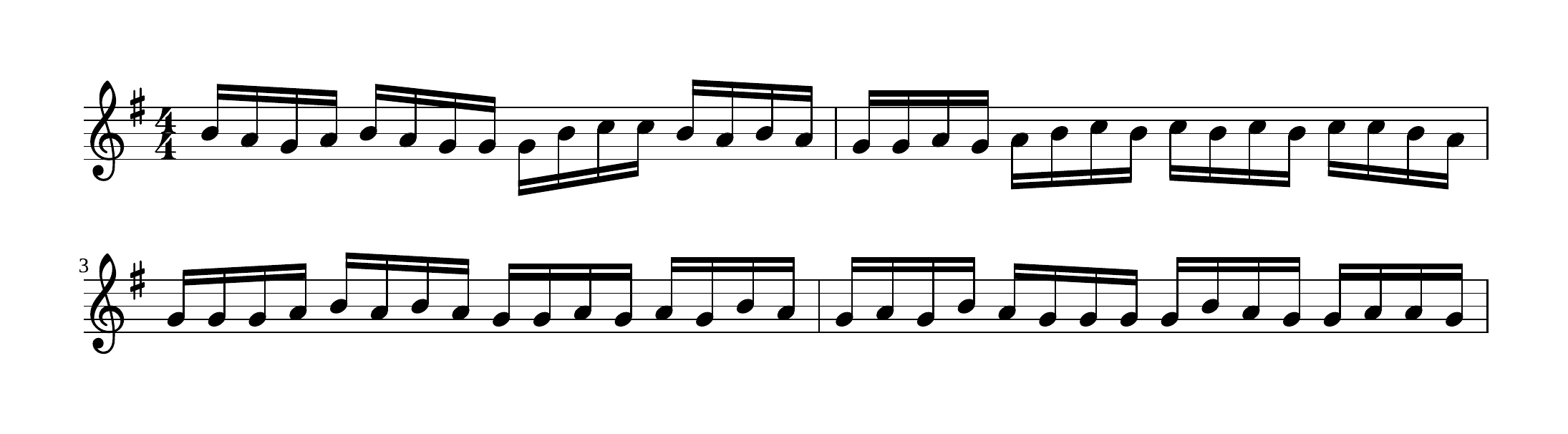}
    \caption{composed by KATH}
  \end{subfigure}
  \begin{subfigure}{\textwidth}
    \center
    \resizebox{0.9\textwidth}{!}{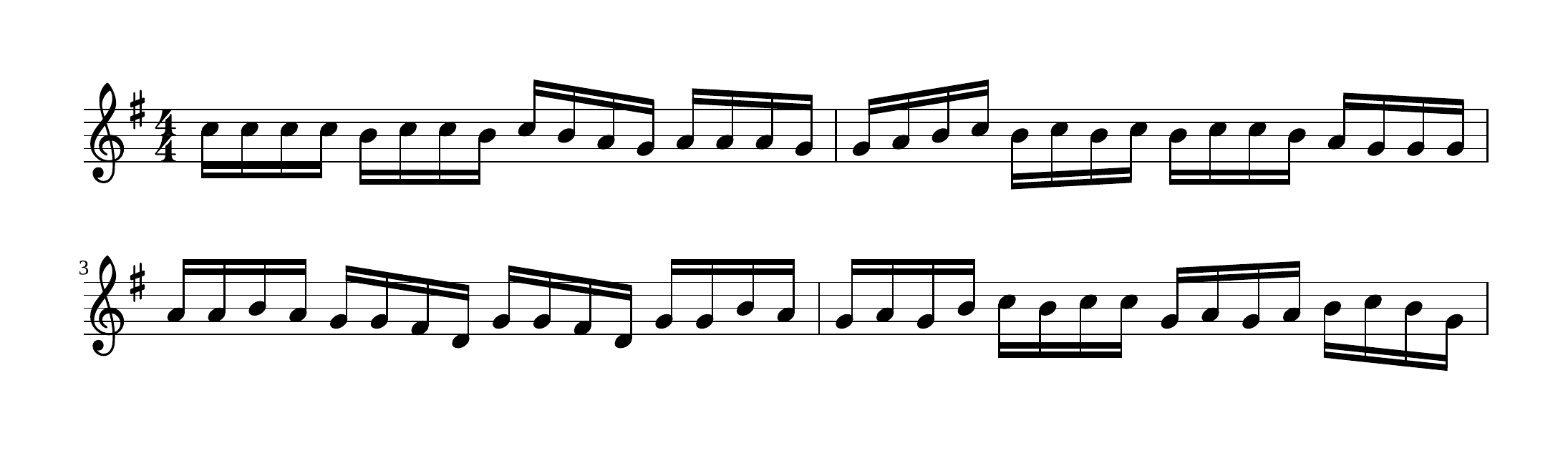}
    \caption{composed by KATH}
  \end{subfigure}  
\end{figure}

\newpage

\tocless\subsection{Melody 2}
\begin{figure}[H]
  \begin{subfigure}{\textwidth}
    \center
    \resizebox{0.9\textwidth}{!}{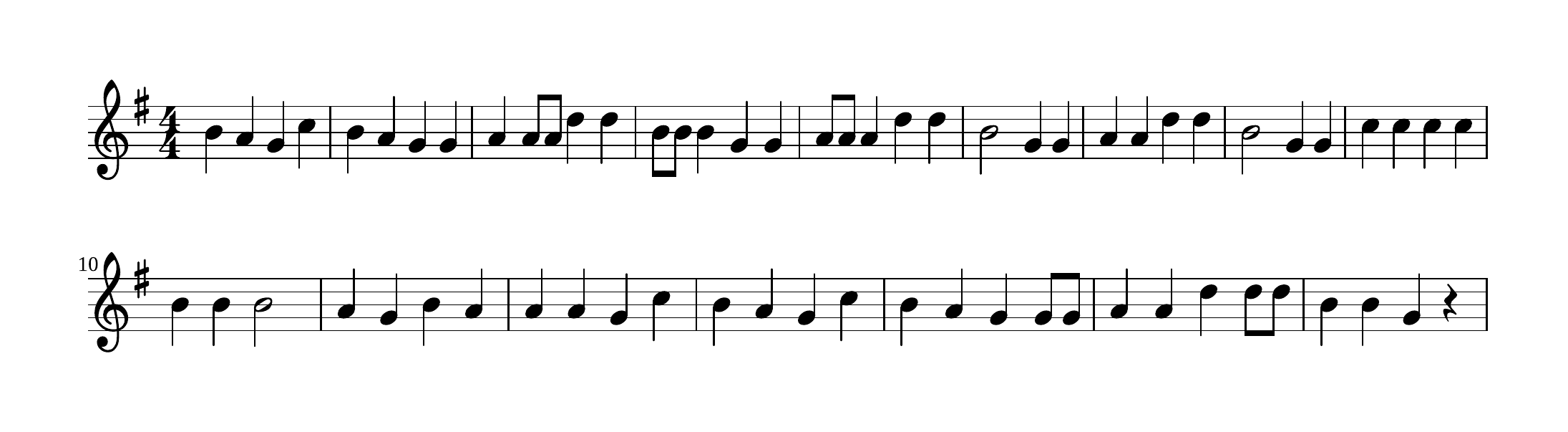}
    \caption{Training melody}
  \end{subfigure}
  \begin{subfigure}{\textwidth}
    \center
    \resizebox{0.9\textwidth}{!}{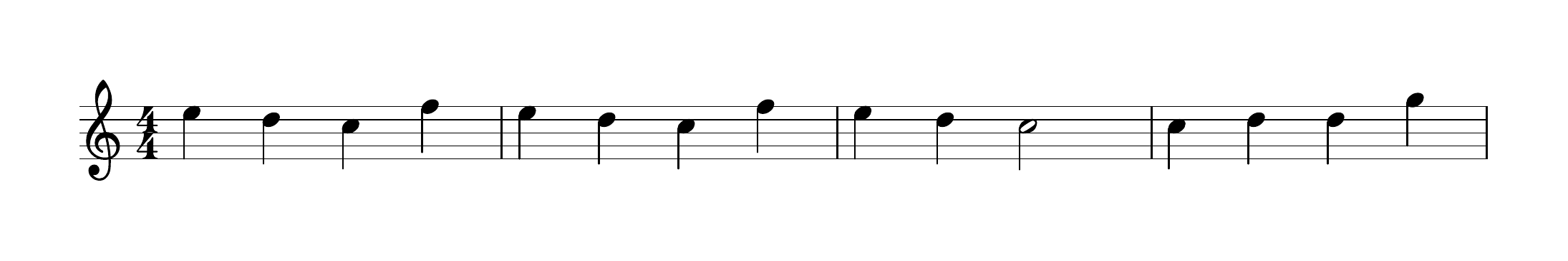}
    \caption{Algorithmically composed melody a}
  \end{subfigure}
  \begin{subfigure}{\textwidth}
    \center
    \resizebox{0.9\textwidth}{!}{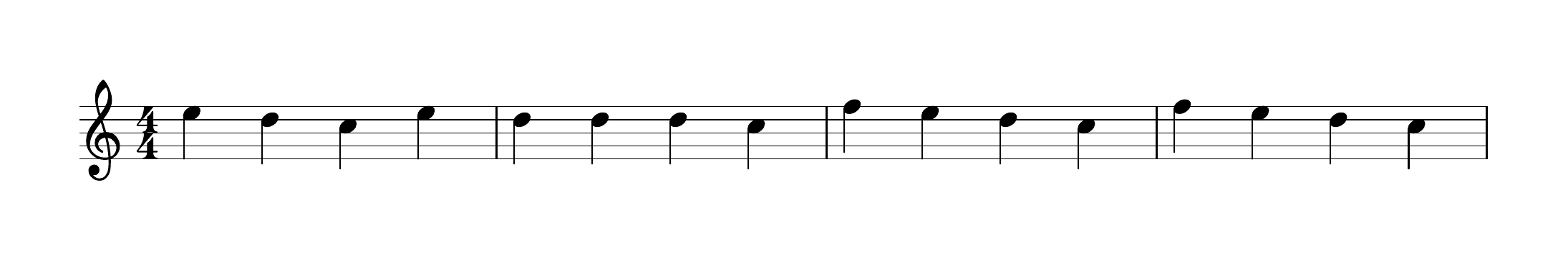}
    \caption{composed by our algorithm}
  \end{subfigure}
  \begin{subfigure}{\textwidth}
    \center
    \resizebox{0.9\textwidth}{!}{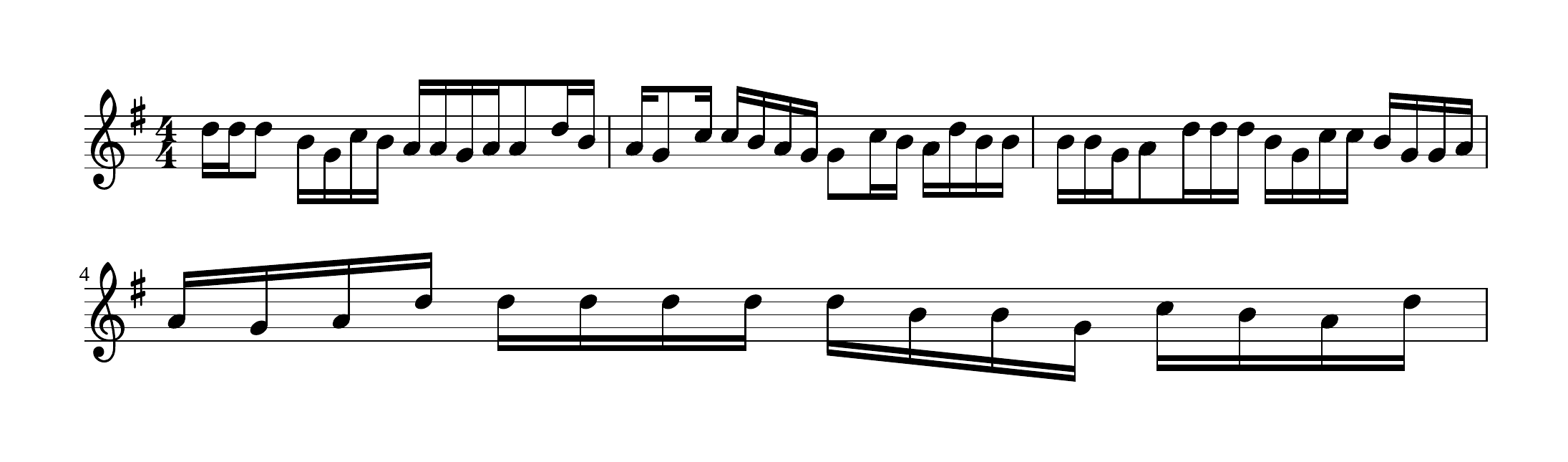}
    \caption{composed by KATH}
  \end{subfigure}
  \begin{subfigure}{\textwidth}
    \center
    \resizebox{0.9\textwidth}{!}{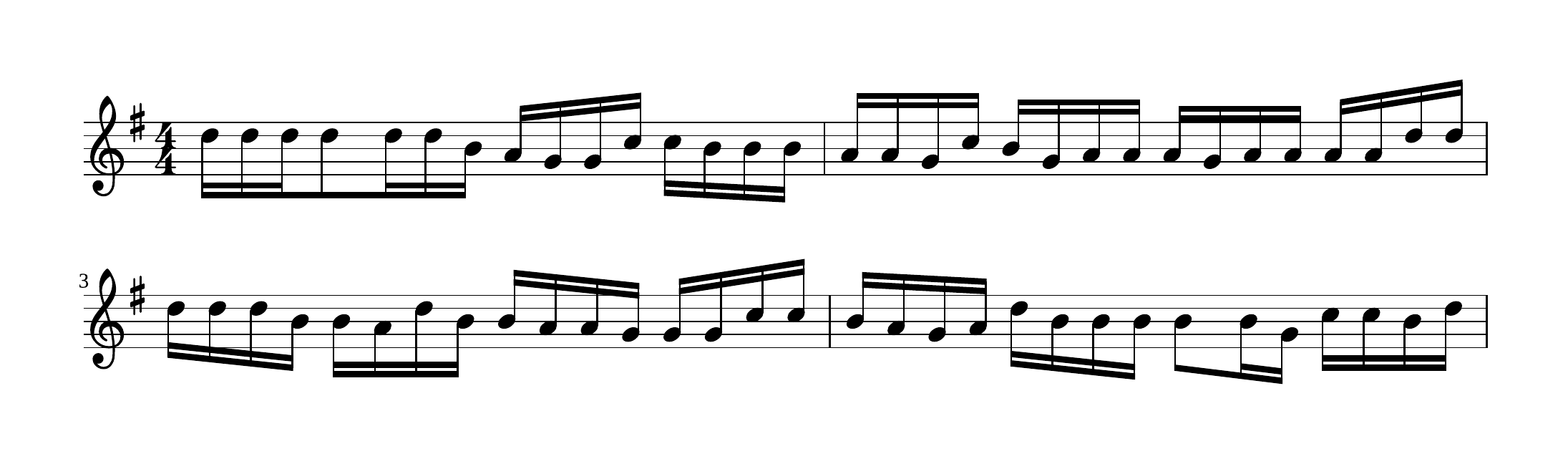}
    \caption{composed by KATH}
  \end{subfigure} 
\end{figure}

\newpage

\tocless\subsection{Melody 3}
\noindent{}You might notice that both melodies generated by our algorithm are the same. Since a lot of randomness is involved in the composition process this is (depending on the width of the search tree) unlikely but possible. We did not filter those occurrences out because we did not want to modify the results in any matter. We actually didn't have a look at the output before starting the listening test.
\begin{figure}[H]
  \begin{subfigure}{\textwidth}
    \center
    \resizebox{0.9\textwidth}{!}{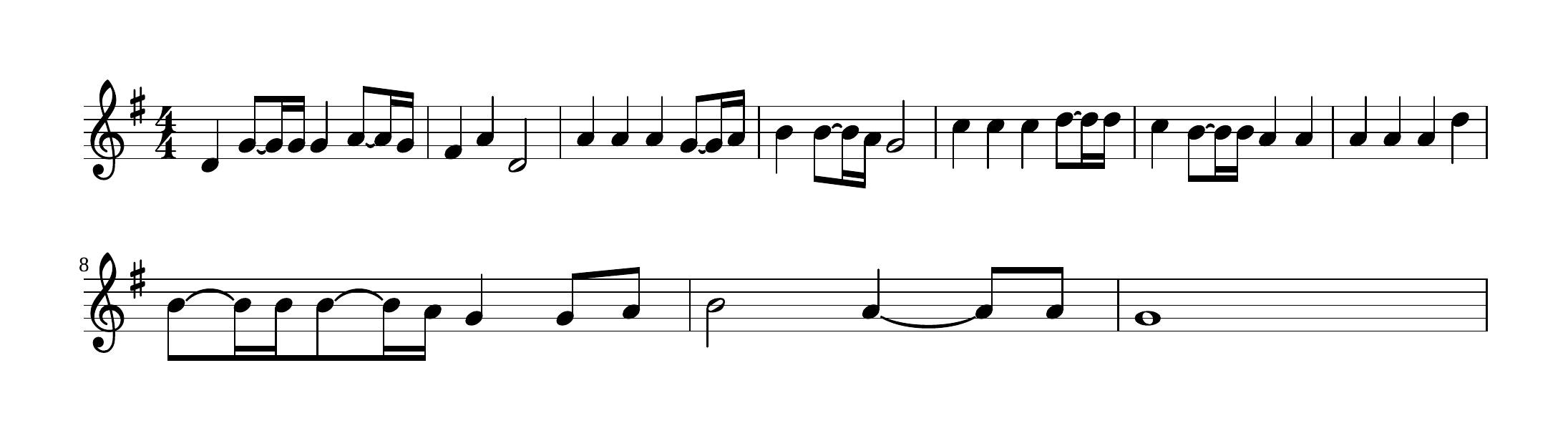}
    \caption{Training melody}
  \end{subfigure}
  \begin{subfigure}{\textwidth}
    \center
    \resizebox{0.9\textwidth}{!}{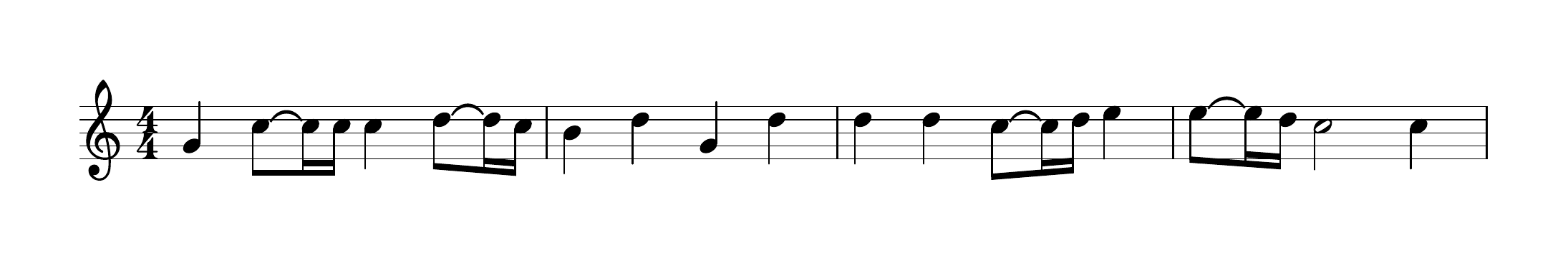}
    \caption{composed by our algorithm}
  \end{subfigure}
  \begin{subfigure}{\textwidth}
    \center
    \resizebox{0.9\textwidth}{!}{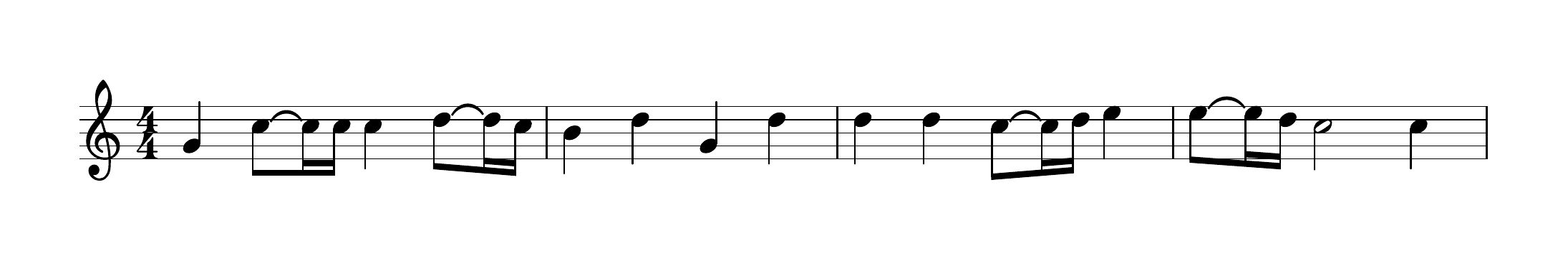}
    \caption{composed by our algorithm}
  \end{subfigure}
  \begin{subfigure}{\textwidth}
    \center
    \resizebox{0.9\textwidth}{!}{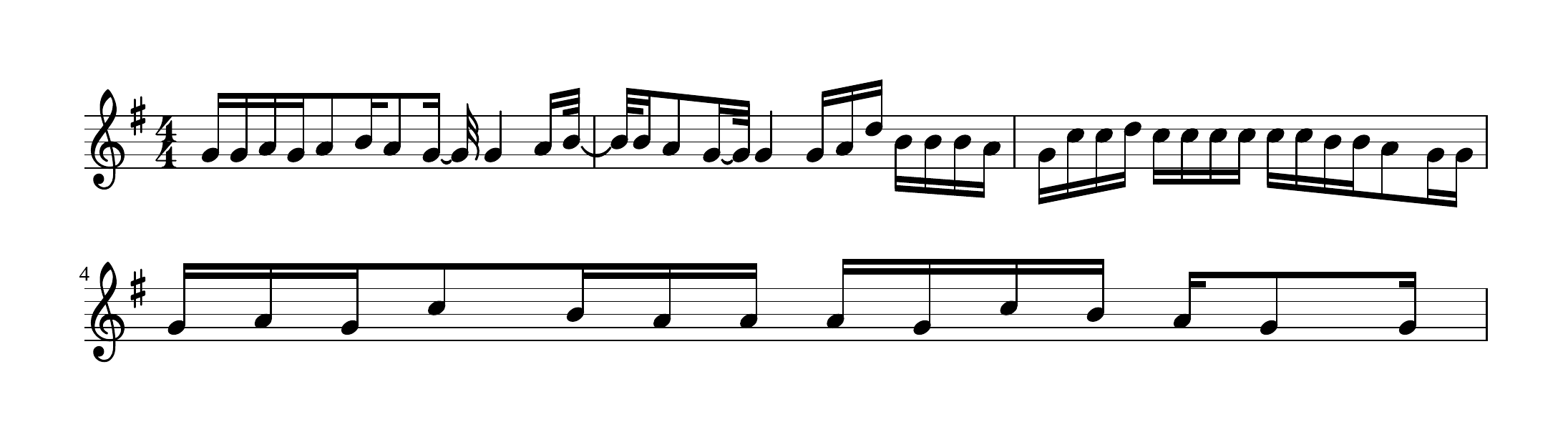}
    \caption{composed by KATH}
  \end{subfigure}
  \begin{subfigure}{\textwidth}
    \center
    \resizebox{0.9\textwidth}{!}{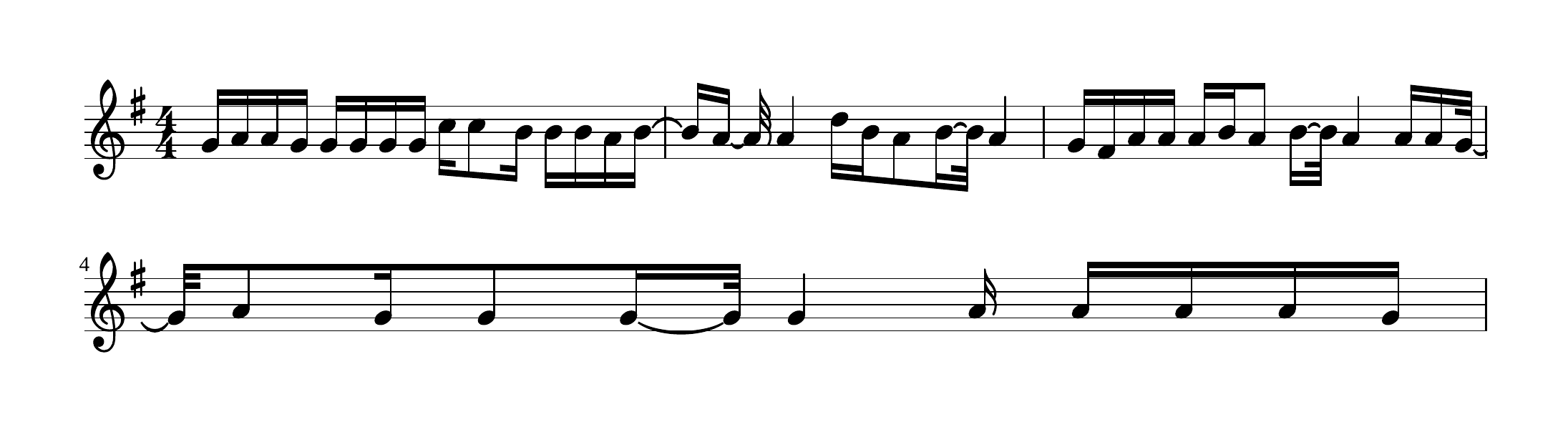}
    \caption{composed by KATH}
  \end{subfigure} 
\end{figure}

\newpage

\tocless\subsection{Melody 4}
\begin{figure}[H]
  \begin{subfigure}{\textwidth}
    \center
    \resizebox{0.9\textwidth}{!}{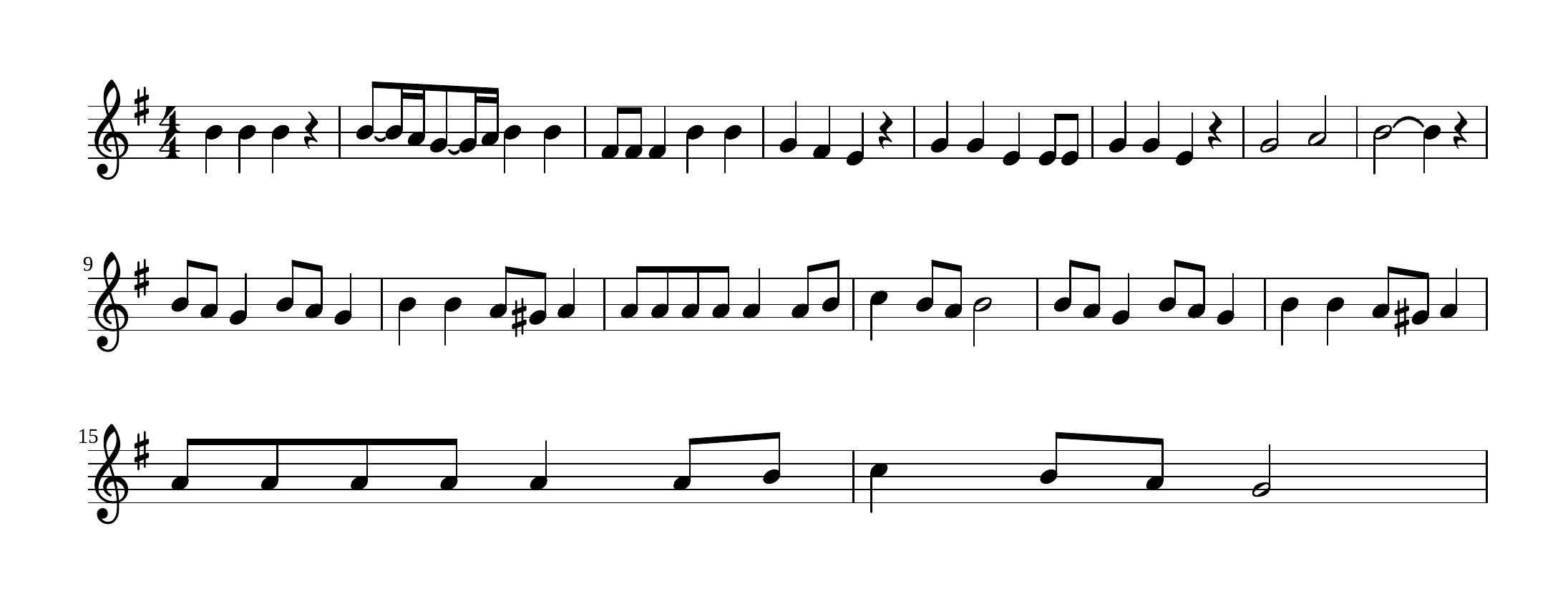}
    \caption{Training melody}
  \end{subfigure}
  \begin{subfigure}{\textwidth}
    \center
    \resizebox{0.9\textwidth}{!}{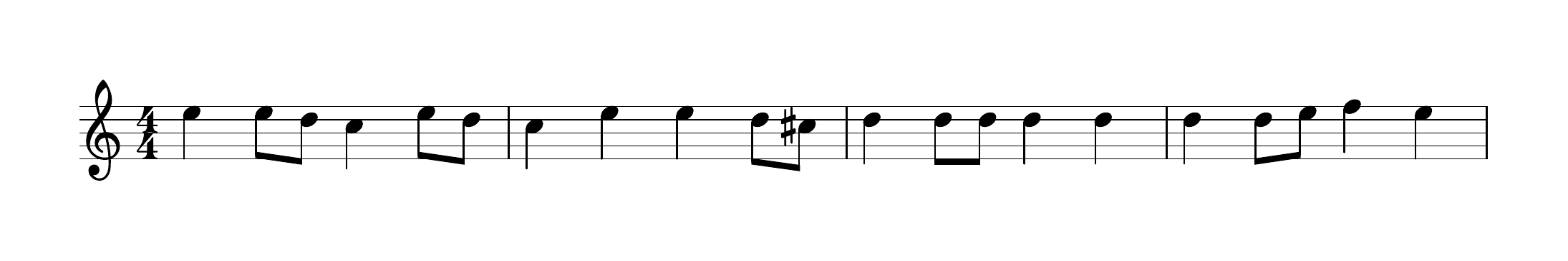}
    \caption{composed by our algorithm}
  \end{subfigure}
  \begin{subfigure}{\textwidth}
    \center
    \resizebox{0.9\textwidth}{!}{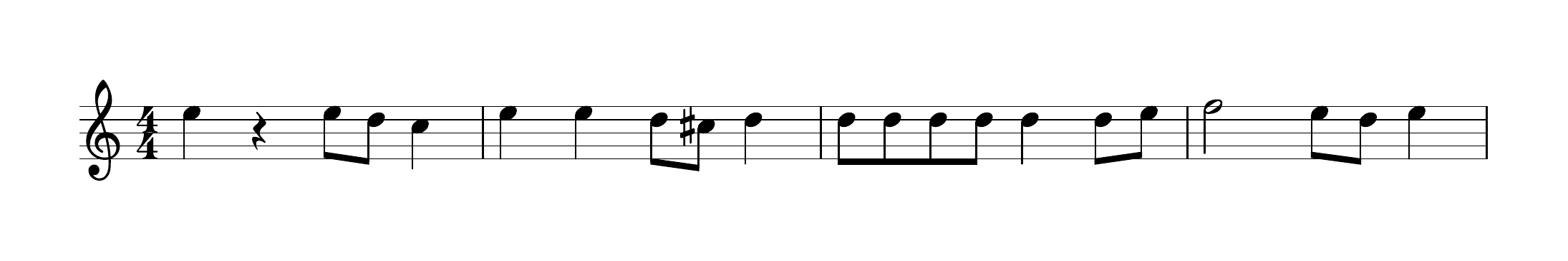}
    \caption{composed by our algorithm}
  \end{subfigure}
  \begin{subfigure}{\textwidth}
    \center
    \resizebox{0.9\textwidth}{!}{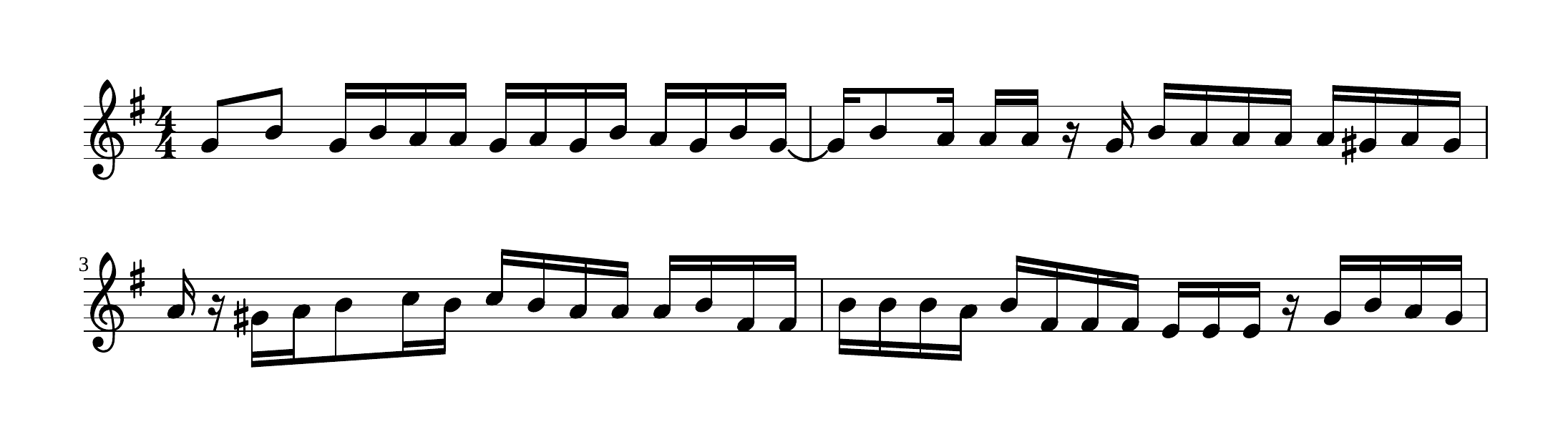}
    \caption{composed by KATH}
  \end{subfigure}
  \begin{subfigure}{\textwidth}
    \center
    \resizebox{0.9\textwidth}{!}{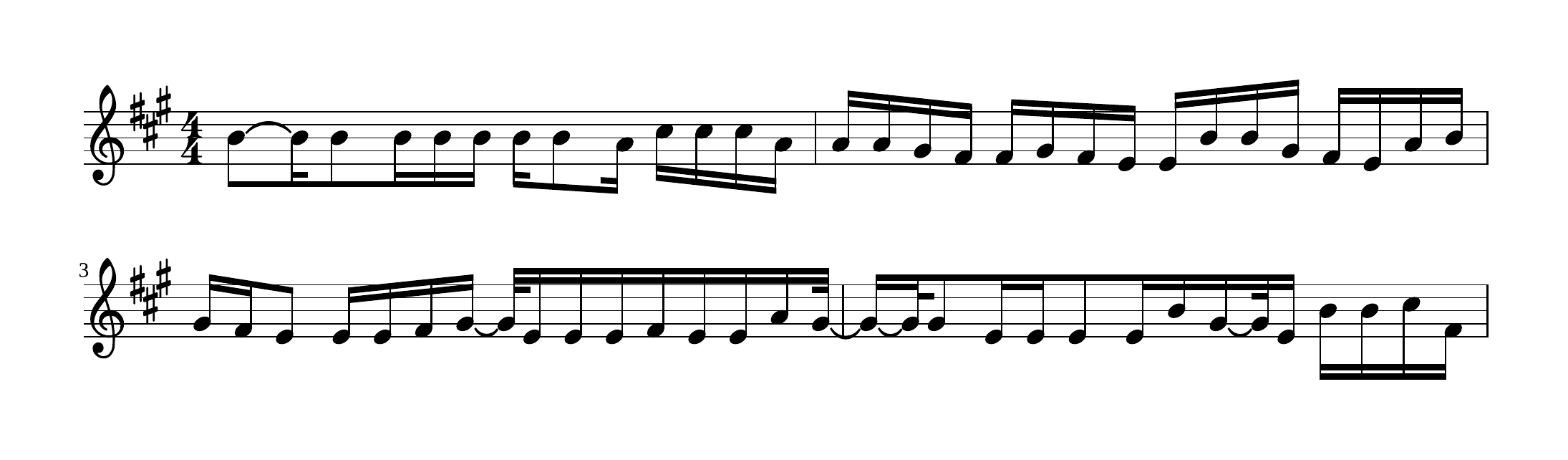}
    \caption{composed by KATH}
  \end{subfigure}
\end{figure}

\newpage

\tocless\subsection{Melody 5}
\begin{figure}[H]
  \begin{subfigure}{\textwidth}
    \center
    \resizebox{0.9\textwidth}{!}{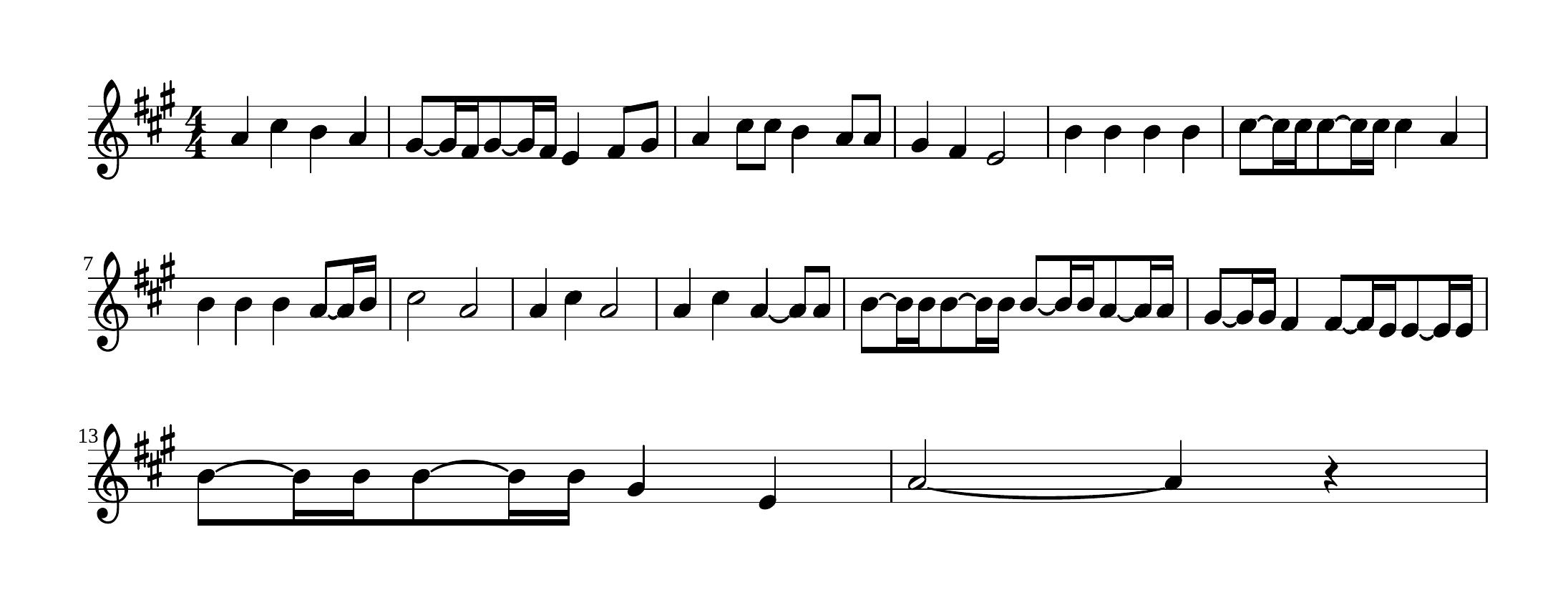}
    \caption{Training melody}
  \end{subfigure}
  \begin{subfigure}{\textwidth}
    \center
    \resizebox{0.9\textwidth}{!}{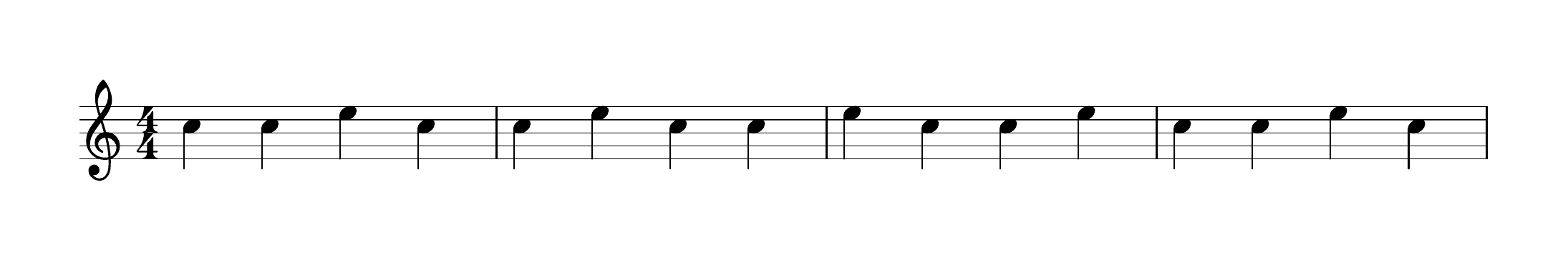}
    \caption{composed by our algorithm}
  \end{subfigure}
  \begin{subfigure}{\textwidth}
    \center
    \resizebox{0.9\textwidth}{!}{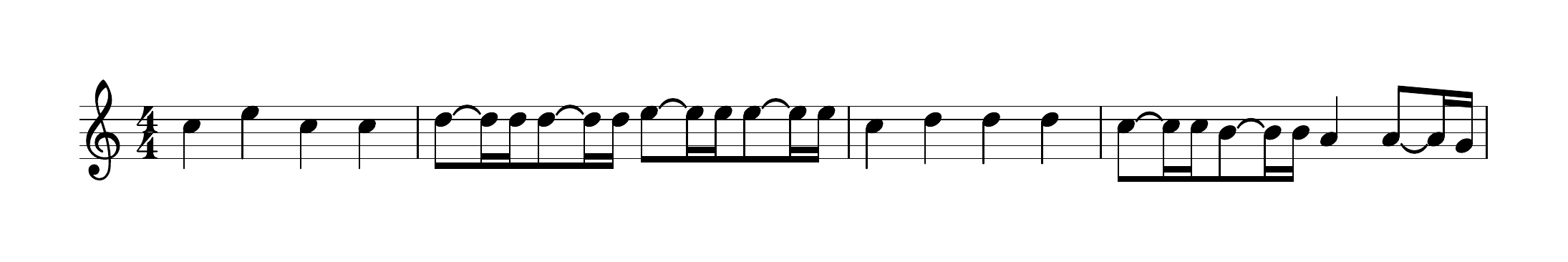}
    \caption{composed by our algorithm}
  \end{subfigure}
  \begin{subfigure}{\textwidth}
    \center
    \resizebox{0.9\textwidth}{!}{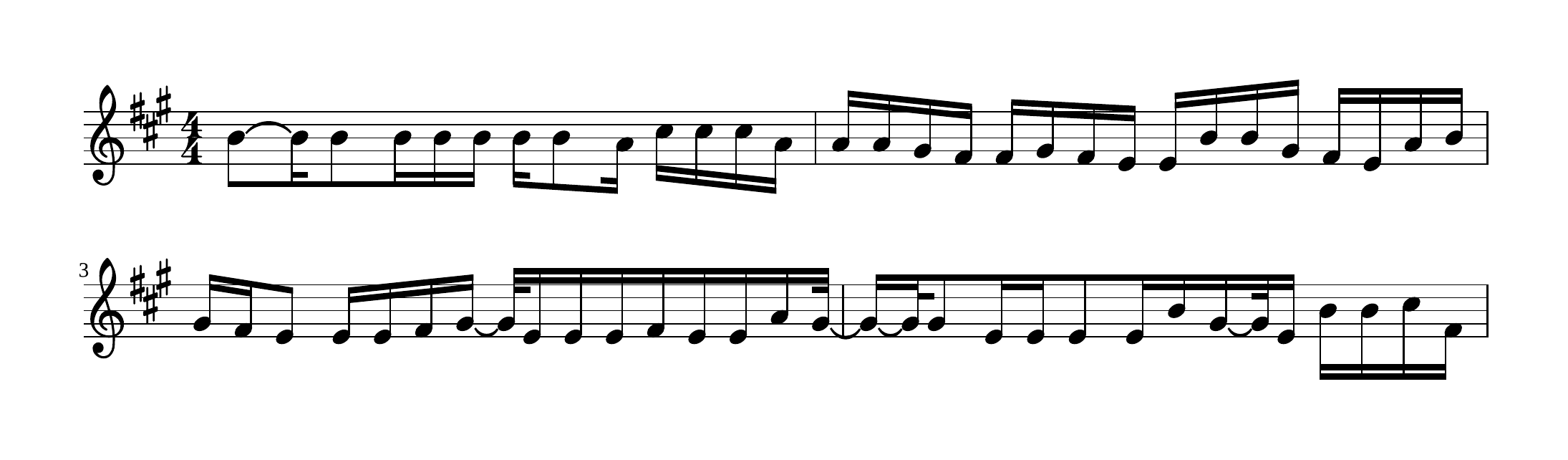}
    \caption{composed by KATH}
  \end{subfigure}
  \begin{subfigure}{\textwidth}
    \center
    \resizebox{0.9\textwidth}{!}{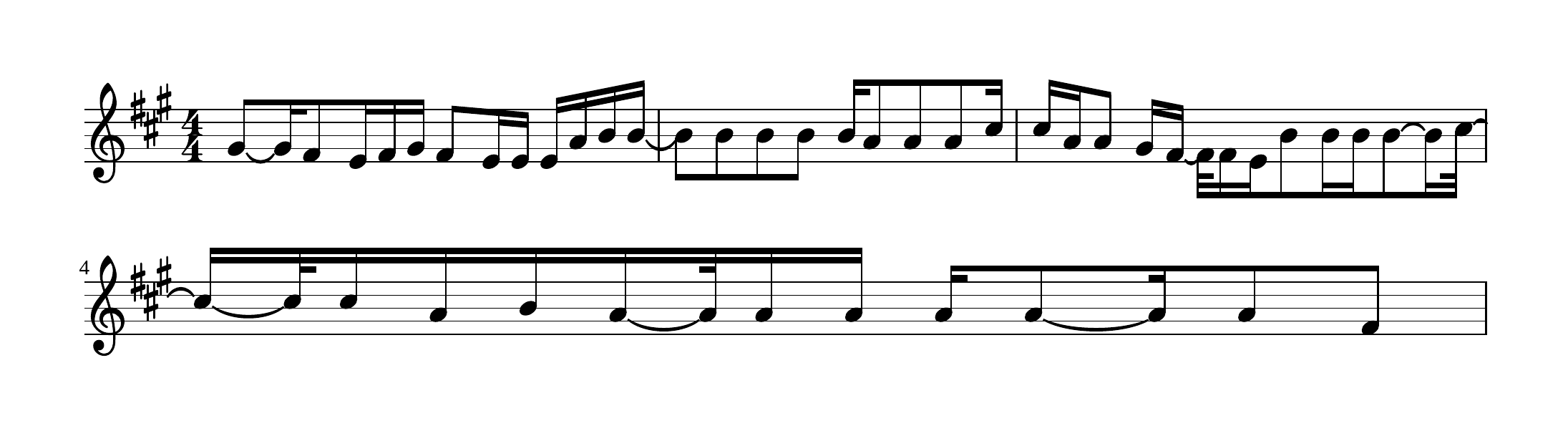}
    \caption{composed by KATH}
  \end{subfigure} 
\end{figure}

%\newpage
%
%\section{Source code}
%\label{appendix:source_code}
%
%\lstinputlisting{code/automatic_content_generation.py}
%
%\lstinputlisting{code/melody_generator.py}
%
%\lstinputlisting{code/statistics_extraction.py}
%
%\lstinputlisting{code/mm.py}
%
%\lstinputlisting{code/contour.py}
%
%\lstinputlisting{code/midi.py}
%
%\lstinputlisting{code/util.py}
%
%\lstinputlisting{code/config.py}

%------------------ !!! RE-INCLUDE !!!------------------------

% Eidesstattliche Erklärung
%\addcontentsline{toc}{section}{Eidesstattliche Erklärung}
%------------------ !!! RE-INCLUDE !!!------------------------
%\include{erklaerung}
%------------------ !!! RE-INCLUDE !!!------------------------

% leere Abschlussseite
%\newpage
%\thispagestyle{empty} % erzeugt Seite ohne Kopf- / Fusszeile
%\section*{ }

\end{document}